\documentclass{article}
\usepackage[utf8]{inputenc}

\usepackage{graphicx}

\usepackage{amsmath,amssymb,amsfonts,graphicx,latexsym,a4wide, enumerate, xcolor}
\usepackage{srcltx}

\usepackage{multirow}
\usepackage{mathtools}
\usepackage{tabularx}
\usepackage{makecell}
\usepackage{subcaption}
\usepackage{adjustbox}

\usepackage[backref=page]{hyperref}
\renewcommand*{\backref}[1]{}

\usepackage[letterpaper, margin=1in]{geometry} 
\usepackage[ruled,linesnumbered]{algorithm2e}

\newcommand{\R}{\mathbb{R}}

\definecolor{myCyan}{HTML}{08fcfc}

\newtheorem{rem}{Remark}[section]

\newtheorem{example}{Example}[section]

\newcommand{\Rn}{\mathbb{R}^n}
\newcommand{\E}{\mathbb{E}}

\newcommand{\Sepsbwd}{S}

\newcommand{\epsgeneratorfwd}{\mathcal{A}_{\epsilon,t}}
\newcommand{\epsgeneratorinv}{\mathcal{A}_{\epsilon,T-t}}
\newcommand{\mufwd}{\nu}
\newcommand{\mubwd}{\mu}
\newcommand{\Sxx}{Q}
\newcommand{\Sx}{q}
\newcommand{\Sc}{r}

\newcommand{\ydata}{y_{obs}}
\newcommand{\Bayesparam}{\theta}
\newcommand{\Gpriorcenter}{\theta^P}
\newcommand{\Pprior}{P_{prior}}
\newcommand{\Dinsde}{D}
\DeclareMathOperator{\Tr}{Tr}
\newcommand{\NNparam}{W}

\newcommand{\ny}{N_y}
\newcommand{\nz}{N_z}
\newcommand{\nt}{N_t}

\newcommand{\nprior}{M}

\newcommand{\sympos}{S_{++}^n}

\newcommand{\footremember}[2]{%
    \footnote{#2}
    \newcounter{#1}
    \setcounter{#1}{\value{footnote}}%
}

\title{HJ-sampler: A Bayesian sampler for inverse problems of a stochastic process by leveraging Hamilton--Jacobi PDEs and score-based generative models\footremember{equal}{Tingwei Meng and Zongren Zou contributed equally to this work.}}
\author{Tingwei Meng\footremember{ucla}{Department of Mathematics, UCLA, Los Angeles, CA 90025 USA (tingwei@math.ucla.edu).} \and Zongren Zou\footremember{brown}{Division of Applied Mathematics, Brown University, Providence, RI 02912 USA (zongren\_zou@brown.edu).} \and J\'er\^ome Darbon\footremember{brown2}{Corresponding author. Division of Applied Mathematics, Brown University, Providence, RI 02912 USA (jerome\_darbon@brown.edu).} \and George Em Karniadakis\footremember{George}{Division of Applied Mathematics, Brown University, Providence, RI 02912 USA, and Pacific Northwest National Laboratory, Richland, WA 99354 USA (george\_karniadakis@brown.edu).}}

\date{}

\begin{document}
\maketitle

\begin{abstract}
The interplay between stochastic processes and optimal control has been extensively explored in the literature. With the recent surge in the use of diffusion models, stochastic processes have increasingly been applied to sample generation. This paper builds on the log transform, known as the Cole-Hopf transform in Brownian motion contexts, and extends it within a more abstract framework that includes a linear operator. Within this framework, we found that the well-known relationship between the Cole-Hopf transform and optimal transport is a particular instance where the linear operator acts as the infinitesimal generator of a stochastic process. We also introduce a novel scenario where the linear operator is the adjoint of the generator, linking to Bayesian inference under specific initial and terminal conditions. Leveraging this theoretical foundation, we develop a new algorithm, named the HJ-sampler, for Bayesian inference for the inverse problem of a stochastic differential equation with given terminal observations. The HJ-sampler involves two stages: (1) solving the viscous Hamilton-Jacobi partial differential equations, and (2) sampling from the associated stochastic optimal control problem. Our proposed algorithm naturally allows for flexibility in selecting the numerical solver for viscous HJ PDEs. We introduce two variants of the solver: the Riccati-HJ-sampler, based on the Riccati method, and the SGM-HJ-sampler, which utilizes diffusion models. We demonstrate the effectiveness and flexibility of the proposed methods by applying them to solve Bayesian inverse problems involving various stochastic processes and prior distributions, including applications that address model misspecifications and quantifying model uncertainty.
\end{abstract}


\section{Introduction}

Uncertainty Quantification (UQ) plays a vital role in scientific computing, helping to quantify and manage the inherent uncertainties in complex models and simulations \cite{roy2011comprehensive, psaros2023uncertainty}. Within UQ, two significant areas of active research are Bayesian inference and sampling from data distributions. Bayesian inference has garnered considerable interest within the scientific computing community due to its ability to rigorously combine prior information with observational data, addressing model uncertainties and enhancing predictive capabilities \cite{von2011bayesian, liu2001monte, calvetti2007introduction, zou2024neuraluq, zou2023uncertainty, zou2024leveraging}. 
Meanwhile, the second area—sampling from data distributions—has gained popularity in the machine learning and AI communities.
This surge in interest is largely driven by the success of generative models, particularly diffusion models, which excel at generating high-quality samples from complex, high-dimensional distributions~\cite{song2020score,de2021diffusion,zhang2024wasserstein,zhang2023mean,deveney2023closing,yang2023diffusion}. 
Diffusion models, and more generally score-based methods, have been employed not only for data generation but also for solving a wide range of scientific computing problems, including inverse problems~\cite{song2021solving}, sampling~\cite{sohl2015deep,zhang2023diffusion,zhang2024wasserstein,bruna2024posterior}, distribution modification~\cite{wang2024protein}, mean-field problems~\cite{lu2024score,zhou2024deep}, control problems~\cite{wei2024closed,wei2024generative}, forward and inverse partial differential equations (PDEs)~\cite{wu2024uncertainty}, and Schrödinger bridge problems~\cite{chen2021likelihood,shi2024diffusion,somnath2023aligned,hamdouche2023generative}. This broad applicability makes diffusion models a powerful tool in both AI and scientific computing.

In this work, we establish connections between these two subfields of UQ. Specifically, we observe that the Bayesian posterior distribution for certain inverse problems involving stochastic processes can be represented as a PDE, which is further linked to a stochastic optimal control problem. Based on this connection, we design an algorithm that solves certain Bayesian sampling problems by addressing the associated stochastic optimal control problem. This algorithm bears similarities to diffusion models, particularly a class known as Score-based Generative Models (SGMs). By highlighting this connection, our work offers a novel link between Bayesian inference, generative models, and traditional stochastic optimal control, providing new directions for exploration and development.

Theoretically, this connection arises as a special case of the log transform, where the initial condition incorporates the observation, and the terminal condition reflects the prior distribution in the Bayesian inference problem. The log transform has been extensively studied in the literature~\cite{fleming1985stochastic,feng2006large,gao2022revisit,mielke2014relation} and is often referred to as the Cole-Hopf transform~\cite{evans2022partial,leger2019geometric,leger2021hopf} when applied to Brownian motion. This transform connects the nonlinear system, consisting of Hamilton-Jacobi (HJ) equations coupled with Fokker-Planck equations, to its linear counterpart, the Kolmogorov forward and backward equations. Traditionally, the Cole-Hopf transform is used to simplify the solution of nonlinear PDEs by converting them into linear PDEs, which are generally easier to solve. In probability theory, it is also employed as a tool for sampling, known as Doob's \(h\)-transform. Beyond these established uses, we introduce a novel application of the log transform within Bayesian inference, a context that presents some open questions and opportunities for further research (see the discussion in the summary).

From a more detailed and practical perspective, we consider the Bayesian inference problem where the likelihood is determined by \(Y_T | Y_0\), with \(Y_t\) governed by a Stochastic Differential Equation (SDE). The goal is to solve the inverse problem of the stochastic process, specifically inferring the position of \(Y_t\) given an observation at its terminal position \(Y_T\). For this class of problems, we propose an algorithm called the HJ-sampler, which generates sample paths from the posterior distribution by solving the corresponding stochastic optimal control problem. The algorithm consists of two steps: first, computing the control, and second, sampling the optimal trajectories. Different versions of the algorithm arise based on the method used to compute the control. In this paper, we present two such versions: the Riccati-HJ-sampler, which applies the Riccati method, and the SGM-HJ-sampler, which leverages the SGM method. A key advantage of the HJ-sampler is its flexibility. The first step, which computes the control, is independent of the second step, which samples the trajectories. This allows for flexibility in adjusting the observation position \(Y_T\), the time \(T\), and the discretization size without requiring a recomputation of the control. Beyond the proposed algorithm itself, this connection between Bayesian inference and control theory allows us to harness techniques from both control algorithms and diffusion models for efficient Bayesian sampling. Additionally, it offers a potential Bayesian interpretation of diffusion models and opens up avenues for their generalization.

While the log transform has been employed in various sampling algorithms in the literature~\cite{bernton2019schr,zhang2022koopman,zhang2021sampling,hartmann2017variational,chopin2020introduction}, our approach differs in its application. These existing methods, and also other sampling algorithms such as such as Hamiltonian Monte Carlo \cite{neal2012mcmc}, variational inference \cite{blei2017variational, hoffman2013stochastic, ranganath2014black}, and Langevin-type Monte Carlo \cite{welling2011bayesian, girolami2011riemann}, typically include the target distribution as part of the terminal condition, requiring an explicit analytical formula for the density function. In contrast, our approach assumes that only the prior density or prior samples are available, with the likelihood being provided by a stochastic process, which may not have an analytical form. In other words, the terminal condition in our method encodes only the prior information, while the likelihood is embedded within the stochastic process. This decoupling provides the flexibility discussed earlier and ensures that the sampling process directly corresponds to the original stochastic process, allowing us to generate entire sample paths rather than finite-dimensional sample points.

The remainder of the paper is organized as follows. In Section~\ref{sec:log_transform}, we detail the log transform at an abstract level, followed by specific cases in Sections~\ref{sec:log_Xt} and~\ref{sec:log_bayesian}. Section~\ref{sec:log_Xt} connects the log transform to stochastic optimal transport (SOT) and Schr\"odinger bridge problems, while Section~\ref{sec:log_bayesian} focuses on the special case related to Bayesian inference setups, which is the primary contribution of this paper. Based on the content in Section~\ref{sec:log_bayesian}, an algortihm caled the HJ-sampler algorithm is presented in Section~\ref{sec:HJsampler}, and numerical results are provided in Section~\ref{sec:numerics}. Finally, Section~\ref{sec:summary} summarizes the findings, discusses limitations, and outlines future research directions. Additional technical details are provided in the appendix. In Figure~\ref{fig:illustration_roadmap}, we present the roadmap of this paper, highlighting the relationships between the main concepts and methodologies discussed.

\begin{figure}[htbp]
    \includegraphics[width=\textwidth]{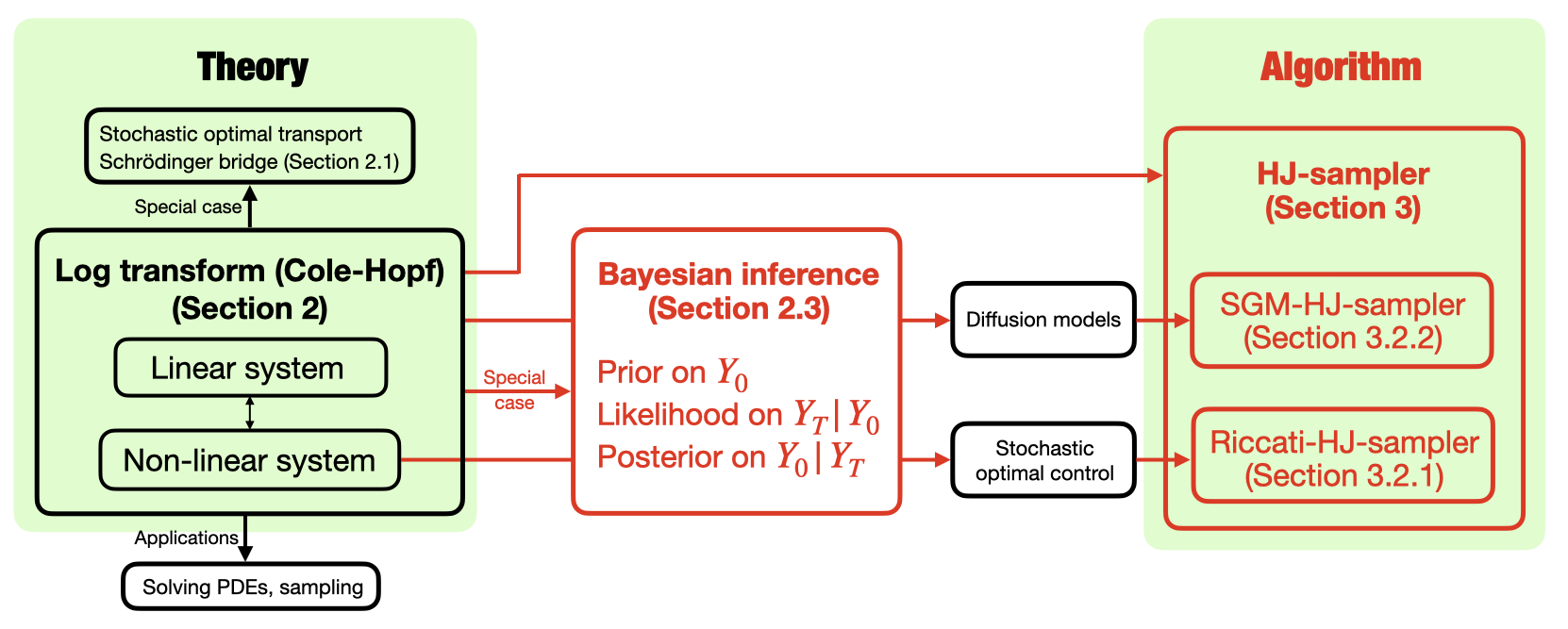}
    \caption{
    Roadmap of this paper. The black sections represent well-known concepts in the literature, while the red sections indicate our contributions.
    } 
    \label{fig:illustration_roadmap}
\end{figure}

\section{The log transform: bridging linear and non-linear systems in the context of stochastic processes}\label{sec:log_transform}

This section introduces the log transform, a mathematical technique that establishes a connection between linear and nonlinear systems. 
We investigate this transform in the context of a versatile linear operator, \(\epsgeneratorfwd\), which may represent differential, difference, integral operators, or combinations thereof.
In Sections~\ref{sec:log_Xt} and~\ref{sec:log_bayesian}, examples are provided where $\epsgeneratorfwd$ functions as either the infinitesimal generator of a stochastic process or its adjoint. The subscript~$\epsilon$ indicates the inclusion of a positive parameter in the operator, reflecting the degree of stochasticity in these examples. 
The operator may also depend on the spatial variable $x$ and the time variable $t$. We put $t$ in the subscript because different equations may involve the operator at different times, while we omit the dependence on $x$ for simplicity of notation.

We introduce two functions, $\mubwd$ and $\mufwd$, that map from $\Rn \times [0, T]$ to $\R$ and comply with the linear system specified below:
\begin{equation}\label{eqt:forward_backward_Kolmogorov}
\begin{dcases}
\partial_t \mubwd = \epsgeneratorinv \mubwd, \\
\partial_t \mufwd = \epsgeneratorfwd^* \mufwd,
\end{dcases}
\end{equation}
where each operator is applied at the point \((x,t)\) for any \(x \in \mathbb{R}^n\) and \(t \in [0,T]\). The notation \(\epsgeneratorfwd^*\) denotes the adjoint of \(\epsgeneratorfwd\) in $L^2(\Rn)$ with respect to the spatial variable \(x\).
The log transform establishes a nonlinear relationship between the pairs $(\mubwd, \mufwd)$ and $(\rho, \Sepsbwd)$ as follows:
\begin{equation}\label{eqt:transform_mu_to_rho}
\rho(x,t) = \mubwd(x,T-t)\mufwd(x,t), \quad \Sepsbwd(x,t) = \epsilon\log \mubwd(x,T-t),
\end{equation}
where the constant \(\epsilon\) in the second equation is the same as the hyperparameter in the operator \(\epsgeneratorfwd\).
The corresponding inverse transformation reads
\begin{equation}
\mubwd(x,t) = e^{\frac{\Sepsbwd(x,T-t)}{\epsilon}}, \quad
\mufwd(x,t) = \rho(x,t) e^{-\frac{\Sepsbwd(x,t)}{\epsilon}}.
\end{equation}
This leads to a coupled nonlinear system for $\rho$ and $\Sepsbwd$:
\begin{equation}\label{eqt:coupled_PDE_fwd}
\begin{dcases}
\partial_t \rho + \rho e^{-\frac{\Sepsbwd}{\epsilon}}\epsgeneratorfwd e^{\frac{\Sepsbwd}{\epsilon}} - e^{\frac{\Sepsbwd}{\epsilon}}\epsgeneratorfwd^* (\rho e^{-\frac{\Sepsbwd}{\epsilon}}) = 0, \\
\partial_t \Sepsbwd + \epsilon e^{-\frac{\Sepsbwd}{\epsilon}}\epsgeneratorfwd e^{\frac{\Sepsbwd}{\epsilon}} = 0,
\end{dcases}
\end{equation}
where each operator is applied at the point \((x,t)\) for any \(x \in \mathbb{R}^n\) and \(t \in [0,T]\).
We refer to the first equation in~\eqref{eqt:coupled_PDE_fwd} as the (generalized) Fokker-Planck equation and the second as the (generalized) HJ equation. As shown in Example~\ref{eg:log_transform_BM} and further discussed in Appendices~\ref{appendix:log_sde} and~\ref{appendix:HJsampler_sde}, when $\epsgeneratorfwd$ corresponds to an SDE, the equations in~\eqref{eqt:coupled_PDE_fwd} reduce to the Fokker-Planck equation and the viscous HJ PDE. In such instances, we refer to the second equation in~\eqref{eqt:coupled_PDE_fwd} as the viscous HJ PDE, but we refrain from using the term ``viscous" for general cases due to the potential absence of a Laplacian term in the equation in such contexts.

This transformation illustrates the connection between linear and nonlinear systems. The log transform, particularly when $\epsgeneratorfwd$ acts as the infinitesimal generator of a stochastic process, is recognized in the literature. For context, we briefly touch upon this aspect in Section~\ref{sec:log_Xt}, offering it as part of the broader narrative. Our novel contribution emerges in Section~\ref{sec:log_bayesian}, where we establish a new link to the Bayesian framework by considering $\epsgeneratorfwd$ as the adjoint of the infinitesimal generator. This insight is crucial for the algorithm we develop later in the paper. By integrating these elements into a comprehensive theoretical framework, we highlight the interconnectedness of diverse applied mathematics fields, such as stochastic processes, control theory, neural networks, and PDEs. This interdisciplinary approach fosters the potential for leveraging algorithms developed in one field to address problems in another, encouraging innovative cross-disciplinary applications.

\subsection{The connection between the log transform and stochastic optimal control}\label{sec:log_Xt}

\begin{figure}[htbp]
    \includegraphics[width=\textwidth]{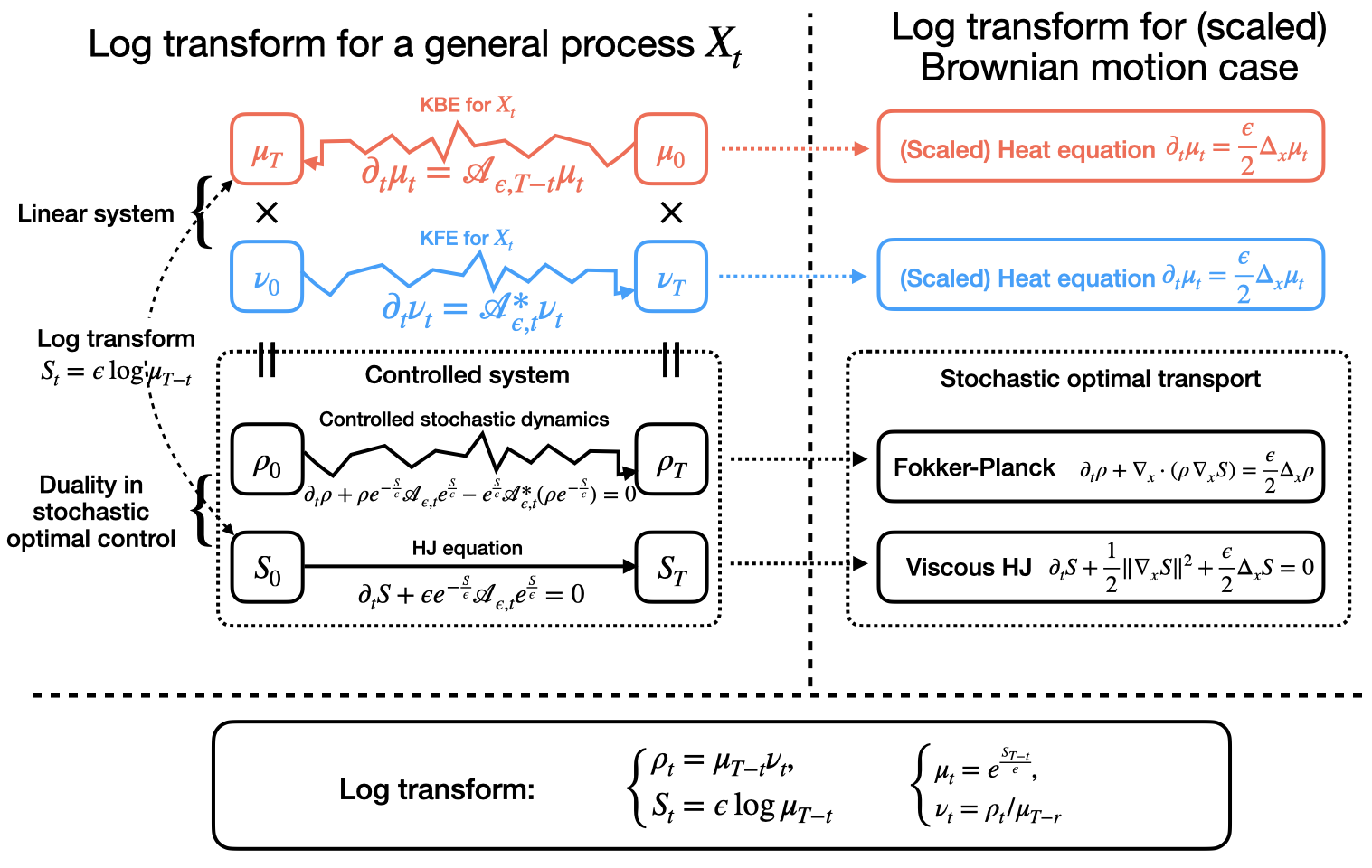}
    \caption{
    This figure illustrates the log transform~\eqref{eqt:transform_mu_to_rho} applied when $\epsgeneratorfwd$ is as the infinitesimal generator of a stochastic process $X_t$. On the left, a general process is depicted, while on the right, the specific instance of a scaled Brownian motion (see Example~\ref{eg:log_transform_BM}) is presented. The time orientation selected here is consistent with that used in stochastic optimal control or stochastic optimal transport problems, aligning with the reversal of the viscous HJ PDE (see Remark~\ref{rem:time_direction} for details).} 
    \label{fig:illustration_fig1}
\end{figure}

In this section, we choose the linear operator $\epsgeneratorfwd$ in~\eqref{eqt:forward_backward_Kolmogorov} and~\eqref{eqt:coupled_PDE_fwd} to be the infinitesimal generator of a stochastic (Feller) process $X_t$ in $\mathbb{R}^n$ from $t=0$ to $t=T$.
This operator incorporates $\epsilon$ to denote the level of stochasticity. Its adjoint is represented by $\epsgeneratorfwd^*$. For further mathematical details on the concepts discussed here, we refer readers to~\cite{liggett2010continuous,pavliotis2014stochastic}.

The linear system in equation~\eqref{eqt:forward_backward_Kolmogorov} encompasses the Kolmogorov backward equation (KBE) for $\mubwd$ and the Kolmogorov forward equation (KFE) for $\mufwd$. 
The non-linear system~\eqref{eqt:coupled_PDE_fwd} constitutes a system for stochastic optimal control, with the source of stochasticity being intrinsically linked to the underlying process $X_t$. Since the connection works for a general stochastic process, the stochastic nature of the controlled system may extend beyond Brownian motion, enabling consideration of jump processes as well.
When $X_t$ adheres to an SDE, the two systems~\eqref{eqt:forward_backward_Kolmogorov} and~\eqref{eqt:coupled_PDE_fwd} become two PDE systems (see Example~\ref{eg:log_transform_BM} and Appendix~\ref{appendix:log_sde}). Furthermore, general stochastic processes could lead to systems characterized by discretized PDEs, integral equations, or integro-differential equations. 
An illustrative example of the scaled Poisson process is provided in Appendix~\ref{appendix:log_Poisson}.
An illustration of a general process, as well as a specific case of a scaled Brownian motion, is presented in Figure~\ref{fig:illustration_fig1}.

Under this setup, the log transform elucidates the relationship between the coupled forward-backward Kolmogorov equations and controlled stochastic systems. This transformation has been extensively studied in the literature (see, for example, \cite{fleming1985stochastic,feng2006large,gao2022revisit,mielke2014relation}). It is also referred to as the Cole-Hopf transform~\cite{evans2022partial} in the context of Brownian motion.
The efficacy of the log transform in numerical methodologies stems from its ability to linearize the nonlinear system~\eqref{eqt:coupled_PDE_fwd}, simplifying the complexity inherent in non-linear dynamics and associated stochastic optimal control challenges.

\begin{rem}
[Initial or terminal conditions] The log transform's effectiveness between equations~\eqref{eqt:forward_backward_Kolmogorov} and~\eqref{eqt:coupled_PDE_fwd} is invariant to the initial or terminal conditions applied, provided these conditions are consistently adapted following the transformation. This principle allows for varied applications, as demonstrated with subsequent examples linking the system to certain mean-field games (MFG), stochastic optimal control, stochastic optimal transport (SOT), and stochastic Wasserstein proximal operators, among other areas. For a general process, it is also related to the Schr\"odinger bridge problem. See the following example for more details.
\end{rem}

\begin{example}[Brownian motion]\label{eg:log_transform_BM}
We consider the scenario where the underlying process is a scaled Brownian motion, denoted by $X_t = \sqrt{\epsilon} W_t$, where $W_t$ represents a Brownian motion in $\mathbb{R}^n$. 
The infinitesimal generator $\epsgeneratorfwd$ and its adjoint operator $\epsgeneratorfwd^*$ are characterized by $\epsgeneratorfwd = \epsgeneratorfwd^* = \frac{\epsilon}{2}\Delta_x$, resulting in both the KBE and KFE in~\eqref{eqt:forward_backward_Kolmogorov} being heat equations:
    \begin{equation}
    \partial_t \mu = \frac{\epsilon}{2}\Delta_x \mu, \quad\quad 
    \partial_t \nu = \frac{\epsilon}{2}\Delta_x \nu.
    \end{equation}
The other non-linear system~\eqref{eqt:coupled_PDE_fwd} manifests as the Fokker-Planck PDE and the viscous HJ PDE (with a distinct sign from the usual case, see Remark~\ref{rem:time_direction}):
\begin{equation}\label{eqt:coupled_PDE_BMcase}
    \begin{split}
    \partial_t \rho + \nabla_x \cdot(\rho \nabla_x \Sepsbwd) = \frac{\epsilon}{2} \Delta_x \rho, \quad\quad \partial_t \Sepsbwd + \frac{1}{2} \|\nabla_x \Sepsbwd\|^2 + \frac{\epsilon}{2}\Delta_x \Sepsbwd = 0.
    \end{split}
    \end{equation}
In this scenario (and generally when the underlying process $X_t$ is described by an SDE, as detailed in Appendix~\ref{appendix:log_sde}), the log transform, known as the Cole-Hopf transform, has been extensively examined in the literature~\cite{evans2022partial,leger2019geometric,leger2021hopf}. For more applications of the Cole-Hopf transform, see~\cite{darbon2021bayesian,osher2023hamilton,heaton2023global,chaudhari2018deep,zhang2024wasserstein}.

By introducing distinct sets of initial or terminal conditions, the coupled PDE system~\eqref{eqt:coupled_PDE_BMcase} can be linked to first-order optimality conditions of various problems, including specific MFG, stochastic optimal control, SOT, or stochastic Wasserstein proximal operator.
    For instance, if we assign the initial condition $\rho_0$ to $\rho$ and the terminal condition $-J$ to $\Sepsbwd$, the coupled PDE system~\eqref{eqt:coupled_PDE_BMcase} is linked to the following MFG:
\begin{equation}
\min\left\{\int_0^T \int_{\Rn}\frac{1}{2}\|v(x,s)\|^2 \rho(x,s) dxds + \int_{\Rn}J(x) \rho(x,T)dx\colon \frac{\partial \rho}{\partial t} + \nabla_x\cdot(v\rho) = \frac{\epsilon}{2}\Delta_x \rho, \,\rho(x,0) = \rho_0(x)\right\},
\end{equation}
which is also associated with the stochastic Wasserstein proximal point of $\mathcal{F}(\mu) = \int_{\mathbb{R}^n}J(x)\mu(x)dx$ at $\rho_0$ (see~\cite{tan2023noise,li2023kernel,han2024tensor}).
Specifically, if $\rho_0$ represents a Dirac mass centered at a point $z_0$, this MFG problem reduces to the following stochastic optimal control problem:
\begin{equation}
\min\left\{ \E\left[\int_0^T \frac{1}{2}\|v_s\|^2 ds + J(Z_T)\right]\colon dZ_s = v_sds + \sqrt{\epsilon} dW_s, Z_0 = z_0\right\},
\end{equation}
whose value equals $-\Sepsbwd(z_0, 0)$.
Alternatively, if we specify both initial and terminal conditions for $\rho$ (denoted by $\rho_0$ and $\rho_T$), the PDE system~\eqref{eqt:coupled_PDE_BMcase} becomes associated with the following SOT problem:
\begin{equation}
\min\left\{ \int_0^T \int_{\Rn}\frac{1}{2}\|v(x,s)\|^2\rho(x,s) dxds \colon \frac{\partial \rho}{\partial t} + \nabla_x\cdot(v\rho) = \frac{\epsilon}{2}\Delta_x \rho,\, \rho(x,0) = \rho_0(x),\, \rho(x,T) = \rho_T(x)\right\}.
\end{equation}
Similar results hold for a general SDE. More details are provided in Appendix~\ref{appendix:log_sde}.

The exploration of these connections can be broadened to encompass scenarios characterized by a wider range of stochastic behaviors. This expansion necessitates the consideration of controlled stochastic processes that incorporate stochastic elements beyond the realm of Brownian motion. 
Furthermore, these connections are intricately linked to the large deviation principle governing the underlying stochastic processes. While the literature has thoroughly examined cases involving Brownian motion from the perspective of the large deviation principle, as detailed in \cite{budhiraja2000variational, budhiraja2011variational, fleming1985stochastic, delarue2020master, chetrite2015variational}, more complex stochastic processes have been explored in \cite{budhiraja2019analysis, fleming1989asymptotic, sheu1985stochastic, gao2022revisit, catuogno2022large, leonard2000large, lynch1987large, de1994large, jakobsen2023master}. Additional research has been directed towards elucidating the connections between these models and gradient flows within certain probabilistic spaces, as seen in \cite{adams2011large, fathi2016gradient, renger2013microscopic, peletier2014variational, jordan1998variational}. However, extending the gradient flow concept to encompass jump processes demands a novel approach to defining geometry within probability spaces, diverging from traditional interpretations based on Wasserstein space, as discussed in \cite{mielke2011gradient, mielke2016generalization, mielke2014relation, chalub2021gradient, erbar2014gradient}. Consequently, establishing a direct correlation between the general logarithmic transformation and gradient flow remains elusive.
\end{example}

\begin{rem}\label{rem:time_direction}
[Regarding the time direction]
The viscous HJ PDE in~\eqref{eqt:coupled_PDE_BMcase} exhibits a different sign in front of the diffusion term compared to the traditional viscous HJ PDE. This discrepancy arises from the direction of time. To maintain consistency with the controlled stochastic process in MFG or SOT, the time direction is reversed. In essence, upon applying time reversal to $\Sepsbwd$ and subsequently taking the negative sign, the function $\tilde S(x,t) = -\Sepsbwd(x,T-t)$ satisfies the traditional viscous HJ PDE:
\begin{equation*}
\partial_t \tilde S(x,t) + \frac{1}{2}\|\nabla_x \tilde S(x,t)\|^2 = \frac{\epsilon}{2}\Delta_x \tilde S(x,t).
\end{equation*}
With this discrepancy in signs, the log transform becomes $\mubwd(x,t) = \exp\left(\frac{\Sepsbwd(x,T-t)}{\epsilon}\right) = \exp\left(-\frac{\tilde S(x,t)}{\epsilon}\right)$, thereby restoring the correct sign in the traditional Cole-Hopf transform applied to $\tilde S$.
\end{rem}

\subsection{The connection between log transform and Bayesian inference}\label{sec:log_bayesian}

\begin{figure}[htbp]
    \includegraphics[width=\textwidth]{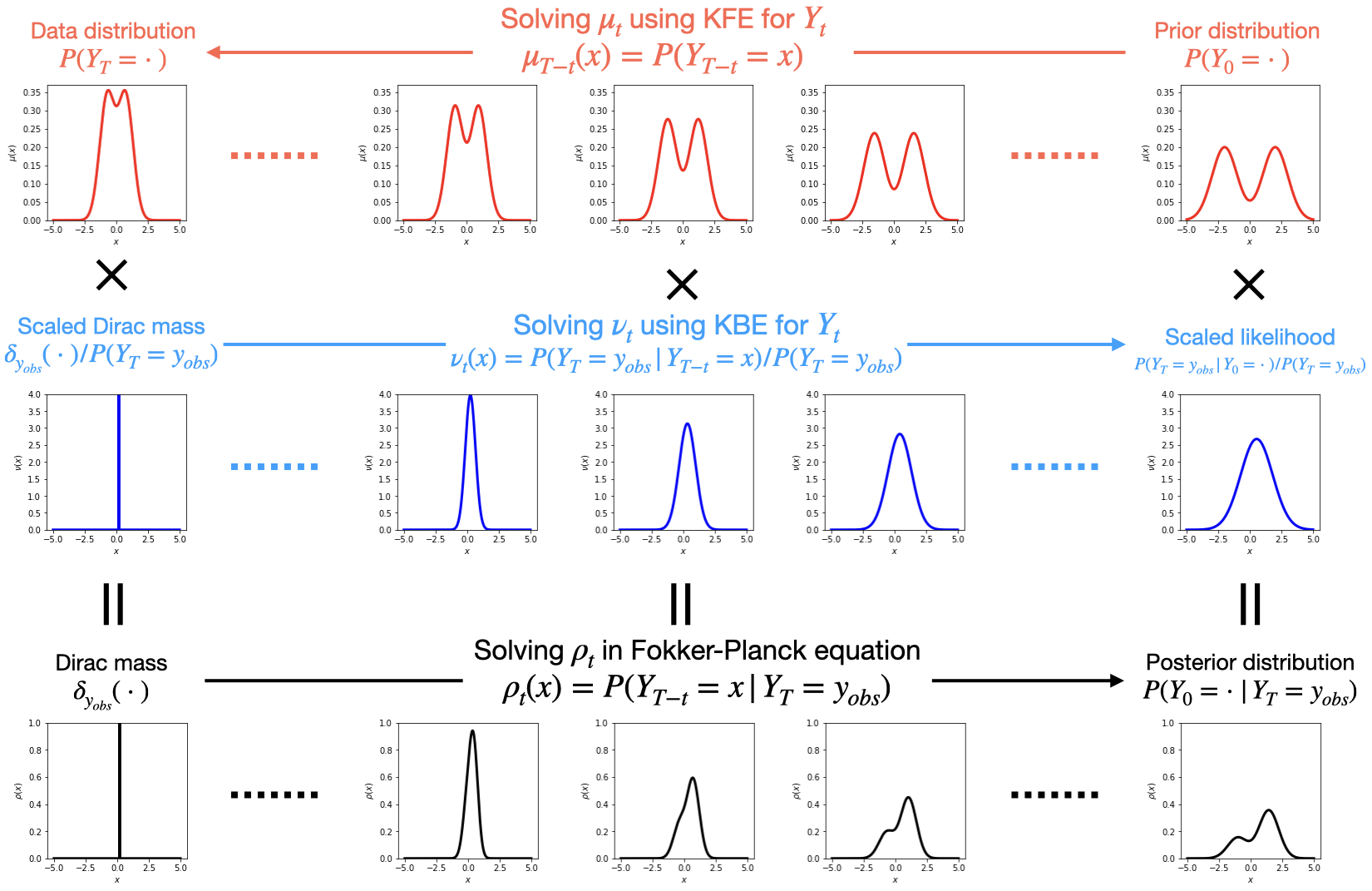}
    \caption{
Depiction of the log transform~\eqref{eqt:transform_mu_to_rho} when the linear operator \(\epsgeneratorinv\) acts as the adjoint of the infinitesimal generator for the stochastic process \(Y_t\), illustrating its application in Bayesian inference. With specific initial and terminal conditions, the function \(\mubwd\) evolves from the prior distribution to the data distribution, while \(\rho\) evolves from a Dirac delta centered at the observation \(\ydata\) of \(Y_T\) to the corresponding posterior distribution. The first line shows the evolution of \(\mubwd\) from right to left, while the second and third lines display the evolutions of \(\mufwd\) and \(\rho\) from left to right. The figures depict the graphs of the respective density functions, and the relationships among the three lines represent the first part of the log transform~\eqref{eqt:transform_mu_to_rho}.
    } \label{fig:illustration_Bayesian1}
\end{figure}

In this section, we delve into the scenario where the linear operator $\epsgeneratorinv$ is the adjoint operator to the infinitesimal generator of an underlying stochastic process \(Y_t\), and we illustrate its relevance to Bayesian inference. To our knowledge, this connection, along with related algorithms for a general process, has yet to be documented in existing literature. The relationship between the Cole-Hopf formula and Bayesian inference has been previously examined in~\cite{darbon2021bayesian}, albeit with a focus on scaled Brownian motion as the underlying process and on computing the posterior mean rather than conducting posterior sampling.

Consider an \(n\)-dimensional stochastic (Feller) process \(Y_t\), with \(\epsgeneratorinv\) representing the adjoint of its infinitesimal generator. In contrast to the previous section, the first equation in~\eqref{eqt:forward_backward_Kolmogorov} describes the KFE, while the second corresponds to the KBE. The initial condition for \(\mubwd\) is defined as the marginal density function of \(Y_0\), and for \(\mufwd\), it is set as the scaled Dirac delta function \(\frac{\delta_{z}(\cdot)}{P(Y_T = z)}\) at a fixed point \(z \in \R^n\). According to the properties of the KFE, the density function \(\mubwd(\cdot, t)\) evolves to match the marginal density of \(Y_t\). For \(\mufwd\), the KBE ensures that \(\mufwd(x,t) = \E[\mufwd(Y_T,0) | Y_{T-t} = x] = \int_{\Rn} \frac{\delta_{z}(y)}{P(Y_T = z)} P(Y_T = y | Y_{T-t} = x) \, dy = \frac{P(Y_T = z | Y_{T-t} = x)}{P(Y_T = z)}\). Through the log transform~\eqref{eqt:transform_mu_to_rho}, the function \(\rho\) is given by
\begin{equation}\label{eqt:rho_bayesian}
\begin{split}
\rho(x,0) &= \mubwd(x,T)\mufwd(x,0) = P(Y_T = x)\frac{\delta_{z}(x)}{P(Y_T = z)} = \delta_{z}(x), \\
\rho(x,t) &= \mubwd(x,T-t)\mufwd(x,t) = P(Y_{T-t} = x)\frac{P(Y_T = z | Y_{T-t} = x)}{P(Y_T = z)} = P(Y_{T-t} = x | Y_T = z),
\end{split}
\end{equation}
indicating that \(\rho(\cdot,t)\) represents the conditional density of \(Y_{T-t}\) given \(Y_T = z\). If \(z\) is the observed value of \(Y_T\), then \(\rho(\cdot, t)\) provides the Bayesian posterior density for \(Y_{T-t} | Y_T = z\).

From a Bayesian perspective, given the initial and terminal conditions, \(\mubwd(\cdot, t)\) evolves from the prior density \(P(Y_0)\) at \(t=0\) to the data density \(P(Y_T)\) at \(t=T\), while \(\rho(\cdot, t)\) evolves from the Dirac delta \(\delta_z\) at \(Y_T\) for \(t=0\) to the posterior density \(P(Y_0 | Y_T = z)\) at \(t=T\). In terms of the nonlinear system for $\rho$ and $\Sepsbwd$, the function \(\rho\) satisfies~\eqref{eqt:rho_bayesian} when the following initial and terminal conditions are applied for \(\rho\) and \(\Sepsbwd\):
\begin{equation}
\rho(x,0) = \delta_z(x), \quad \Sepsbwd(x,T) = \epsilon \log P(Y_0 = x).
\end{equation}
An illustration is provided in Figure~\ref{fig:illustration_Bayesian1}.

Throughout this paper, we assume that the distribution is either continuous or a mixture of continuous and discrete components, allowing us to represent it either as a density function or as a finite combination of Dirac masses. The terms ``distribution" and ``density function" will be used interchangeably as appropriate in the context.

\begin{rem}[Partial observation]
A significant consideration in Bayesian inference involves scenarios of partial observation, relevant in applications such as image inpainting. 
The computations above remain valid even with only a partial observation of \(Y_T\). 
For example, if \(Y_t\) is a concatenation of \(Y_{t,1} \in \R^m\) and \(Y_{t,2} \in \R^{n-m}\), and only an observation \(\ydata \in \R^m\) for \(Y_{T,1}\) is available, the analysis adapts accordingly.
In this remark, for any generic variable \(x\), the notation \(x_1\) with a subscript `$1$' refers to the vector comprising the first \(m\) elements of \(x\), while \(x_2\), denoted with a subscript `$2$', encompasses the remaining \(n-m\) elements. The formulation of the function \(\mubwd\) remains as previously described, but the initial condition for \(\mufwd\) is modified to \(\mufwd(x,0) = \frac{\delta_{\ydata}(x_1)}{P(Y_{T,1} = \ydata)}\) for any vector \(x = (x_1, x_2)\in\R^n\). Following a computation akin to the earlier one, we derive that \(\mufwd(x,t) = \E[\mufwd(Y_T,0) | Y_{T-t} = x] = \int \frac{\delta_{\ydata}(y_1)}{P(Y_{T,1} = \ydata)} P(Y_{T} = y| Y_{T-t} = x) dy = \frac{P(Y_{T,1} = \ydata| Y_{T-t} = x)}{P(Y_{T,1} = \ydata)}\). Consequently, the function \(\rho\)  satisfies
\begin{equation}\label{eqt:rho_Bayesian_formula}
\begin{split}
\rho(x,0) &= \mubwd(x,T)\mufwd(x,0) = P(Y_T = x)\frac{\delta_{\ydata}(x_1)}{P(Y_{T,1} = \ydata)} = \delta_{\ydata}(x_1) P(Y_{T,2} = x_2 | Y_{T,1} = \ydata),
\\
\rho(x,t) &= \mubwd(x,T-t)\mufwd(x,t) = P(Y_{T-t} = x)\frac{P(Y_{T,1} = \ydata| Y_{T-t} = x)}{P(Y_{T,1} = \ydata)} = P(Y_{T-t} = x | Y_{T,1} = \ydata),
\end{split}
\end{equation}
for any $t\in (0,T)$ and $x= (x_1, x_2)\in\Rn$.
While the computation is feasible in this scenario, the primary challenge in implementing the proposed algorithm in Section~\ref{sec:HJsampler} lies in sampling from \(\rho(\cdot, 0)\), that is, determining how to draw samples from the conditional distribution \(P(Y_{T,2} = x_2 | Y_{T,1} = \ydata)\). Advancing the proposed algorithm to address this issue necessitates additional research.
\end{rem}

This section, along with Section~\ref{sec:log_Xt}, presents two distinct examples of the log transform using different instances of the operator \(\epsgeneratorfwd\).
In Section~\ref{sec:log_Xt}, we investigated the log transform's role when \(\epsgeneratorfwd\) acts as the infinitesimal generator of a stochastic process \(X_t\), linking it to certain MFG, SOT, stochastic optimal control, and the stochastic Wasserstein proximal operator -- fields that are at the forefront of current research. In contrast, this section explores the application of the log transform when \(\epsgeneratorinv\) is the adjoint operator of the infinitesimal generator of \(Y_t\), and its relation to Bayesian inference. In scenarios where the infinitesimal generator is self-adjoint (such as in a Brownian motion process), these two situations are the same. 
When considering SDEs, these two cases are intuitively related in a reversed manner, highlighting a compelling research path into this duality and its potential to connect Bayesian inference with MFGs and related areas.

Until now, our discussion has focused on theoretical connections. In the next section, we will utilize these insights to develop a Bayesian sampling algorithm, named the HJ-sampler, which is designed to solve the inverse problem related to the process \(Y_t\) within a Bayesian framework.

\section{A Bayesian sampling method for inverse problems: HJ-sampler}\label{sec:HJsampler}

As explored in Section~\ref{sec:log_bayesian}, the log transform establishes a connection to the Bayesian framework under specific initial and terminal conditions. This section aims to harness this connection by introducing an algorithm, the HJ-sampler, designed for a particular subset of inverse problems.

Initially, we outline the category of problems addressed. Considering an underlying stochastic process \(Y_t\), with a prior distribution \(\Pprior\) on \(Y_0\), our objective is to infer the solution \(Y_t\) for \(t \in [0,T)\) based on the terminal observation \(Y_T = \ydata\). Essentially, we aim to tackle the inverse problem associated with \(Y_t\) within the Bayesian framework. 
Setting the marginal distribution of \(Y_0\) as the prior distribution \(\Pprior\), our task becomes to obtain samples from the posterior distribution \(P(Y_t | Y_T = \ydata)\) for \(t \in [0,T)\).
Starting from the observation \(\ydata\), the HJ-sampler produces a sequence of posterior samples for \(Y_t\), moving backwards in time from \(t=T\) to \(t=0\), through the resolution of the associated stochastic optimal control problem.

Subsequently, the discussion will progress in two phases. Firstly, in Section~\ref{sec:HJsampler_general}, we introduce the HJ-sampler for general stochastic processes, entailing two primary steps: solving \(\Sepsbwd\) and thereafter sampling from \(\rho\) as defined in equation~\eqref{eqt:coupled_PDE_fwd}. Following this, in Section~\ref{sec:HJsampler_sde}, attention shifts to SDE scenarios, delving into the numerical details of each step.

\subsection{HJ-sampler for general stochastic processes}\label{sec:HJsampler_general}

\begin{figure}[htbp]
    \includegraphics[width=\textwidth]{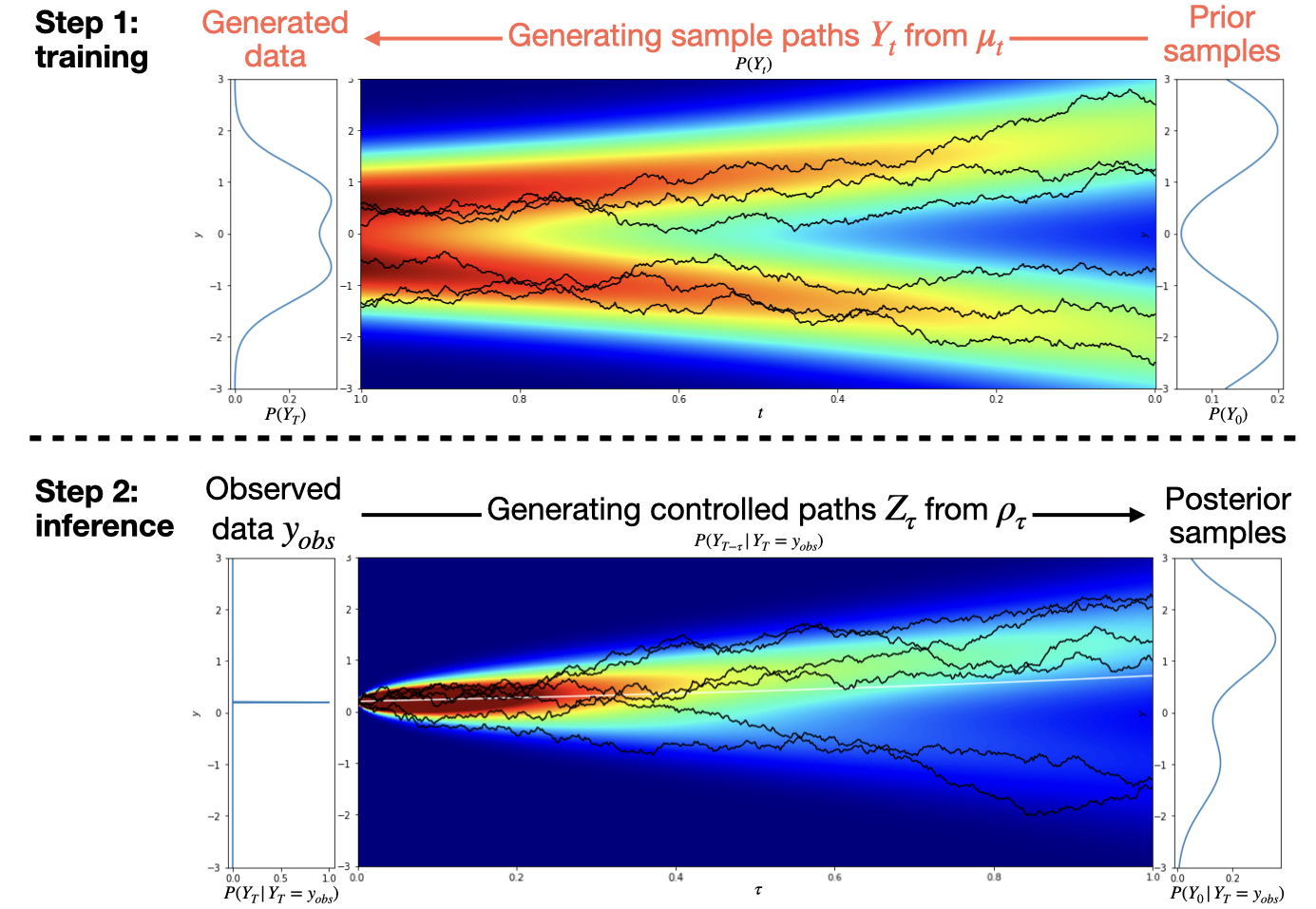}
    \caption{
    The figure illustrates the SGM-HJ-sampler algorithm, consisting of two steps. The first step, shown in the top panel, corresponds to the training phase, where training data is generated by sampling \(Y_t\) from \(\mubwd_t\), and a neural network is trained to approximate the scaled control or score function. The heatmap in the middle represents the evolution of the density function \(\mubwd_t\) from right to left, with time on the horizontal axis and space on the vertical axis. The black curves display the sample paths, demonstrating the training data. The second step, in the bottom panel, represents the inference phase, where posterior samples of \(Y_t \mid Y_T = \ydata\) (with density \(\rho_\tau\)) are generated by sampling the controlled paths \(Z_\tau\). The heatmap shows the evolution of \(\rho_\tau\) from left to right, with the black curves representing the generated sample paths, and the white curve depicting the sample mean of the posterior distribution. The graphs of the initial and terminal densities are displayed on the sides of each panel.
    } 
    \label{fig:illustration_Bayesian2}
\end{figure}

In this section, we present the HJ-sampler, an algorithm designed for a broad class of stochastic processes $Y_t$, where the underlying stochasticity is not limited to traditional Brownian motion dynamics.
As delineated in Section~\ref{sec:log_bayesian}, by setting the terminal condition for $\Sepsbwd$ to $\epsilon\log \Pprior$ and the initial condition for $\rho$ to $\delta_{\ydata}$, we derive that $\rho(\cdot, t)$ represents the conditional distribution of $Y_{T-t}$ given $Y_T=\ydata$. 
This enables the acquisition of posterior samples for $Y_{T-t}$ given $Y_T=\ydata$, utilizing $\rho(\cdot, t)$. 
By sampling from \(\rho(\cdot, t)\) over the interval from \(t = 0\) to \(t = T\), we construct a continuous flow of posterior samples that traces the evolution from \(Y_T\) back to \(Y_0\), conditional on \(Y_T=\ydata\).

Building upon this foundational understanding, we introduce an algorithm named the HJ-sampler, which comprises two primary stages:
\begin{enumerate}
    \item Initially, we solve the HJ equation as specified in the second line of~\eqref{eqt:coupled_PDE_fwd}, which is the following HJ equation with terminal condition:
\begin{equation}\label{eqt:HJsampler_general_HJ}
\begin{dcases}
\partial_t \Sepsbwd + \epsilon e^{-\frac{\Sepsbwd}{\epsilon}}\epsgeneratorfwd e^{\frac{\Sepsbwd}{\epsilon}} = 0,\\
\Sepsbwd(x,T) = \epsilon\log P_{prior}(x).
\end{dcases}
\end{equation}

\item Subsequently, we obtain samples from $\rho$ as delineated in the first line of~\eqref{eqt:coupled_PDE_fwd}, which is expressed as:
\begin{equation}\label{eqt:HJsampler_general_rho}
\begin{dcases}
\partial_t \rho + \rho e^{-\frac{\Sepsbwd}{\epsilon}}\epsgeneratorfwd e^{\frac{\Sepsbwd}{\epsilon}} - e^{\frac{\Sepsbwd}{\epsilon}}\epsgeneratorfwd^* (\rho e^{-\frac{\Sepsbwd}{\epsilon}}) = 0, \\
\rho(x,0) = \delta_{\ydata}(x).
\end{dcases}
\end{equation}
\end{enumerate}

\begin{rem}[Flexibility of the observation time \(T\)]\label{rem:HJ_sampler_flexible_obstime}
The HJ-sampler is designed to sample the posterior distribution of \(Y_t\) given \(Y_T\) for \(t \in [0,T)\). However, it can be generalized to sample the distribution of \(Y_t\) given \(Y_s\) as long as \(0 \leq t < s \leq T\). This flexibility means that the observation time does not need to be fixed at the start. If the observation time \(s\) is initially unknown, the algorithm can be modified by first solving the HJ equation for a sufficiently large \(T\) that is guaranteed to exceed \(s\). In the second step, instead of using the original \(\Sepsbwd\), the time-shifted version \((x,t)\mapsto\Sepsbwd(x, t+T-s)\) is applied to the PDE governing \(\rho\). Sampling from \(\rho\) then provides posterior samples for \(Y_t \mid Y_s = \ydata\). Consequently, if the underlying process and prior distribution remain unchanged, adjusting the observation time \(s\) and location \(\ydata\) does not require re-solving the HJ equation in the first step.
\end{rem}

The HJ-sampler operates through a sequential two-step process. In the first step, prior information is used to set up the terminal condition for \(\Sepsbwd\), after which the function \(\Sepsbwd\) is solved to compute the control for the second step. Since this step does not depend on observational data \(\ydata\), it can be precomputed offline using traditional numerical solvers~\cite{Osher1991High,Jiang2000Weighted,Hu1999Discontinuous,akian2008max,yegorov2021perspectives,mceneaney2006max,Darbon2016Algorithms,kang2017mitigating,kalise2018polynomial,chen2024hopf} or scientific machine learning techniques~\cite{darbon2020overcoming,NakamuraZimmerer2021QRnet,Han2018Solving,bachouch2022deep,wabersich2023data, chen2024leveraging, chen2024leveraging2, zou2024leveraging}. The second step, dependent on the outcomes of the first, uses the observational data to set the initial condition for \(\rho\). This structured separation between prior information and observational data allows for flexibility and precomputation, with the precomputed control applied to generate Bayesian samples once the data is available. In Section~\ref{sec:HJsampler_diffusion}, the application of SGMs for SDEs further illustrates this process, with the first step corresponding to the training phase and the second step to the inference stage.

The main challenge in implementing the HJ-sampler lies in efficiently sampling from \(\rho(\cdot, t)\), which evolves according to the dynamics in~\eqref{eqt:HJsampler_general_rho}. These dynamics relate to stochastic optimal control problems with a terminal cost of \(-\epsilon \log \Pprior\). Optimal control strategies, denoted by \(u^*\), are derived by applying operators to \(\Sepsbwd\), allowing posterior samples to be generated by sampling optimally controlled stochastic trajectories. The type of stochasticity in \(Y_t\) directly influences the nature of the control problem; for example, if \(Y_t\) follows jump process dynamics, the control problem will involve a controlled jump process rather than a traditional controlled SDE. 
In the remainder of this paper, we focus specifically on cases where the underlying process is governed by an SDE, reducing the HJ equation~\eqref{eqt:HJsampler_general_HJ} to a traditional viscous HJ PDE, corresponding to a stochastic optimal control problem. Section~\ref{sec:HJsampler_sde} introduces the details of the HJ-sampler algorithm and its variants for this case, while Section~\ref{sec:numerics} presents the numerical results. Extensions of the HJ-sampler to more general processes are left for future exploration.

\subsection{HJ-sampler for SDEs}\label{sec:HJsampler_sde}
In this section, and throughout the remainder of this paper, we concentrate on stochastic processes described by an \(n\)-dimensional SDE \(dY_t = b(Y_t, t) dt + \sqrt{\epsilon} dW_t\), where \(b: \mathbb{R}^n \times [0, T] \rightarrow \mathbb{R}^n\) acts as the drift component, and \(W_t\) represents Brownian motion in \(\mathbb{R}^n\). While our results are broadly applicable to general SDEs, additional details are provided in Appendix~\ref{appendix:HJsampler_sde}. Here, the operator \(\epsgeneratorinv\) is defined as \(\epsgeneratorinv f = -\nabla_x \cdot (b(x,t)f) + \frac{\epsilon}{2}\Delta_x f\), and the HJ equation~\eqref{eqt:HJsampler_general_HJ} becomes:
\begin{equation}\label{eqt:HJsampler_SDE_HJ}
\begin{dcases}
\partial_t \Sepsbwd - b(x,T-t)\cdot \nabla_x \Sepsbwd(x,t) + \frac{1}{2} \|\nabla_x \Sepsbwd\|^2 + \frac{\epsilon}{2}\Delta_x \Sepsbwd - \epsilon \nabla_x \cdot b(x,T-t)= 0,\\
\Sepsbwd(x,T) = \epsilon\log P_{prior}(x).
\end{dcases}
\end{equation}
The corresponding Fokker-Planck equation~\eqref{eqt:HJsampler_general_rho} evolves into:
\begin{equation}\label{eqt:HJsampler_SDE_rho}
\begin{dcases}
\partial_t \rho + \nabla_x \cdot(\rho (\nabla_x \Sepsbwd - b(x,T-t))) - \frac{\epsilon}{2} \Delta_x \rho =0, \\
\rho(x,0) = \delta_{\ydata}(x),
\end{dcases}
\end{equation}
where the solution \(\rho(\cdot, t)\) is the same as the marginal distribution of an underlying SDE. Consequently, we obtain posterior samples by sampling from this underlying SDE.

In this scenario, our HJ-sampler algorithm comprises two primary steps:
\begin{enumerate}
    \item Numerically solve the HJ PDE~\eqref{eqt:HJsampler_SDE_HJ} to determine \(\Sepsbwd\);
    \item Generate samples from \(\rho\) that satisfies~\eqref{eqt:HJsampler_SDE_rho} by simulating the controlled SDE:
\begin{equation}\label{eq:sampler}
dZ_\tau = (\nabla_x \Sepsbwd(Z_\tau, \tau) - b(Z_\tau, T-\tau)) d\tau + \sqrt{\epsilon} dW_\tau, \quad Z_0 = \ydata.
\end{equation}
For practical implementation, the Euler–Maruyama method~\cite{Kloeden1992Numerical} discretizes the SDE as follows:
\begin{equation}\label{eq:Euler–Maruyama}
Z_{k+1} = Z_k + (\nabla_x \Sepsbwd(Z_{k}, \tau_{k}) - b(Z_k, T-\tau_k)) \Delta \tau + \sqrt{\epsilon \Delta \tau} \xi_k, \quad Z_0 = \ydata,
\end{equation}
where \(\xi_0, \dots, \xi_{\nt-1}\) are independent and identically distributed (i.i.d.) samples from the \(n\)-dimensional standard normal distribution.
\end{enumerate}
Thus, the samples obtained from the discretized SDE \(Z_k\) serve as approximations of the posterior samples for \(Y_{T - k\Delta \tau}\) given \(Y_T = \ydata\). An informal error estimation for the HJ-sampler is provided in Appendix~\ref{appendix:C3_err_est}.

In the HJ-sampler algorithm, the choice of the numerical solver in each step can be adapted based on the practical requirements of the application. 
For instance, in the second step, the Euler–Maruyama method can be replaced with higher-order schemes to achieve higher accuracy.
However, solving the viscous HJ PDE~\eqref{eqt:HJsampler_SDE_HJ} in the first step is more challenging, particularly in high-dimensional settings, making the selection of numerical solvers for this step critical. In lower-dimensional cases, grid-based numerical solvers such as Essentially Non-Oscillatory (ENO) schemes~\cite{Osher1991High}, Weighted ENO schemes~\cite{Jiang2000Weighted}, or the Discontinuous Galerkin method~\cite{Hu1999Discontinuous} can be employed.

In addition to these grid-based methods, this paper introduces two grid-free numerical solvers that have the potential for application to high-dimensional problems:
\begin{itemize}
    \item 
If the Hamiltonian is quadratic and the prior distribution is a Gaussian mixture, the HJ PDEs can be solved using Riccati ODEs (see Section~\ref{sec:HJsampler_riccati}).
\item For more complex cases, we propose neural network-based methods, such as SGMs, as a solution (see Section~\ref{sec:HJsampler_diffusion}).
\end{itemize}

Since the sampling process only requires \(\nabla_x \Sepsbwd\), our focus is on determining \(\nabla_x \Sepsbwd\) rather than solving for \(\Sepsbwd\) itself.

\subsubsection{Riccati-HJ-sampler}\label{sec:HJsampler_riccati}
In scenarios where the prior distribution is Gaussian and the drift term \( b(Y_t, t) \) of the SDE is linear in \( Y_t \), expressed as \( A(t)Y_t + \beta(t) \), where \( A\colon [0,T]\to \R^{n\times n} \) (throughout this paper, \(\R^{m\times n}\) denotes the space of matrices with \(m\) rows and \(n\) columns) and \( \beta\colon [0,T]\to \Rn \) are continuous functions, both the Hamiltonian and the initial condition in~\eqref{eqt:HJsampler_SDE_HJ} adopt quadratic forms. Consequently, the HJ PDE~\eqref{eqt:HJsampler_SDE_HJ} can be efficiently solved using Riccati ODEs~\cite{yong2012stochastic,fleming2012deterministic}.
We apply a numerical ODE solver to solve the corresponding Riccati ODEs and call this version of the HJ-sampler the Riccati-HJ-sampler.

Specifically, if we assume the prior density \(\Pprior\) follows \(\Pprior(\theta) = \frac{1}{\sqrt{(2\pi)^n \det (\Sigma)}} \exp\left(-\frac{1}{2}(\theta - \Gpriorcenter)^T \Sigma^{-1} (\theta - \Gpriorcenter)\right)\), with \(\Gpriorcenter \in \Rn\) and \(\Sigma \in \sympos\) (in this paper, $\sympos$ denotes the set containing all $n \times n$ symmetric positive definite matrices) representing the mean and covariance matrix, respectively, the solution to the viscous HJ PDE~\eqref{eqt:HJsampler_SDE_HJ} is expressed as:
\[
\Sepsbwd(x,t) = -\frac{1}{2} (x-\Sx(T-t))^T\Sxx(T-t)^{-1}(x-\Sx(T-t)) - \Sc(T-t),
\]
where \(\Sxx\colon [0,T] \to \sympos\), \(\Sx\colon [0,T] \to \Rn\), and \(\Sc\colon [0,T] \to \R\) satisfy the Riccati ODE system outlined below:
\begin{equation}\label{eqt:RiccatiODE_sde}
\begin{dcases}
\dot \Sxx(t)= I + \Sxx(t)A(t)^T + A(t) \Sxx(t), \\
\dot \Sx(t) = A(t) \Sx(t) + \beta(t),\\
\dot \Sc(t) = \frac{\epsilon}{2} \Tr(2A(t) + \Sxx(t)^{-1}),
\end{dcases}
\quad\quad \quad\quad 
\begin{dcases}
\Sxx(0) = \frac{1}{\epsilon} \Sigma, \\
\Sx(0) = \Gpriorcenter,\\
\Sc(0) = \frac{n\epsilon}{2}\log(2\pi) + \frac{\epsilon}{2}\log\det(\Sigma).
\end{dcases}
\end{equation}
To facilitate sampling, it is sufficient to compute \(\Sxx\) and \(\Sx\), from which \(\nabla_x \Sepsbwd(x,t)\) can be derived as follows:
\begin{equation}
\nabla_x \Sepsbwd(x,t) = -\Sxx(T-t)^{-1}( x - \Sx(T-t)).
\end{equation}

For Gaussian mixture priors, assume the prior distribution is a convex combination of Gaussian distributions:
\begin{equation}\label{eqt:mixed_Gaussian}
    \Pprior = \sum_{j=1}^{\nprior} w_j \Pprior^j, \text{ where }\Pprior^j(x) = \frac{1}{\sqrt{(2\pi)^n\det(\Sigma_j)}}\exp\left(-\frac{1}{2}(x - \Gpriorcenter_j)^T \Sigma_j^{-1} (x - \Gpriorcenter_j)\right).
\end{equation}
Here, \(w_j\) serves as the mixture coefficient, determining the relative contribution of each Gaussian component in the mixture and satisfying $\sum_{j=1}^\nprior w_j = 1$. Additionally, \(\Gpriorcenter_j \in \Rn\) and \(\Sigma_j \in \sympos\) are the mean vectors and covariance matrices of \(\Pprior^j\), respectively.
Utilizing the log transform, solving the viscous HJ PDE~\eqref{eqt:HJsampler_SDE_HJ} becomes equivalent to solving the KFE with the initial condition set as \(\Pprior\), which results in a linear equation. Consequently, the solution \(\mubwd\) is expressed as
$\mubwd = \sum_{j=1}^N w_j\mubwd_j$,
where \(\mubwd_j\) satisfies the KFE for the initial condition \(\Pprior^j\). Therefore, the solution to the viscous HJ PDE~\eqref{eqt:HJsampler_SDE_HJ} is formulated as:
\begin{equation}
\begin{split}
\Sepsbwd(x,t) &= \epsilon\log \mubwd(x,T-t) = \epsilon\log \left(\sum_{j=1}^Nw_j\mubwd_j(x,T-t)\right) = \epsilon\log \left(\sum_{j=1}^N w_j\exp\left(\frac{\Sepsbwd_j(x,t)}{\epsilon}\right)\right)\\
&= \epsilon\log \left(\sum_{j=1}^N w_j\exp\left(-\frac{1}{2\epsilon} (x-\Sx_{j}(T-t))^T\Sxx_{j}(T-t)^{-1}(x-\Sx_{j}(T-t)) - \frac{\Sc_{j}(T-t)}{\epsilon}\right)\right), 
\end{split}
\end{equation}
where \(\Sepsbwd_j\) addresses the viscous HJ PDE~\eqref{eqt:HJsampler_SDE_HJ} with the terminal condition \(\epsilon\log \mubwd_j(\cdot, 0)\) and can be computed using \(\Sxx_{j}\), \(\Sx_{j}\), and \(\Sc_{j}\) that solve the Riccati ODEs~\eqref{eqt:RiccatiODE_sde} for each \(j\)-th Gaussian prior $\Pprior^j$.
The optimal control is then given by:
\begin{equation}\label{eqt:control_mixedGaussian}
\begin{adjustbox}{width=0.99\textwidth}$
\begin{split}
\nabla_x \Sepsbwd(x,t) = -\frac{\sum_{j=1}^N w_j \Sxx_{j}(T-t)^{-1}(x - \Sx_{j}(T-t)) \exp\left(-\frac{1}{2\epsilon} (x-\Sx_{j}(T-t))^T\Sxx_{j}(T-t)^{-1}(x-\Sx_{j}(T-t)) - \frac{1}{\epsilon}\Sc_{j}(T-t)\right)}{\sum_{j=1}^N w_j\exp\left(-\frac{1}{2\epsilon} (x-\Sx_{j}(T-t))^T\Sxx_{j}(T-t)^{-1}(x-\Sx_{j}(T-t)) - \frac{1}{\epsilon}\Sc_{j}(T-t)\right)}.
\end{split}
$\end{adjustbox}
\end{equation}
In summary, to manage a mixed Gaussian prior \(\Pprior\), the Riccati-HJ-sampler initially involves solving for \(\Sxx_j\), \(\Sx_j\), and \(\Sc_j\) using the Riccati ODEs described in~\eqref{eqt:RiccatiODE_sde} for each \(j\)-th Gaussian component, \(\Pprior^j\). The control \(\nabla_x \Sepsbwd\) is then derived from~\eqref{eqt:control_mixedGaussian}.

\subsubsection{SGM-HJ-sampler}
\label{sec:HJsampler_diffusion}
In many complex scenarios, traditional numerical solvers struggle to efficiently and flexibly solve the viscous HJ PDE~\eqref{eqt:HJsampler_SDE_HJ}, especially in higher dimensions. 
To overcome these limitations, neural network-based methods can be employed. These approaches allow for approximating either the entire solution \(\Sepsbwd\) or its gradient \(\nabla_x \Sepsbwd\), which is essential for determining the control dynamics in the second step of the HJ-sampler.
In this section, we propose an algorithm called the SGM-HJ-sampler, which integrates a diffusion model known as SGM~\cite{song2020score,de2021diffusion,zhang2024wasserstein,zhang2023mean,deveney2023closing} within the HJ-sampler framework. In the literature, SGMs are used to approximate the score function \(\nabla \log P(Y_t=x)\) using a neural network trained on samples of \(Y_t\). Within our framework, this function corresponds to \(\frac{1}{\epsilon}\nabla_x \Sepsbwd(x,T-t)\) according to the log transform. The neural network is then utilized as a scaled control to generate posterior samples. For more insights into diffusion models and their recent applications, we refer to the survey in~\cite{yang2023diffusion}. Additionally, other neural network approaches for approximating \(\Sepsbwd\) or \(\nabla_x \Sepsbwd\) can also be explored in future research.

The SGM approach is structured into two phases: training and inference. During the training phase, the model learns the score function by minimizing a score-matching loss function using data sampled from a stochastic process. In the inference phase, this learned score function is used to reverse the process, starting from an initial ``noise" state to generate samples that reflect the data distribution. These two phases closely correspond to the two steps of the HJ-sampler. The first step of the HJ-sampler is similar to the training stage of the SGM, where the goal is to train a neural network to approximate the score function \(\nabla \log P(Y_t=x) = \frac{1}{\epsilon}\nabla_x \Sepsbwd(x,T-t)\), using data derived from the process \(Y_t\). Similarly, the second step of the SGM-HJ-sampler aligns with the inference phase of the SGM, where samples are generated from a controlled stochastic process utilizing the learned score function. This alignment makes the SGM method well-suited for integration into the HJ-sampler framework.

While our method shares foundational similarities with SGMs, it introduces distinct variations that set it apart. Standard SGMs and other diffusion models typically involve a forward process that transitions from data to noise, followed by a reverse process that reconstructs the data from the noise. In contrast, our model follows a different structure. We utilize two distinct processes: the first moves from the prior distribution \(P(Y_0) = \Pprior\) to the data distribution \(P(Y_T)\), while the second transitions from an observed datum \(Z_0 = \ydata\) to the posterior distribution \(P(Y_0|Y_T=\ydata)\). As a result, unlike conventional diffusion models that use data points for training and prior samples for inference, our model uses prior samples for training and observation points for inference. These similarities and differences may provide a new perspective on diffusion models, enriching both the theoretical and practical understanding of these methods.

The SGM-HJ-sampler utilizing the SGM approach comprises two distinct steps:
\begin{enumerate}
    \item \textbf{Training stage:} Data is generated by sampling \(Y_t\) using the Euler–Maruyama method as follows:
\begin{equation}\label{eqt:training_sample}
     Y_{k+1, j} = Y_{k, j} + b(Y_{k, j}, t_k)\Delta t + \sqrt{\epsilon \Delta t} \xi_{k,j},
    \end{equation}
where \(\xi_{k,j}\) are i.i.d. standard Gaussian, and \(Y_{0,j}\) are prior samples, for \(k = 1,\dots, \nt-1\) and \(j = 1,\dots, \ny\). Here, \(Y_{k,j}\) denotes the \(j\)-th sample for \(Y_{t_k}\). We use \(\nt\) to denote the number of time discretizations and \(\ny\) to denote the number of sample paths in training. A neural network \(s_\NNparam\) (where \(\NNparam\) denotes the trainable parameters) is trained to fit \(\nabla \log P(Y_t)\) using the implicit score-matching function:
\begin{equation}\label{eqt:sgm_loss}
\sum_{k=1}^{\nt} \sum_{j=1}^{\ny} \lambda_k\left(\frac{1}{2}\|s_\NNparam(Y_{k,j},t_k)\|^2 +  \nabla_{x}\cdot s_\NNparam(Y_{k,j}, t_k)\right),
\end{equation}
where \(\lambda_k > 0\) are tunable weighting hyperparameters. 
For high-dimensional problems, it is standard practice to enhance scalability by employing the sliced version~\cite{song2020sliced,song2020score}:
\begin{equation}\label{eqt:slice_sgm_loss}
\sum_{k=1}^{\nt} \sum_{j=1}^{\ny} \lambda_k\left(\frac{1}{2}\|s_\NNparam(Y_{k,j},t_k)\|^2 +  \sum_{\ell=1}^{N_v} v_\ell^T\nabla_{x}(s_\NNparam(Y_{k,j}, t_k)^T v_\ell) \right),
\end{equation}
where \(v_\ell\in \Rn\) are i.i.d. samples from a standard Gaussian distribution. For more details and discussion about the loss function, see Appendix~\ref{appendix:C22_SGM}.
    \item \textbf{Inference stage:} The discretized process \(Z_k\) is sampled according to the equation:
\begin{equation}\label{eqt:sgm_inference}
Z_{k+1} = Z_k + (\epsilon s_\NNparam(Z_{k}, T - \tau_{k}) - b(Z_k, T - \tau_k)) \Delta \tau + \sqrt{\epsilon \Delta \tau} \eta_k, \quad Z_0 = \ydata,
\end{equation}
   where \(\eta_k\) are i.i.d. standard Gaussian samples for \(k=1,\dots,\nz\), distinct from those used in the training stage. This is the Euler–Maruyama discretization for \(Z_\tau\), and \(Z_k\) is a sample for \(Z_{\tau_k}\). Note that we use different notations for the sample size and time grid size in training (\(\ny\) and \(\Delta t\)) and inference (\(\nz\) and \(\Delta \tau\)) to emphasize that these two discretizations do not need to be the same. Note that the inference stage of the SGM-HJ-sampler differs from that of the standard SGM: rather than starting from an initial ``noise" state, the process begins at the observation point \(\ydata\).
\end{enumerate}
This structured approach ensures that our model is both innovative and aligned with established methodologies while offering potential for future exploration. In this method, we do not require an explicit prior probability density function, nor do we obtain the posterior probability density function directly. Instead, the approach relies on prior samples and generates posterior samples.

\begin{table}[h]
    \footnotesize
    \centering
    \begin{tabular}{c|c|c|c|c}
    \hline
    \hline
    & $\mubwd(\cdot, 0)$ & $\mubwd(\cdot, T)$ & $\rho(\cdot, 0)$ & $\rho(\cdot, T)$\\
    \hline 
    HJ-sampler & prior distribution & data distribution & Dirac mass on observation & posterior distribution \\
    \hline
    Diffusion models & data distribution & prior distribution & $\rho(\cdot, 0) = \mubwd(\cdot, T)$ & $\rho(\cdot, T) = \mubwd(\cdot, 0)$\\
    \hline
    \hline
    \end{tabular}
    \caption{
 Difference between HJ-sampler and diffusion models: the initial and terminal conditions for \(\mubwd\) and \(\rho\) have different meanings in the corresponding problems, i.e., posterior sampling for HJ-sampler and data generation for diffusion models.
} 
    \label{tab:difference_HJsampler_diffusion}
\end{table}

\begin{rem}[Difference between SGM-HJ-sampler and SGM]
Although our training and inference stages share similar formulas with diffusion models, the key differences lie in the practical interpretation of the initial and terminal conditions, the specific problems each method is designed to address, the nature of the challenges inherent in these problems, and the directions for future improvements.

Diffusion models or SGMs are designed to generate samples from the underlying distribution of the data. Therefore, their data density function (which corresponds to our \(\mubwd(\cdot,0)\)) is unknown. A stochastic process is chosen to transform the data distribution into a simpler distribution (which corresponds to our \(\mubwd(\cdot,T)\)). This process is manually selected to balance complexity and implementation difficulty, ensuring that \(\mubwd(\cdot,T)\) is easy to sample. In this setup, \(\mubwd(\cdot,T)\) functions like a prior. A large number of samples are drawn from this prior and then transformed using the reverse process to generate samples in the original data distribution. 
The training and inference steps correspond to the forward and reverse processes, enabling the use of ODEs or their solution operators, rather than the reverse SDE, to enhance the efficiency of the sampling stage.

However, in our method, we aim at Bayesian inference rather than sample generation from a distribution described by data. Moreover, the stochastic processes of \(Y_t\) and \(Z_t\) are not in a forward and reverse relationship; they differ by a factor of \(\mufwd\). 
Thus, we have four different initial and terminal densities: \(\mubwd(\cdot, 0)\), \(\mubwd(\cdot, T)\), \(\rho(\cdot, 0)\), and \(\rho(\cdot, T)\), instead of just two. Our interpretation of these four densities also differs. See Table~\ref{tab:difference_HJsampler_diffusion} for more details. 
In our case, the process $Y_t$ is governed by the model of the underlying dynamics, which may be complicated, but we assume that samples from \(Y_t\) can still be obtained in the first step. In terms of the second step, we cannot apply the techniques designed for SGMs to accelerate the inference process, as \(Y_t\) and \(Z_\tau\) are not reverse processes of each other, and the control \(\nabla_x \Sepsbwd\) differs from \(\epsilon \nabla_x \log \rho\). According to the theory connecting SDEs and ODEs, the marginal distribution for \(Z_\tau\) is the same as the marginal distribution of the following ODE:
\begin{equation}
\dot {\tilde Z}_\tau = \epsilon s_\NNparam(\tilde Z_{\tau}, T - \tau) - b(\tilde Z_\tau, T - \tau) - \frac{\epsilon}{2}\nabla_x \log \rho(\tilde Z_\tau, \tau),
\end{equation}
provided that the initial distributions of \(Z_0\) and \(\tilde Z_0\) are the same. However, we cannot use the ODE for \(\tilde Z_\tau\) to accelerate the inference process due to the lack of information on \(\nabla_x \log \rho\). Even if we can generate samples from this ODE, the sample paths of \(\tilde Z_\tau\) differ from those of \(Z_\tau\), and only the marginal distributions match. Consequently, the advantage of trajectory sampling is lost.

While the previous two paragraphs outline the differences between the SGM and SGM-HJ-sampler methods, these distinctions also point to divergent directions for future development. 
In the literature~\cite{song2020improved,song2023consistency,karras2022elucidating,karras2024analyzing}, key directions in the development of diffusion models include finding effective ways to add noise (i.e., SDE design), improving the training process by adjusting loss weights and data generation methods, and accelerating inference processes. The future improvements for the SGM-HJ-sampler are different. Since the underlying SDE process is determined by the problem setup and is usually complicated, we cannot design the SDE or the training data distribution.
Moreover, as discussed earlier, we cannot utilize ODE-based sampling techniques like those in diffusion models. Thus, future work may focus on enhancing the loss design by integrating more model-specific knowledge into the loss function (see~\cite{hu2024score} for example). In the context of Bayesian inference problems, a simpler form of the prior distribution \(\mubwd(\cdot, 0)\) could allow for more efficient incorporation of prior information, providing a basis for further algorithmic extensions.
\end{rem}

\begin{rem}[Theoretical unification of SGM-HJ-sampler and SGM]\label{rem:similarity_SGM}
These two methods can be understood within the same theoretical framework. Through the log transform, the diffusion model is a special case where the function \(\mufwd(\cdot, 0)\) is constantly equal to 1, and \(\mubwd(\cdot, 0)\) represents the data distribution. In contrast, the SGM-HJ-sampler is a special case where \(\mufwd(\cdot, 0)\) is a scaled Dirac delta function \(\frac{\delta_{\ydata}(\cdot)}{P(Y_T = \ydata)}\) at the observation point \(\ydata\), and \(\mubwd(\cdot, 0)\) corresponds to the prior distribution. From this perspective, both methods can be viewed as applications of the log transform, albeit with different interpretations of the initial conditions. A related perspective on SGMs is also explored in~\cite{berner2022optimal}, which interprets diffusion models through the lens of optimal control.

From the perspective of diffusion models, the SGM-HJ-sampler can also be viewed as a conditional sampling variant of SGM. According to diffusion model theory, the following SDE
\begin{equation}
d\tilde Y_{\tau} = (\epsilon s_\NNparam(\tilde Y_{\tau}, T - \tau) - b(\tilde Y_{\tau}, T - \tau))d\tau + \sqrt{\epsilon} dW_\tau, \quad \tilde Y_0 \stackrel{d}{=} Y_T,
\end{equation}
where $\stackrel{d}{=}$ denotes equality in distribution, provides the reverse process of \(Y_t\), meaning \((\tilde Y_0, \tilde Y_\tau)\) is distributionally equivalent to \((Y_T, Y_{T-\tau})\).
If we change the distribution of \(\tilde Y_0\) from that of \(Y_T\) to a Dirac mass $\delta_{\ydata}$ while keeping the SDE unchanged, we obtain the process \(Z_\tau\) in~\eqref{eq:sampler}. Since the SDE does not change, the conditional distribution of \(Z_\tau\) given \(Z_0 = \ydata\) is the same as the distribution of \(\tilde Y_\tau\) given \(\tilde Y_0 = \ydata\), which equals the posterior distribution of \(Y_{T-\tau}\) given \(Y_T = \ydata\). This reasoning underpins the SGM-HJ-sampler from the perspective of diffusion model theory.
\end{rem}

\section{Numerical examples}\label{sec:numerics}

In this section, we demonstrate the accuracy, flexibility, and generalizability of our proposed HJ-sampler by applying it to four test problems. In Section~\ref{sec:numerics_eg1}, we begin with a verification test, using 1D and 2D scaled Brownian motions to present quantitative results that showcase the HJ-sampler's performance. In Section~\ref{sec:numerics_eg2}, we consider the underlying process \(Y_t\) to be an Ornstein–Uhlenbeck (OU) process, demonstrating how the HJ-sampler can handle model uncertainty and misspecification in ODEs~\cite{zou2024correcting}. Specifically, in the model misspecification part of this example, a second-order ODE is incorrectly modeled as a linear ODE. We introduce uncertainty by adding white noise to this misspecified ODE, transforming it into an OU process, and then solve the inference problem using the HJ-sampler. 
The third example, detailed in Section~\ref{sec:numerics_eg3}, illustrates the method's potential in addressing model misspecification in nonlinear ODEs.
Finally, Section~\ref{sec:numerics_eg4} demonstrates the method's scalability by solving a 100-dimensional problem. 

For the first example and the 1D OU process case, we have analytical formulas for the posterior density functions, which serve as ground truths for quantitative comparison. These cases also have analytical solutions to the viscous HJ PDEs~\eqref{eqt:HJsampler_SDE_HJ}. We refer to the version of the HJ-sampler that utilizes this analytical solution as the analytic-HJ-sampler, which serves as a baseline to isolate and distinguish errors from the first and second steps of the HJ-sampler. The analytical formulas and ground truths are provided in Appendix~\ref{appendix:analytical_numeical_examples}. The algorithm's flexibility in sampling from \( P(Y_t | Y_s = \ydata) \) for \( 0 \leq t < s \leq T \) and \( \ydata \in \Rn \) (see Remark~\ref{rem:HJ_sampler_flexible_obstime}) is demonstrated by selecting various \( t \) and \( s \) in several examples. Single precision is used in all numerical experiments for illustration purposes. Additional details on the numerical implementations can be found in Appendix~\ref{sec:details}, with further results in Appendix~\ref{appendix:additional_numerics}. The code for these examples will be made publicly available upon acceptance of this paper.

\subsection{Brownian motion}\label{sec:numerics_eg1}
In this section, we consider scaled 1D and 2D Brownian motions \(dY_t = \sqrt{\epsilon} dW_t\), where \(W_t\) is a standard Brownian motion and \(\epsilon > 0\) is the hyperparameter indicating the level of stochasticity, as the underlying process. We solve the Bayesian inverse problem by sampling from \(P(Y_t \mid Y_s = \ydata)\), where \(0 \leq t < s \leq T\), using the proposed HJ-sampler. The objective is to infer the value of \(Y_t\) from the underlying stochastic model, given the observation \(Y_s = \ydata\) and the prior on \(Y_0\). The posterior distribution for this problem has an analytical solution, serving as the ground truth for error computation and quantitative analysis of the HJ-sampler's performance. Additionally, since the viscous HJ PDE~\eqref{eqt:HJsampler_SDE_HJ} also has an analytical solution, we can compare the performance of the SGM-HJ-sampler and the analytic-HJ-sampler to isolate the errors arising from the first and second steps.

\subsubsection{1D cases}\label{sec:example_1_1}

\begin{table}[h]
    \footnotesize
    \begin{subtable}[h]{\textwidth}
    \centering
    \begin{tabular}{c|c|c|c|c|c|c}
    \hline
    \hline
    & $\ydata=-2$ & $\ydata=-1$  & $\ydata=0$  & $\ydata=1.5$ & $\ydata=3$ & $\ydata \sim P(Y_T)$\\
    \hline 
    analytic-HJ-sampler & $0.0024$ & $0.0023$ & $0.0037$ & $0.0018$ & $0.0018$ & $0.0024\pm0.0006$\\
    \hline
    SGM-HJ-sampler & $0.0054$ & $0.0069$ & $0.0103$ & $0.0175$ & $0.0208$ & $0.0104\pm0.0051$\\
    \hline
    \hline
    \end{tabular}
    \caption{\(W_1\) errors of posterior samples for \(Y_0 \mid  Y_T = \ydata\), computed for different values of \(\ydata\) (\(\Delta\tau = 0.01\)).
    }
    \end{subtable}
    \vfill
    \vfill
    \begin{subtable}[h]{\textwidth}
    \centering
    \begin{tabular}{c|c|c|c|c|c}
    \hline
    \hline
    & Metric & $\Delta \tau = 0.5$  & $\Delta \tau = 0.1$  & $\Delta \tau = 0.01$ & $\Delta \tau = 0.001$ \\
    \hline 
    analytic- & $W_1$ & $0.1141$ & $0.0217$ & $0.0022$ & $0.0008$\\
    HJ-sampler & Wall time (s)  & $0.0$ & $0.2$ & $2.0$ & $28.6$\\
    \hline
    SGM- & $W_1$ & $0.1140$ & $0.0224$ & $0.0053$ & $0.0040$\\
    HJ-sampler & Wall time (s) & $0.3$ & $1.1$ & $10.5$ & $106.8$ \\
    \hline
    \hline
    \end{tabular}
    \caption{$W_1$ errors and the wall time for sampling from $P(Y_0\mid Y_T=-3)$, with different values of $\Delta\tau$ in \eqref{eqt:sgm_inference}.}
    \end{subtable}
    \caption{The quantitative results for the scaled 1D Brownian motion with Gaussian prior. In (a), we compare the performance of the analytic-HJ-sampler and SGM-HJ-sampler on specific values of \(\ydata\) (columns 2–6) and on 1,000 random samples of \(\ydata\) drawn from \(P(Y_T)\) (rightmost column). In (b), we examine the performance and computational wall time of both the analytic-HJ-sampler and SGM-HJ-sampler for different \(\Delta \tau\) values in \eqref{eqt:sgm_inference}. The performance is measured by Wasserstein-1 distances (\(W_1\)) between the samples from the exact posterior distribution \(P(Y_0 \mid Y_T = \ydata)\) (Gaussian) and the posterior samples generated by the proposed algorithm. The \(W_1\) distances are computed using \(1 \times 10^6\) samples, and the neural network in the SGM-HJ-sampler is trained on snapshots taken every \(\Delta t = 0.01\).
    } 
    \label{tab:example_1_1}
\end{table}

In this section, we validate the proposed HJ-sampler through numerical experiments on 1D problems with different prior distributions, including Gaussian, Gaussian mixture, and a mixture of uniform distributions.

We first assume a standard Gaussian prior for \( Y_0 \), where \( Y_0 \sim \mathcal{N}(0, 1^2) \) with \(\epsilon = 1\) and \(T = 1\). For the SGM-HJ-sampler, the interval \([0, T]\) is uniformly discretized with a step size of \(\Delta t = 0.01\) in~\eqref{eqt:training_sample} to generate training data. In the inference stage, we set \(\Delta \tau = 0.01\) in~\eqref{eqt:sgm_inference} to obtain posterior samples by solving the controlled SDE. To quantitatively evaluate the posterior samples, we compute the Wasserstein-1 distance (\(W_1\)) between the exact posterior samples and those obtained using both the analytic-HJ-sampler and SGM-HJ-sampler. The results for specific values of the observation \(\ydata\) for $Y_T$ are presented in Table~\ref{tab:example_1_1}(a) and Figure~\ref{fig:example_1_0}, which demonstrate that both samplers produce high-quality posterior samples, though the analytic-HJ-sampler consistently outperforms the SGM-HJ-sampler. This difference highlights the errors arising from the two steps of the HJ-sampler: the analytic-HJ-sampler only incurs error from discretizing the controlled SDE in the second step, whereas the SGM-HJ-sampler also introduces error from the neural network approximation in the first step. We also assess their performance over 1,000 random samples of \(\ydata\) drawn from \(P(Y_T)\), reporting the mean and standard deviation of \(W_1\) in the rightmost column of Table~\ref{tab:example_1_1}(a). 
Next, we analyze the performance and computational wall time of both HJ-samplers with varying \(\Delta \tau\). Table~\ref{tab:example_1_1}(b) shows the trade-off between accuracy and efficiency in both samplers. Notably, for the SGM-HJ-sampler, the \(W_1\) errors improve at a slower rate when \(\Delta \tau\) is reduced from \(0.01\) to \(0.001\) compared to the analytic-HJ-sampler. This indicates that the error from the neural network, trained with \(\Delta t = 0.01\), becomes the dominant factor, constraining further error reduction despite a smaller \(\Delta \tau\).
Wall times are measured on a standard laptop CPU (13th Gen Intel(R) Core(TM) i9-13900HX, 2.20 GHz, 16 GB RAM).

\begin{figure}[ht!]
    \centering
    \begin{subfigure}[b]{1\textwidth}
        \centering
        \includegraphics[width=1\textwidth]{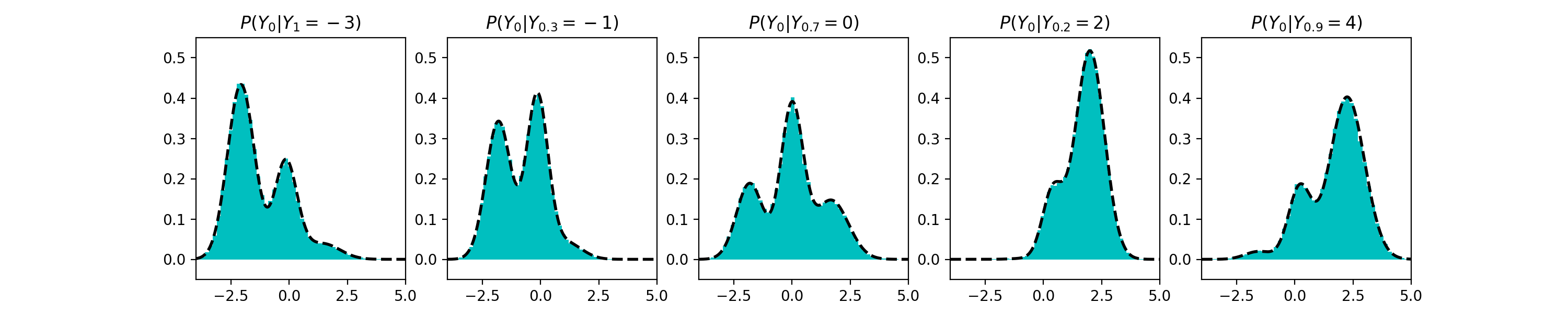}
        \caption{\(P(Y_0 \mid Y_s = \ydata)\) for different \(\ydata \in \mathbb{R}\) and \(s \in (0, T]\).
        }
    \end{subfigure}
    \begin{subfigure}[b]{1\textwidth}
        \centering
        \includegraphics[width=1\textwidth]{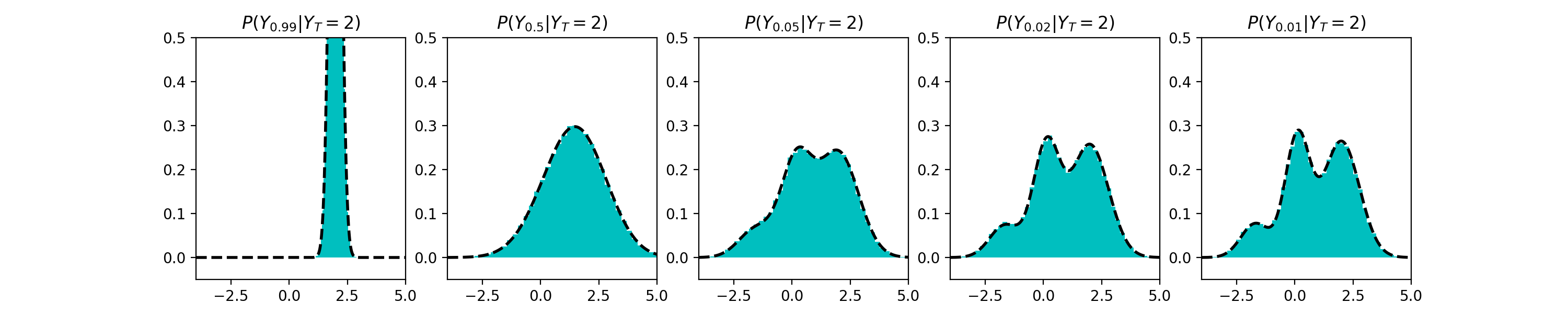}
        \caption{$P(Y_t\mid Y_s = \ydata)$ for different $t\in[0, T)$ with fixed observation $\ydata=2$ and time $s=T$.}
    \end{subfigure}
    \begin{subfigure}[b]{1\textwidth}
        \centering
        \includegraphics[width=1\textwidth]{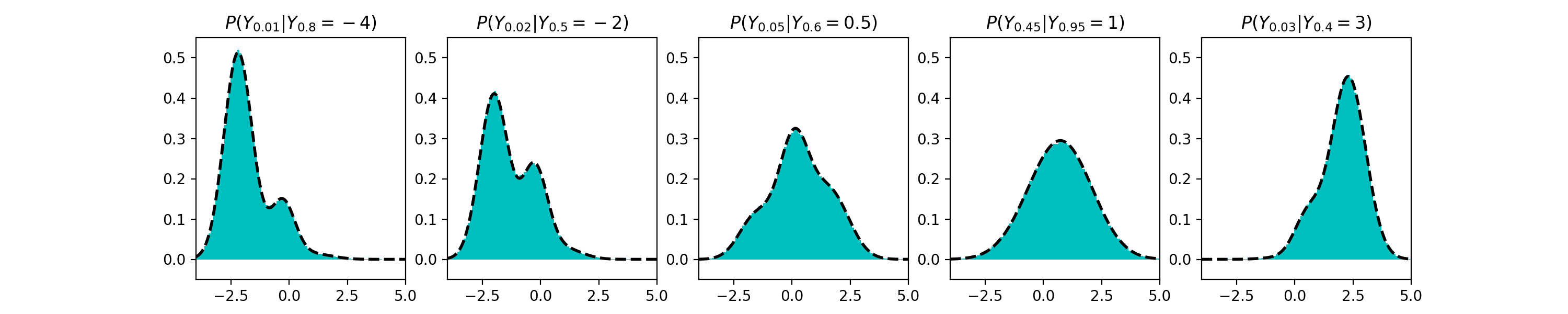}
        \caption{$P(Y_t\mid Y_s=\ydata)$ for different $\ydata\in \R$ and $0\leq t < s \leq T$.}
    \end{subfigure}
    \caption{
    Histograms depicting the distribution of posterior samples for the scaled 1D Brownian motion case with a Gaussian mixture prior, across different observation times \(s\) and data values \(\ydata\). For all cases, the posterior samples, obtained from the SGM-HJ-sampler, utilize the same pretrained neural network, trained on \(t \in [0, T]\) with \(T = 1\). The \textbf{black} dashed lines represent the exact posterior density functions (Gaussian mixture). Each histogram is generated from \(1 \times 10^6\) samples.
    } 
    \label{fig:example_1_case2}
\end{figure}

\begin{table}[h]
    \footnotesize
    \centering
    \begin{tabular}{c|c|c|c|c|c}
    \hline
    \hline
    & $Y_{0.01}\mid Y_{0.8}=-4$  & $Y_{0.02}\mid Y_{0.5}=-2$  & $Y_{0.05}\mid Y_{0.6}=0.5$ & $Y_{0.45}\mid Y_{0.95}=1$ & $Y_{0.03}\mid Y_{0.4}=3$\\
    \hline 
     analytic-HJ-sampler & $0.0019$ & $0.0029$ & $0.0034$ & $0.0026$ & $0.0027$ \\
    \hline
     SGM-HJ-sampler & $0.0098$ & $0.0072$ & $0.0069$ & $0.0110$ & $0.0079$\\
    \hline
    \hline
    \end{tabular}
    \caption{$W_1$ error for the scaled 1D Brownian motion case with a Gaussian mixture prior. The $W_1$ distances are calculated between $1 \times 10^6$ posterior samples of $Y_t \mid Y_s = \ydata$ obtained using two versions of the HJ-sampler (with $\Delta \tau = 0.001$) and $1 \times 10^6$ samples from the exact posterior distribution (Gaussian mixture).} 
    \label{tab:example_1_case2}
\end{table}

\begin{figure}[h]
    \centering
    \begin{subfigure}[b]{1\textwidth}
         \centering
         \includegraphics[width=\textwidth]{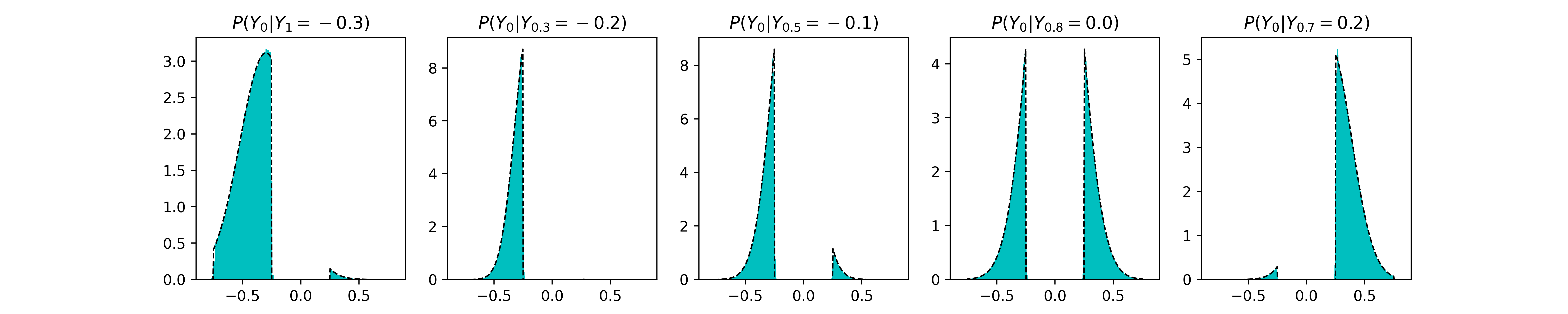}
         \caption{The analytic-HJ-sampler.}
     \end{subfigure}
     \begin{subfigure}[b]{1\textwidth}
         \centering
         \includegraphics[width=\textwidth]{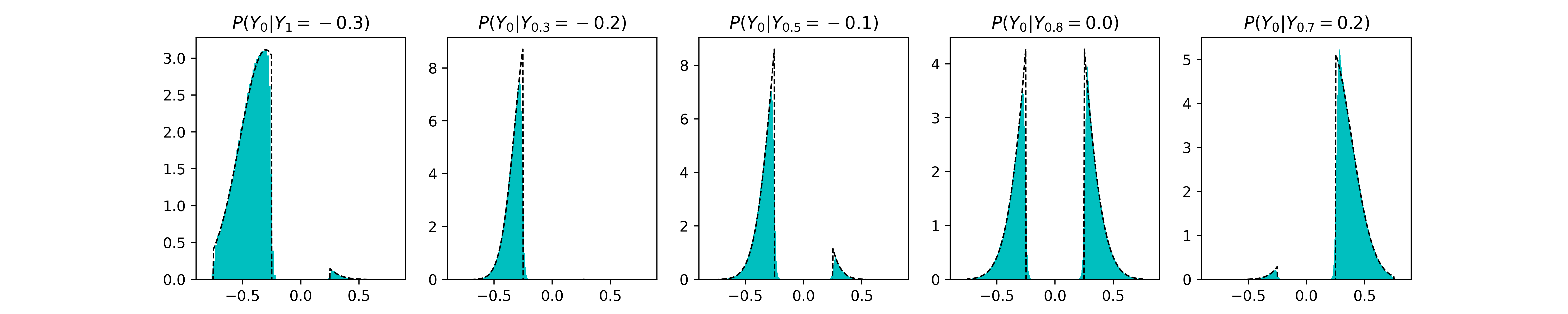}
         \caption{The SGM-HJ-sampler.}
     \end{subfigure}
    \caption{Histograms depicting the distribution of posterior samples of \(Y_0 \mid Y_s = \ydata\) for the scaled 1D Brownian motion case with a prior distribution consisting of a mixture of uniform distributions, across different observation times \(s\) and data values \(\ydata\). In all cases, the posterior samples, obtained using the SGM-HJ-sampler, utilize the same pretrained neural network, trained on \(t \in [0, T]\) with \(T = 1\). The \textbf{black} dashed lines represent the exact posterior density functions. Each histogram is generated from \(1 \times 10^6\) samples. The prior is a mixture of two uniform distributions, \(\mathcal{U}[-0.75, -0.25)\) and \(\mathcal{U}[0.25, 0.75)\), with equal weights.}
    \label{fig:example_1_case3}
\end{figure}

We then apply the HJ-sampler to a 1D Gaussian mixture prior. Specifically, the prior of \(Y_0\) is a mixture of three Gaussians, \(\mathcal{N}(0, 0.5^2)\), \(\mathcal{N}(-2, 0.8^2)\), and \(\mathcal{N}(2, 0.6^2)\), with equal weights, and we set \(\epsilon = 1\), \(T = 1\), and \(\Delta \tau = 0.001\). We demonstrate the flexibility of the SGM-HJ-sampler by generating posterior samples for the following cases:
\begin{enumerate}[(a)]
    \item \(P(Y_0 \mid Y_s = \ydata)\) for \(\ydata \in \R\) and \(s \in (0, T]\),
    \item \(P(Y_t \mid Y_s = \ydata)\) for \(t \in [0, T)\), $s = T$, and $\ydata = 2$,
    \item \(P(Y_t \mid Y_s = \ydata)\) for \(\ydata \in \R\) and \(0 \leq t < s \leq T\).
\end{enumerate}
The results, shown in Figure~\ref{fig:example_1_case2}(a), (b), and (c), indicate that the posterior samples from the SGM-HJ-sampler agree with the exact posterior density functions. Table~\ref{tab:example_1_case2} provides quantitative results for case (c). Notably, in all cases, the neural network is trained once before knowing the observation \(\ydata\) and time \(s\), and after receiving the data, the second step of the HJ-sampler generates posterior samples without needing re-training. This flexibility is discussed in Remark~\ref{rem:HJ_sampler_flexible_obstime}.

Finally, we test the method on a more challenging case where the prior distribution is compactly supported. We consider a prior consisting of a mixture of two uniform distributions, \(\mathcal{U}[-0.75, -0.25)\) and \(\mathcal{U}[0.25, 0.75)\), with equal weights, and set \(\epsilon = 0.05\), \(T = 1\), and \(\Delta \tau = 0.001\). Using both the analytic-HJ-sampler and SGM-HJ-sampler, we generate posterior samples of \(Y_0\) given different observation points \(\ydata\) for various observation times \(s \in (0, T]\). Results are presented in Figure~\ref{fig:example_1_case3}, showing that both HJ-samplers are capable of producing reliable posterior samples that align closely with the exact posterior density functions.
However, the analytic-HJ-sampler outperforms the SGM-HJ-sampler due to the latter's neural network approximation error.

\subsubsection{A 2D case}

\begin{figure}[ht!]
    \centering
    \begin{subfigure}[b]{1\textwidth}
         \centering
         \includegraphics[width=0.2\textwidth]{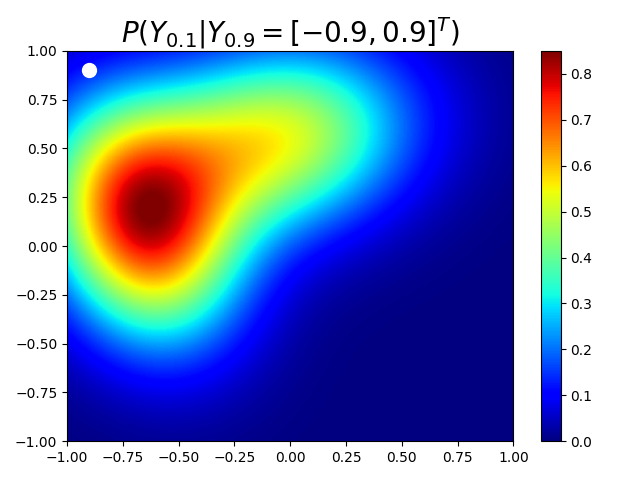}
         \includegraphics[width=0.2\textwidth]{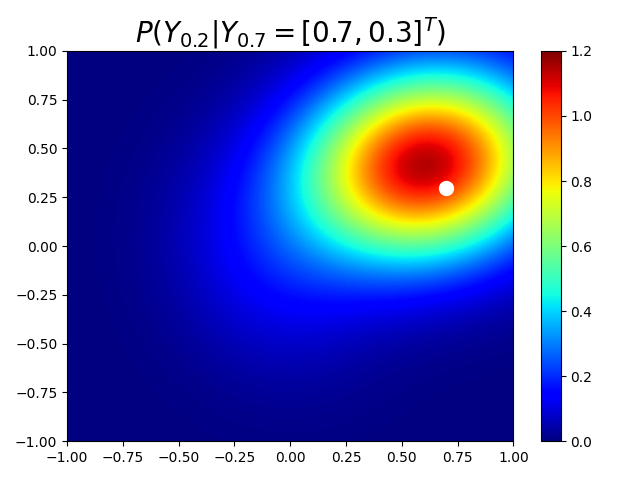}
         \includegraphics[width=0.2\textwidth]{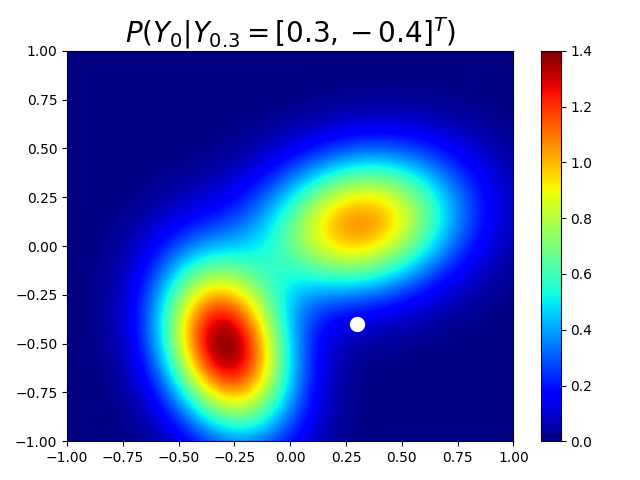}
         \includegraphics[width=0.2\textwidth]{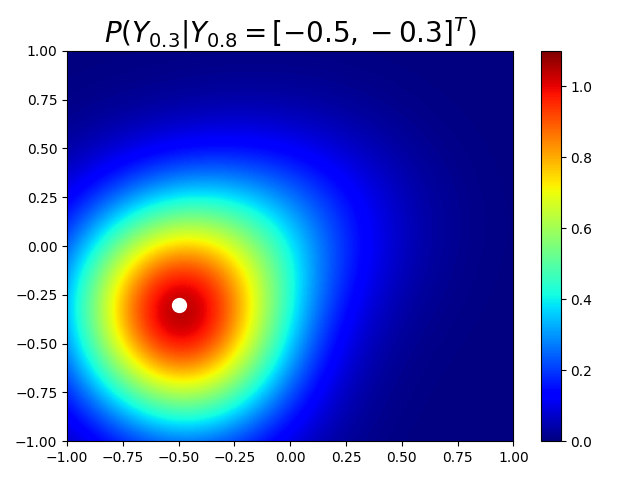}
         \caption{The exact posterior distribution.}
    \end{subfigure}
    \begin{subfigure}[b]{1\textwidth}
         \centering
         \includegraphics[width=0.2\textwidth]{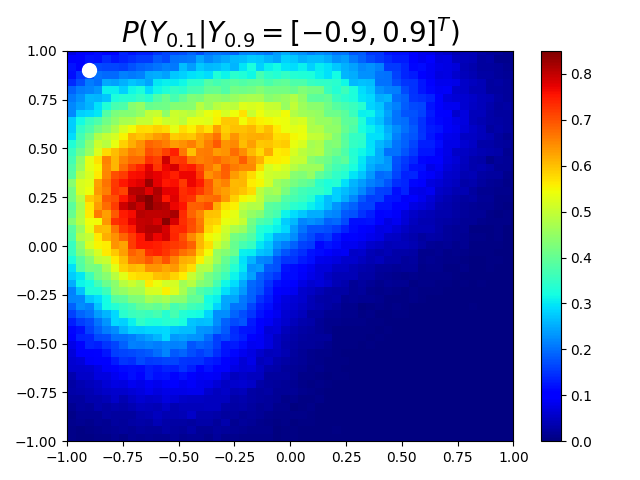}
         \includegraphics[width=0.2\textwidth]{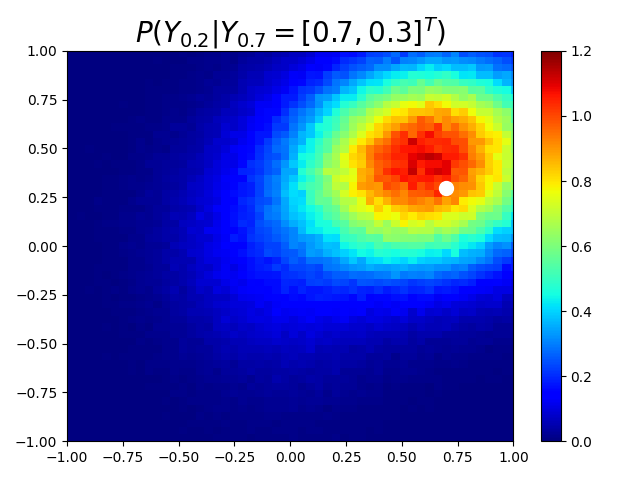}
         \includegraphics[width=0.2\textwidth]{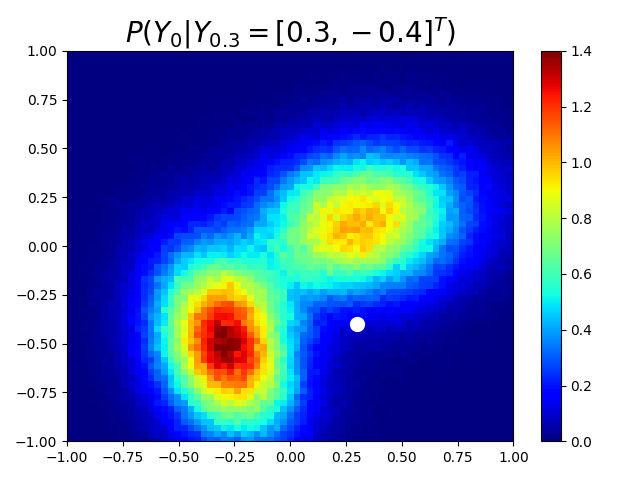}
         \includegraphics[width=0.2\textwidth]{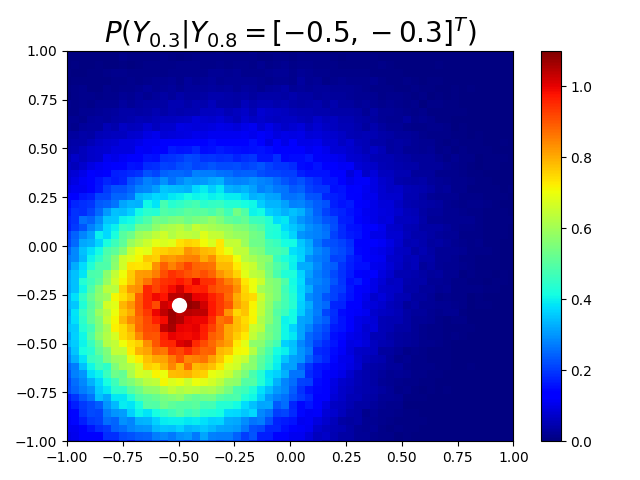}
         \caption{Posterior samples obtained from the SGM-HJ-sampler.}
    \end{subfigure}
    \caption{
 Heatmaps of the scaled 2D Brownian motion case with a Gaussian mixture prior. Posterior samples of \(Y_t\) given the value of \(Y_s\) for \(0 \leq t < s \leq T\) are compared to the exact posterior distribution (Gaussian mixture). In (a), the exact posterior density functions are displayed for specific values of \(t\), \(s\), and \(Y_s\). In (b), the corresponding posterior samples obtained from the SGM-HJ-sampler are presented as histograms. The neural network in the SGM-HJ-sampler was trained on \(t \in [0, T]\), with \(T = 1\).
    } 
    \label{fig:example_1_case4_2}
\end{figure}

\begin{table}[h]
    \footnotesize
    \centering
    \begin{tabular}{c|c|c|c|c}
    \hline
    \hline
    & $Y_{0.1}\mid Y_{0.9}=[-0.9, 0.9]^T$  & $Y_{0.2}\mid Y_{0.7}=[0.7, 0.3]^T$  & $Y_0\mid Y_{0.3}=[0.3, -0.4]^T$ & $Y_{0.3}\mid Y_{0.8}=[-0.5, 0.3]^T$\\
    \hline 
    analytic-HJ-sampler & $0.0007$ & $0.0007$ & $0.0008$ & $0.0007$ \\
    \hline
    SGM-HJ-sampler & $0.0126$ & $0.0056$ & $0.0025$ & $0.0031$ \\
    \hline
    \hline
    \end{tabular}
    \caption{Sliced \(W_1\) errors for the scaled 2D Brownian motion case with a Gaussian mixture prior. For each value of \(t\), \(s\), and \(\ydata\), the \(W_1\) error is computed between \(1 \times 10^6\) samples from the exact posterior distribution \(P(Y_t \mid Y_s = \ydata)\) (Gaussian mixture) and the posterior samples generated using the proposed methods.
    } 
    \label{tab:example_1_case4}
\end{table}

We now consider a 2D case with the prior for \(Y_0\) given as a mixture of two Gaussian distributions with means \(\mu_1 = [0.5, 0.5]^T\) and \(\mu_2 = [-0.5, -0.5]^T\), and covariance matrices \(\Sigma_1 = \begin{bmatrix} 0.25 & 0.05 \\ 0.05 & 1/9 \end{bmatrix}\) and \(\Sigma_2 = \begin{bmatrix} 0.0625 & -0.05\\ -0.05& 0.25 \end{bmatrix}\), respectively. We set \(\epsilon=0.5\) and \(T=1\), and uniformly discretize \([0, T]\) with a step size \(\Delta t = 0.01\) to generate training data. For the inference, we set \(\Delta \tau = 0.001\) when solving the controlled SDE. Posterior samples of \(Y_t\) for different values of \(Y_s\) (where \(0 \leq t < s \leq T\)) are generated using both the analytic-HJ-sampler and the SGM-HJ-sampler. The results from the SGM-HJ-sampler are displayed in Figure~\ref{fig:example_1_case4_2}, which show strong agreement between the posterior samples and the exact density functions. To quantitatively assess the sample quality, we compute the sliced $W_1$ distance between the posterior samples and the exact distributions. The results, shown in Table~\ref{tab:example_1_case4}, indicate that while both methods perform well, the analytic-HJ-sampler generally yields better results. The sliced $W_1$ distance between two \(n\)-dimensional distributions \(\mu\) and \(\nu\) is computed as:
\[
\E_{d \sim \mathcal{U}(\mathbb{S}^{n-1})} \left[ W_1 \left( (\mathcal{P}_d)_\#\mu, (\mathcal{P}_d)_\#\nu \right) \right],
\]
where \(\mathcal{U}(\mathbb{S}^{n-1})\) is the uniform distribution on the unit sphere, \(\mathcal{P}_d\) is the projection along direction \(d\) (i.e., \(\mathcal{P}_d(x) = \langle d, x \rangle\)), and \((\mathcal{P}_d)_\# \mu\) and \((\mathcal{P}_d)_\# \nu\) denote the push-forwards of the distributions \(\mu\) and \(\nu\), respectively. The $W_1$ distance, \(W_1((\mathcal{P}_d)_\# \mu, (\mathcal{P}_d)_\# \nu)\), is computed between samples of \((\mathcal{P}_d)_\# \mu\) and \((\mathcal{P}_d)_\# \nu\). The sliced $W_1$ distance is then calculated by taking the expectation over random directions \(d\) sampled from \(\mathcal{U}(\mathbb{S}^{n-1})\). To compute this sliced $W_1$ distance, we use a Monte Carlo approximation with 50 samples for \(d\), utilizing the Python Optimal Transport library \cite{flamary2021pot}.

\subsection{Ornstein–Uhlenbeck process}\label{sec:numerics_eg2}

In this section, we consider the OU process
\begin{equation}\label{eqt:OU_sec4.2}
dY_t = -BY_t\,dt + \sqrt{\epsilon}\,dW_t,    
\end{equation}
where \(B \in \R^{n \times n}\) is a constant matrix whose eigenvalues have positive real parts, and \(W_t\) is a Brownian motion in \(\R^n\). The prior distribution is assumed to be either Gaussian or a Gaussian mixture. This setup is a specific instance of the process discussed in Section~\ref{sec:HJsampler_riccati}, where \(A(t) = -B\) and \(\beta(t) = 0\). A more general version of the OU process, such as \(dY_t = (c - BY_t)\,dt + \sqrt{\epsilon}\,\sigma\,dW_t\), involving an additional constant vector \(c \in \R^n\) and a constant matrix \(\sigma \in \R^{n \times n}\), can also be solved in a similar manner (as a special case discussed in Appendix~\ref{appendx:numerical_HJsampler}). However, we focus on the simpler case here for illustrative purposes. Both the Riccati-HJ-sampler and SGM-HJ-sampler can be applied to solve this problem. Additionally, in the 1D case, the viscous HJ PDE~\eqref{eqt:HJsampler_SDE_HJ} has an analytical solution, enabling the use of the analytic-HJ-sampler for comparison as a sanity check.

In this scenario, the functions \(\Sx\) and \(\Sc\) in the Riccati ODE~\eqref{eqt:RiccatiODE_sde} have analytical solutions, given by \(\Sx(t) = e^{-tB} \Gpriorcenter\) and \(\Sc(t) = \frac{\epsilon}{2} \log ((2\pi\epsilon)^n \det(\Sxx(t)))\). The other function, \(\Sxx\), can be solved using an ODE solver. Although the Riccati ODE method for solving the marginal density function of the OU process is known in the literature (e.g.,~\cite{kolokoltsov2010nonlinear}), we extend this by connecting it to the viscous HJ PDE and control problems via the log transform.

\begin{table}[h]
    \footnotesize
    \centering
    \begin{tabular}{c|c|c}
    \hline
    \hline
    analytic-HJ-sampler &  Riccati-HJ-sampler &  SGM-HJ-sampler \\
    \hline 
    $0.0085\pm0.0014$ & $0.0086\pm0.0013$ & $0.0103\pm0.0024$ \\
    \hline
    \hline
    \end{tabular}
    \caption{Verification of the proposed methods for a 1D OU process with a Gaussian prior. \(W_1\) distances are computed between the posterior samples of \(Y_0 \mid Y_T = \ydata\) obtained from three HJ-samplers and the samples from the exact posterior distribution (Gaussian). We randomly draw $1,000$ samples of \(\ydata\) from \(P(Y_T)\) (Gaussian) and present the mean and standard deviation of the \(W_1\) distances across these values of \(\ydata\). The \(W_1\) distance is evaluated based on \(1 \times 10^6\) samples. 
    } 
    \label{tab:example_2}
\end{table}

We first perform a sanity check for the presented algorithm by inferring the value of \(Y_0\) given \(Y_T = \ydata\) for a 1D OU process, \(dY_t = -\alpha Y_t dt + \sqrt{\epsilon}dW_t\), with \(\alpha = 3\) and \(\epsilon = 1.5\). The prior distribution is assumed to be the standard Gaussian distribution \(\mathcal{N}(0, 1^2)\), and we set \(T = 1\). We apply the analytic-HJ-sampler, Riccati-HJ-sampler, and SGM-HJ-sampler to obtain posterior samples of \(Y_0\). The forward Euler scheme with a step size of \(1 \times 10^{-4}\) is used to solve the Riccati ODE in the Riccati-HJ-sampler, while \(\Delta\tau = 0.01\) is employed in the inference stage of all three HJ-samplers to solve the controlled SDE. We sample 1,000 values of \(\ydata\) randomly from \(P(Y_T)\) and compute the \(W_1\) distance between the posterior samples generated by the HJ-samplers and the exact posterior distribution of \(Y_0 \mid Y_T = \ydata\) (Gaussian). The mean and standard deviation of the \(W_1\) distances across different values of \(\ydata\) are presented in Table~\ref{tab:example_2}, showing that all three HJ-samplers perform well in producing posterior samples.

After this sanity check, in the following two sections, we demonstrate the ability of the proposed HJ-samplers to handle model uncertainty and misspecification problems. In the first case (Section~\ref{sec:eg_OU_caseA}), we examine a first-order linear ODE system with model uncertainty present in both equations. In the second case (Section~\ref{sec:eg_OU_caseB}), we address a model misspecification problem where a second-order ODE is incorrectly modeled as a second-order linear ODE. This misspecification is captured by the corresponding linear SDE. By reformulating the SDE as a first-order linear system, only the second equation contains the white noise term, illustrating the algorithm’s capability to handle partial uncertainty within a system. In both cases, the SDE models yield an OU process, allowing us to apply the HJ-samplers developed for OU processes to effectively solve these Bayesian inference problems.

\subsubsection{The first case: model uncertainty in a first-order linear ODE system} \label{sec:eg_OU_caseA}

\begin{figure}[ht!]
    \centering
    \begin{subfigure}[b]{1\textwidth}
         \centering
         \includegraphics[width=0.2\textwidth]{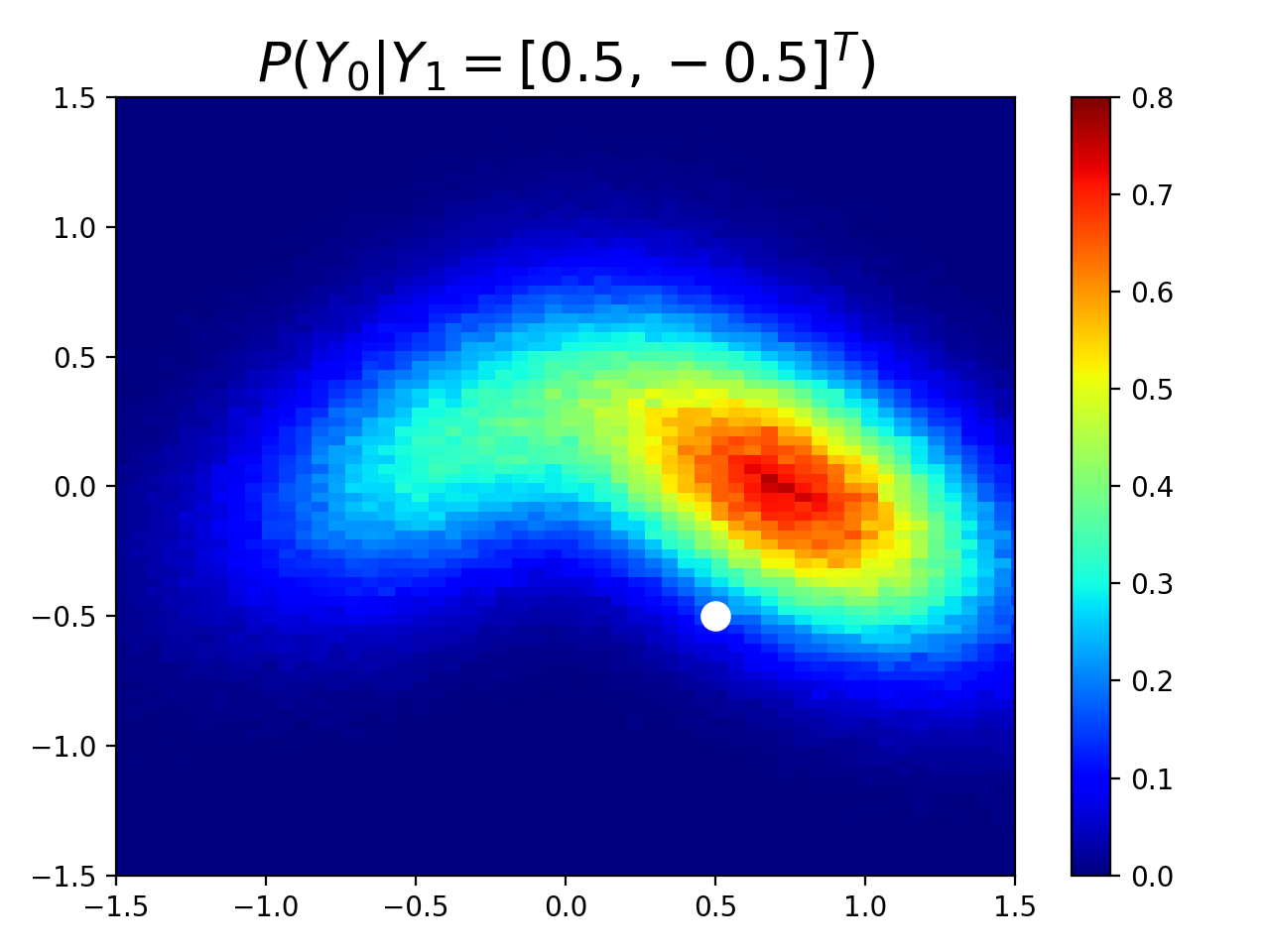}
         \includegraphics[width=0.2\textwidth]{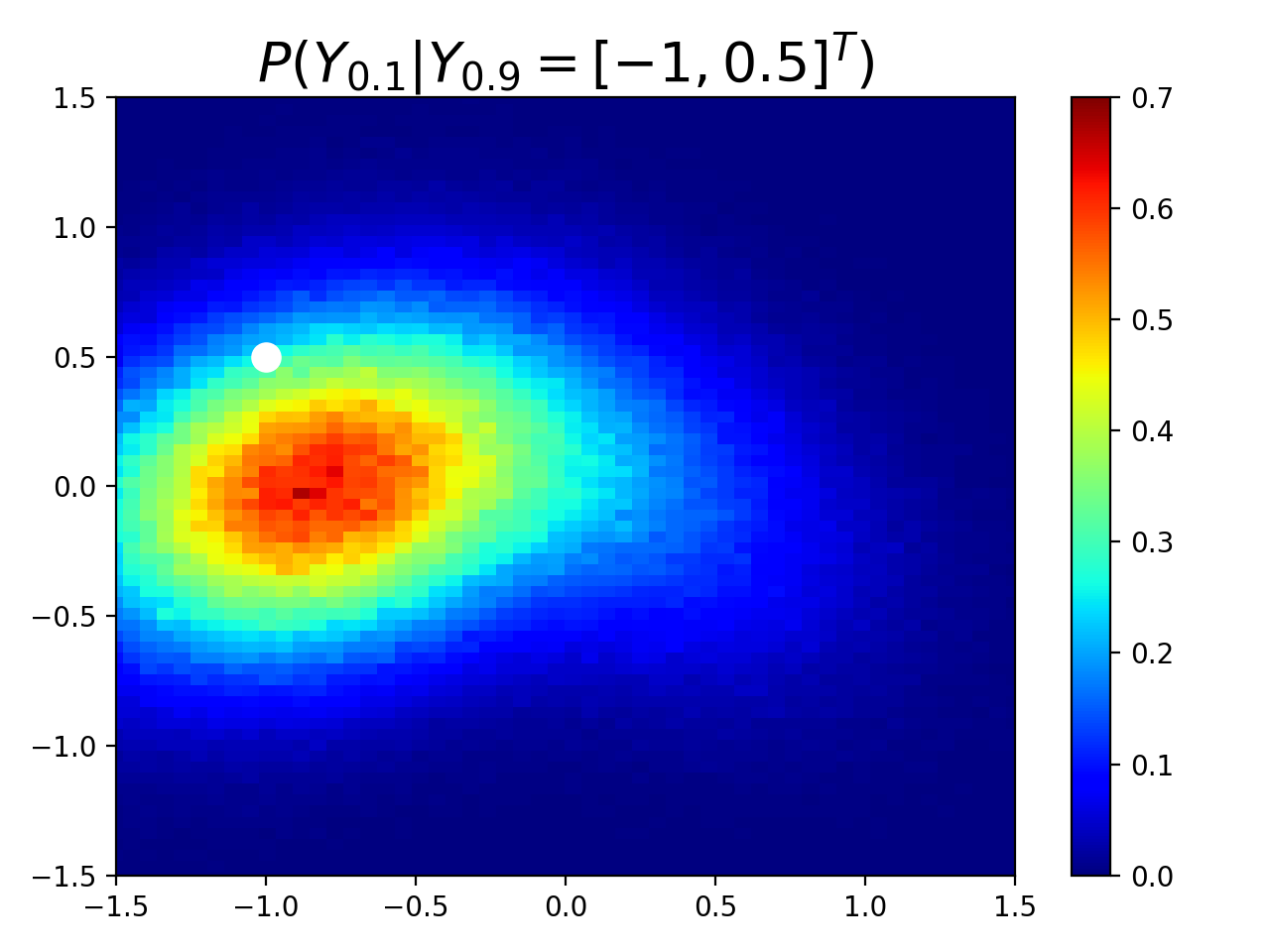}
         \includegraphics[width=0.2\textwidth]{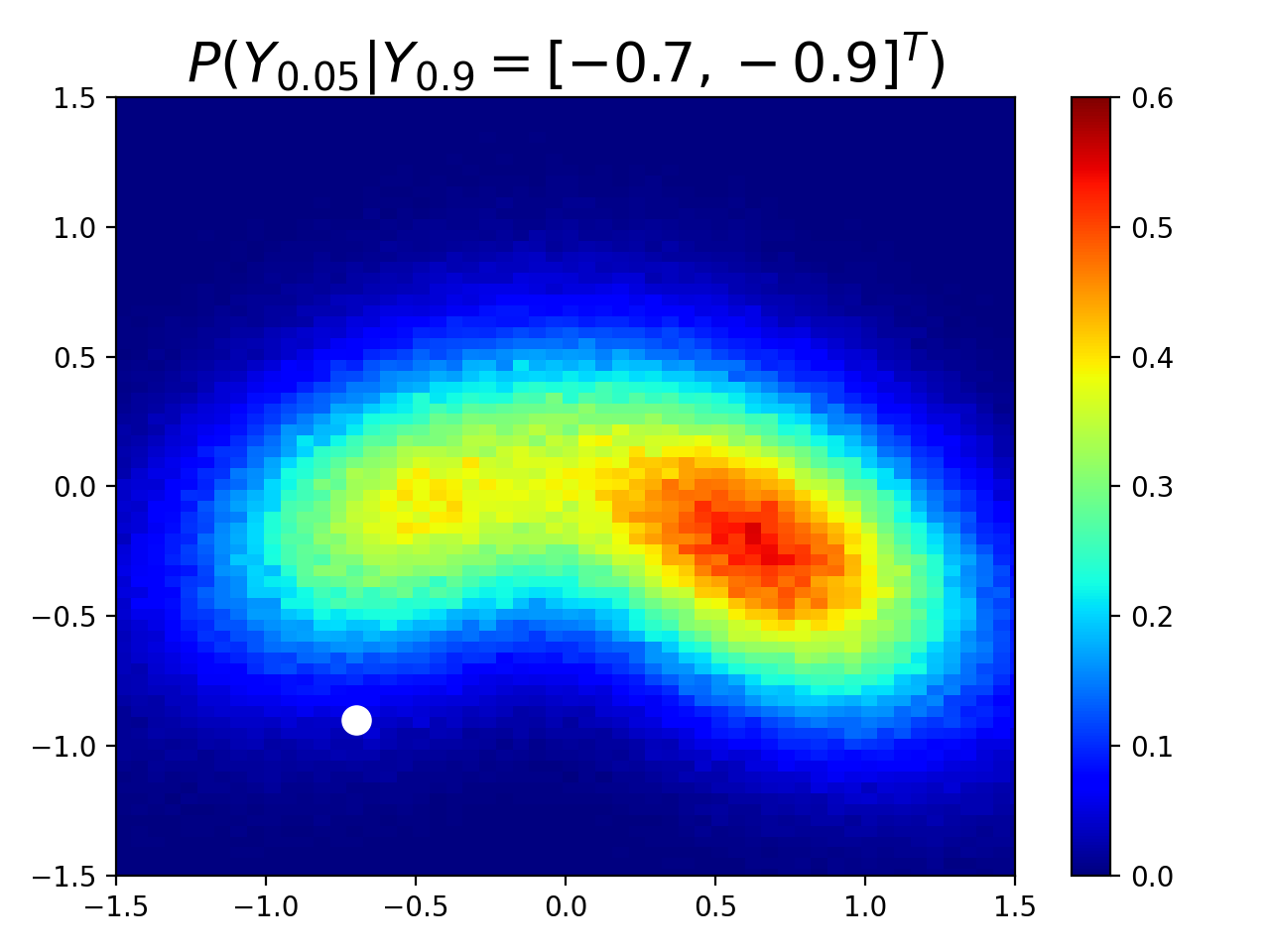}
         \includegraphics[width=0.2\textwidth]{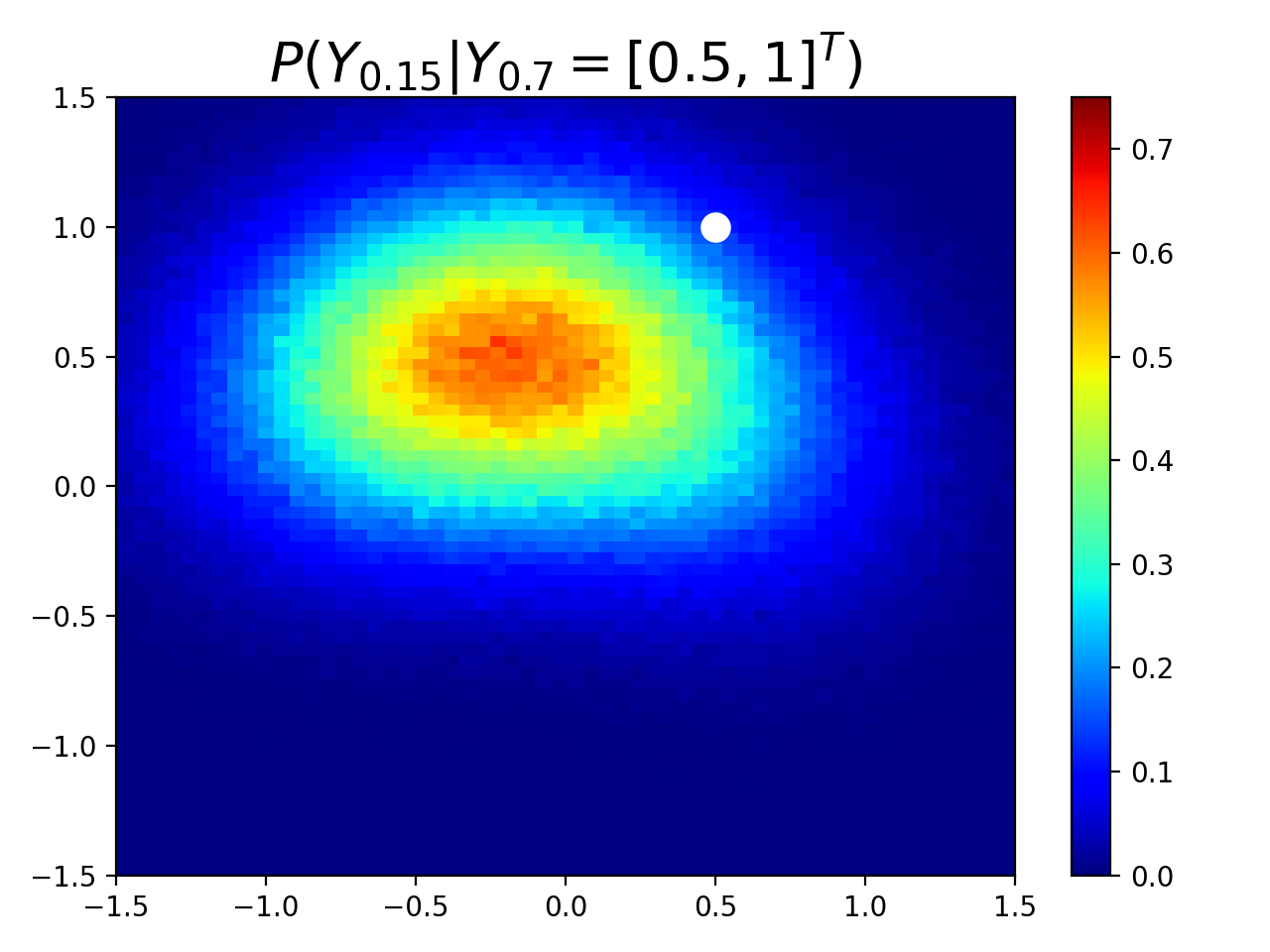}
         \caption{The Riccati-HJ-sampler.}
    \end{subfigure}
    \begin{subfigure}[b]{1\textwidth}
         \centering
         \includegraphics[width=0.2\textwidth]{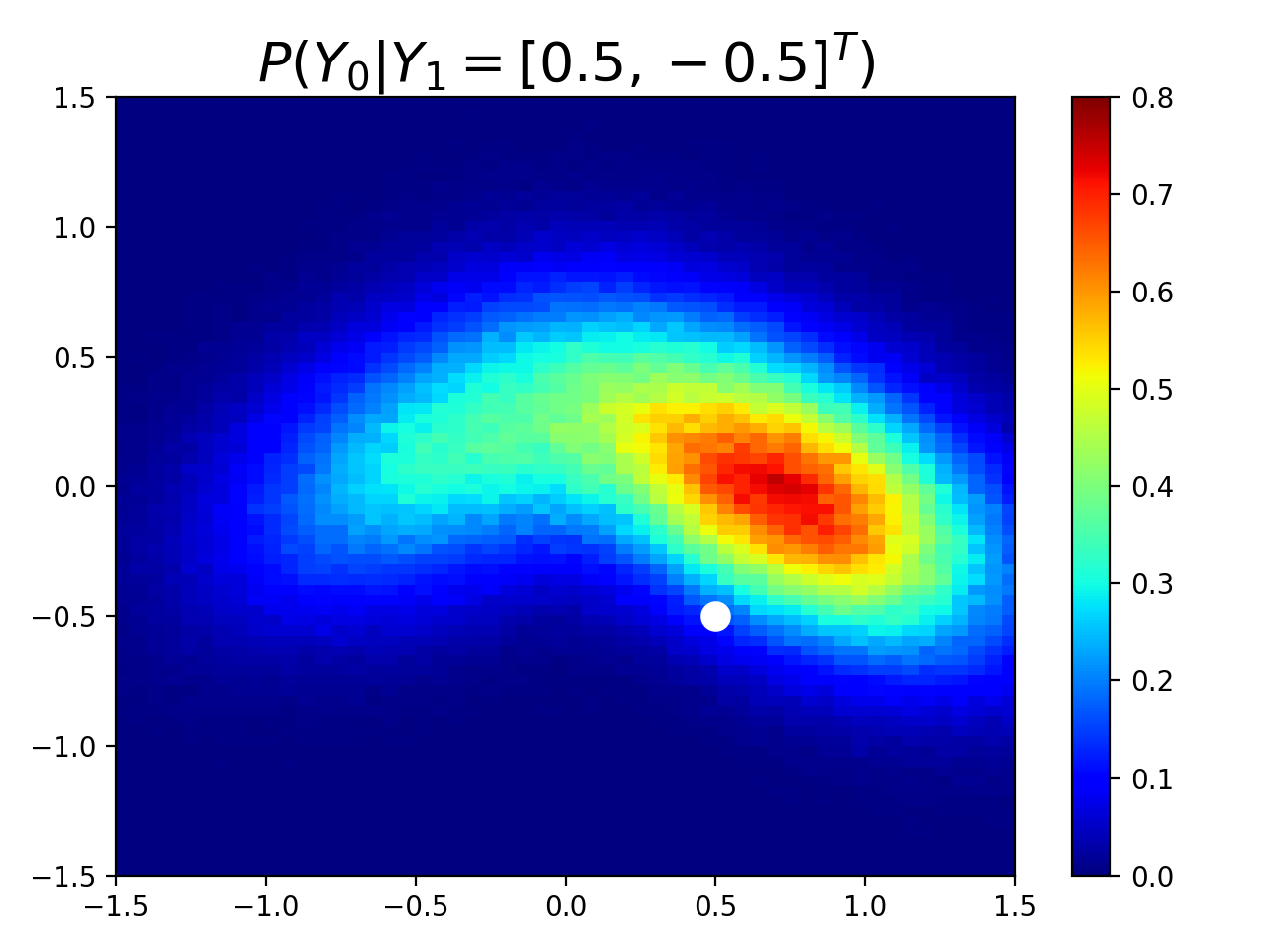}
         \includegraphics[width=0.2\textwidth]{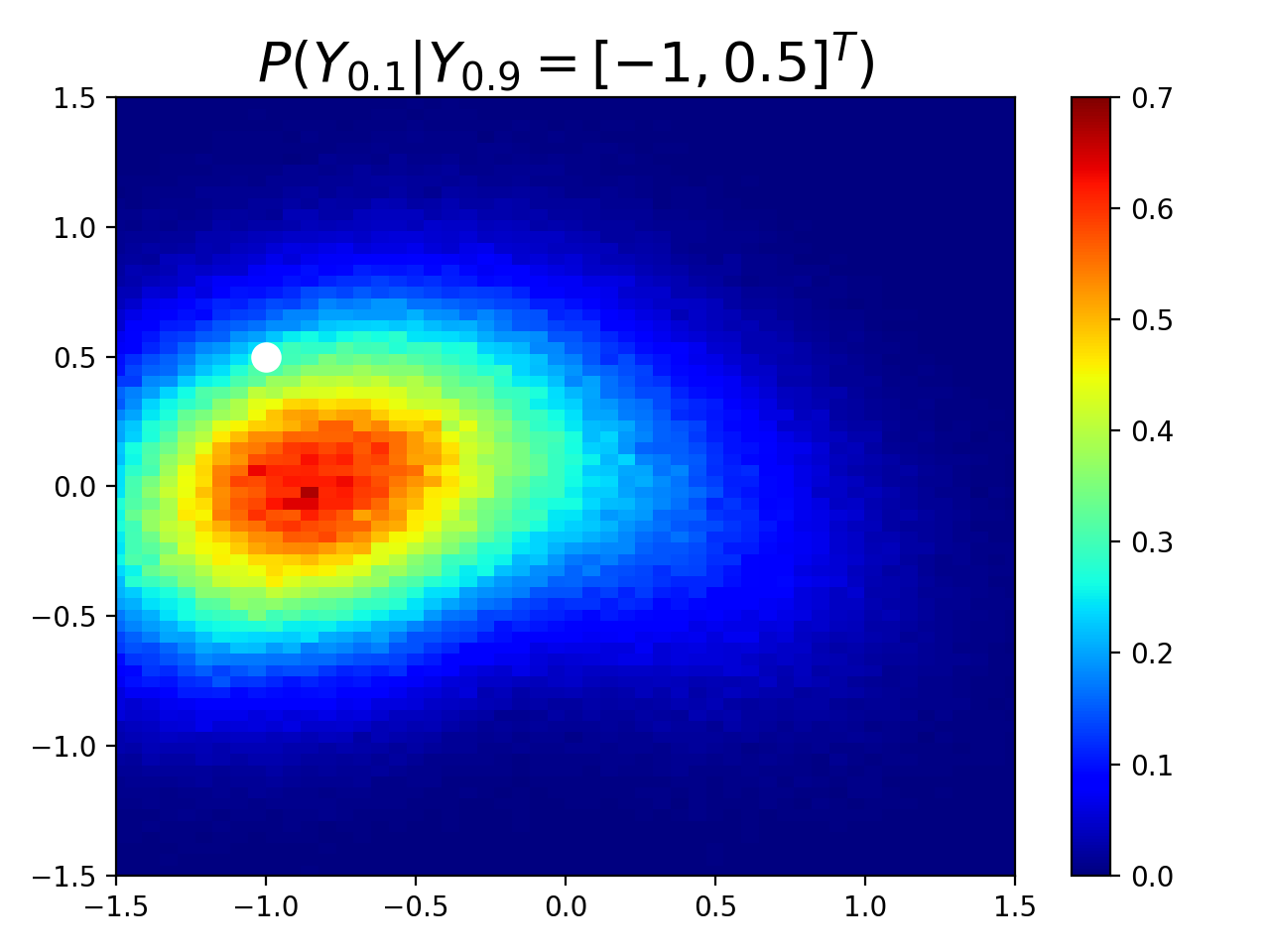}
         \includegraphics[width=0.2\textwidth]{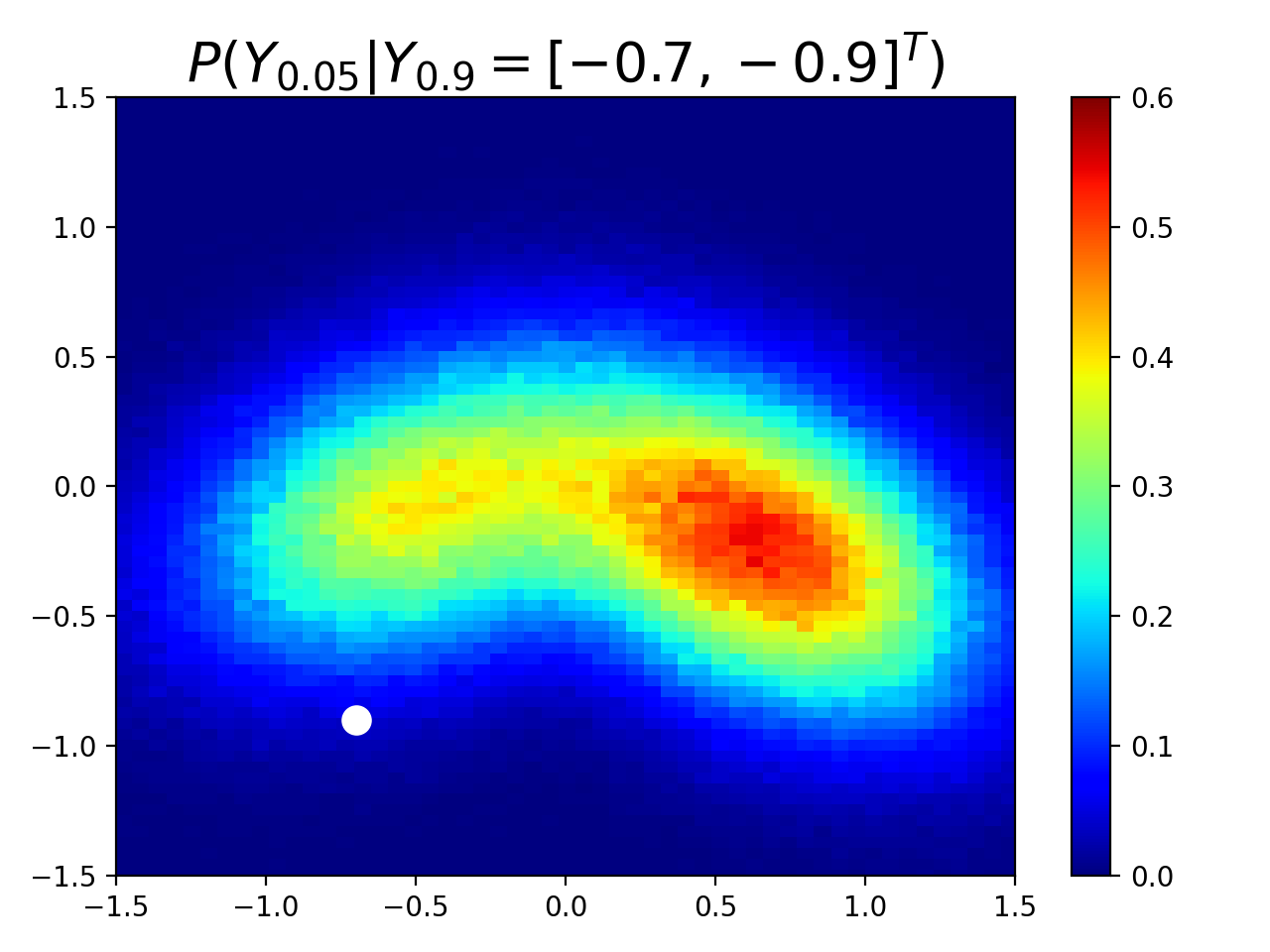}
         \includegraphics[width=0.2\textwidth]{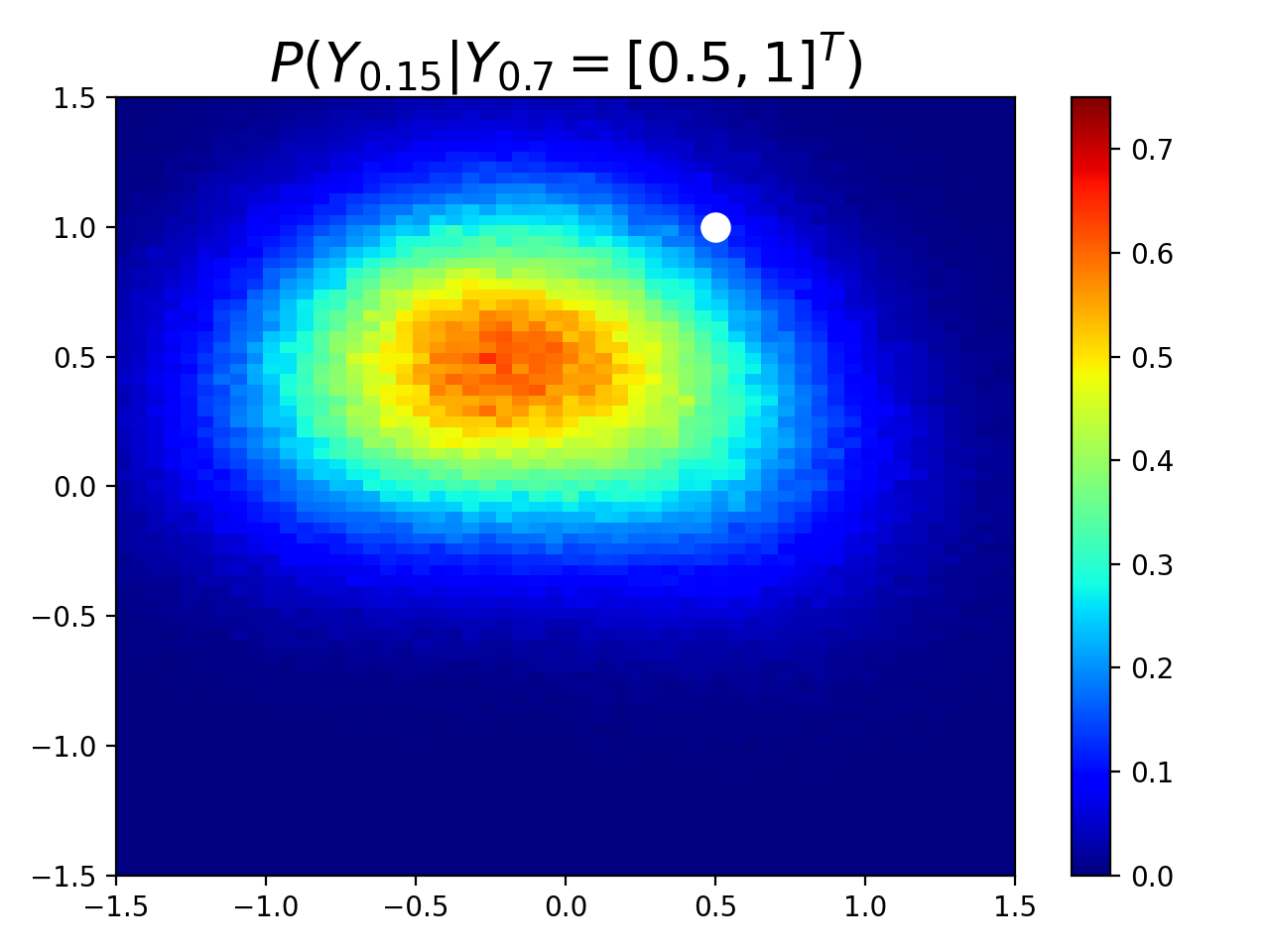}
         \caption{The SGM-HJ-sampler.}
    \end{subfigure}
    \caption{
    Histograms of posterior samples for the 2D OU process with a Gaussian mixture prior. We show the posterior samples of \(Y_t \mid Y_s = \ydata\) computed using the Riccati-HJ-sampler in (a) and SGM-HJ-sampler in (b) for several values of \(\ydata \in \mathbb{R}^2\) and \(0 \leq t<s \leq T\).
    } 
    \label{fig:example_2_1}
\end{figure}

In this section, we demonstrate how to apply the proposed method to handle model uncertainty. We consider the first-order ODE system:
\begin{equation} \label{eq:example_2_1}
\begin{split}
        \frac{dy_1}{dt} &= y_2, 
        \\
        \frac{dy_2}{dt} &= -y_1 - y_2.
    \end{split}
\end{equation}
Model uncertainty is introduced by adding white noise to the right-hand sides of the equations, leading to the OU process~\eqref{eqt:OU_sec4.2} with \(n=2\) and \(B =
\begin{bmatrix}
    0 & -1\\
    1 & 1
\end{bmatrix}\in \mathbb{R}^{2 \times 2}\). Assume the prior distribution of \(Y_0\) is a mixture of two Gaussians with means \(\mu_1 = [-0.7, 0]^T\) and \(\mu_2 = [0.7, 0]^T\) and covariance matrices \(\Sigma_1 = 
\begin{bmatrix}
    0.25 & 0.1\\
    0.1 & 0.16
\end{bmatrix}\) and \(\Sigma_2 = 
\begin{bmatrix}
    0.25 & -0.1\\
    -0.1 & 0.16
\end{bmatrix}\), respectively, with equal weights. We set \(\epsilon = 5\) and \(T = 1\), and discretize \([0, T]\) uniformly with \(\Delta t = 0.01\) to generate the training data for the SGM-HJ-sampler. We apply both the Riccati-HJ-sampler and SGM-HJ-sampler to solve the Bayesian inverse problem, i.e., sampling from \(P(Y_t \mid Y_s = \ydata)\), where \(0 \leq t < s \leq T\), for some specific values of \(\ydata\). 

Since \(B\) is non-diagonal, the exact posterior cannot be derived analytically, and we do not have access to the analytic-HJ-sampler in this case. Therefore, the Riccati-HJ-sampler is used as the reference method to obtain posterior samples of \(Y_t \mid Y_s = \ydata\). We set \(\Delta \tau = 0.001\) to solve the controlled SDE in both HJ-samplers, and the forward Euler scheme with a step size of \(1 \times 10^{-5}\) is employed to solve the Riccati ODE in the Riccati-HJ-sampler.

Results are presented in Figure~\ref{fig:example_2_1}, demonstrating that the SGM-HJ-sampler produces results that closely align with those from the Riccati-HJ-sampler, validating the effectiveness of the SGM-HJ-sampler.

\subsubsection{The second case: model misspecification of a second-order ODE} \label{sec:eg_OU_caseB}

\begin{figure}[ht!]
    \centering
    \includegraphics[width=0.23\textwidth]{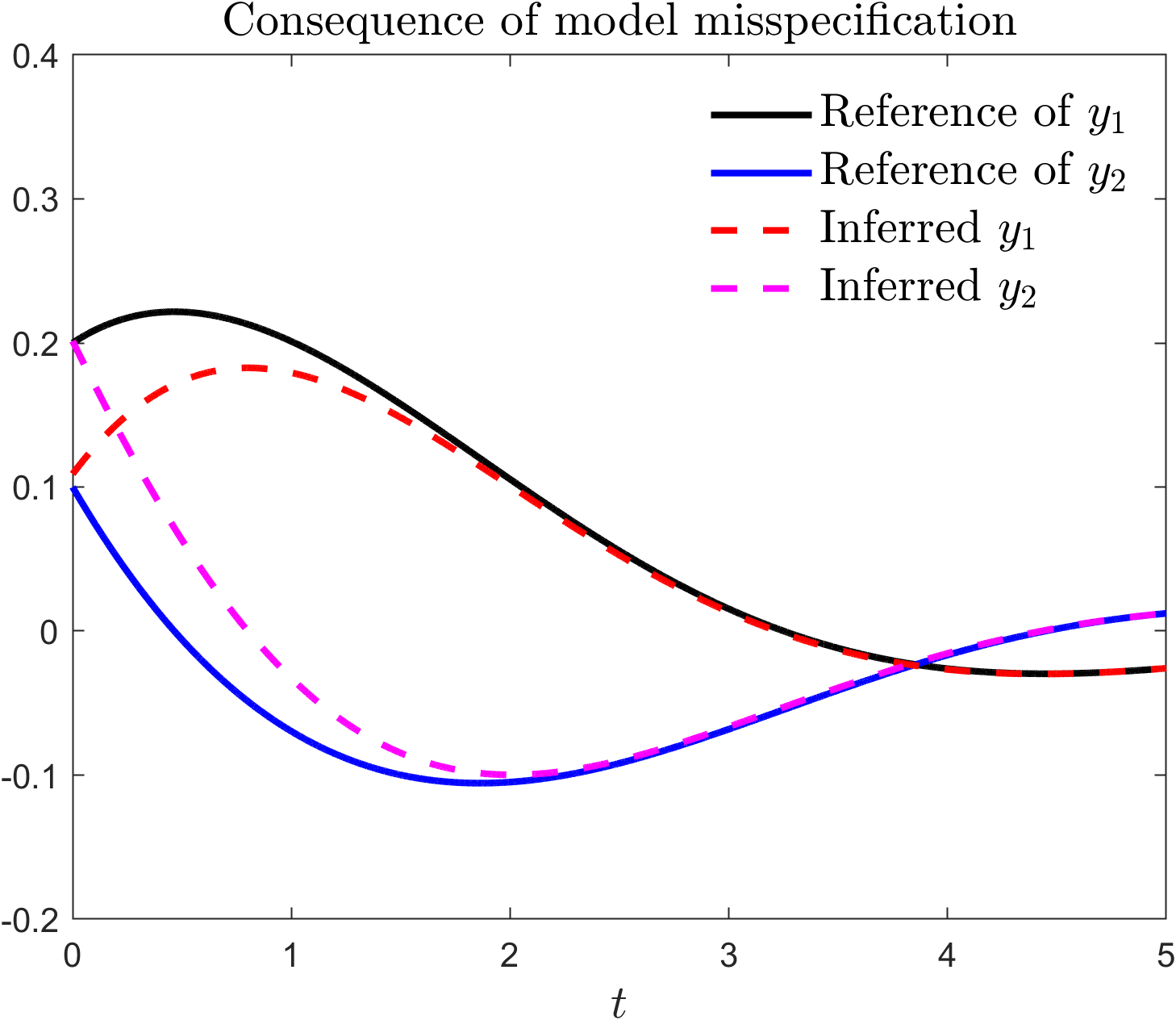}
    \includegraphics[width=0.23\textwidth]{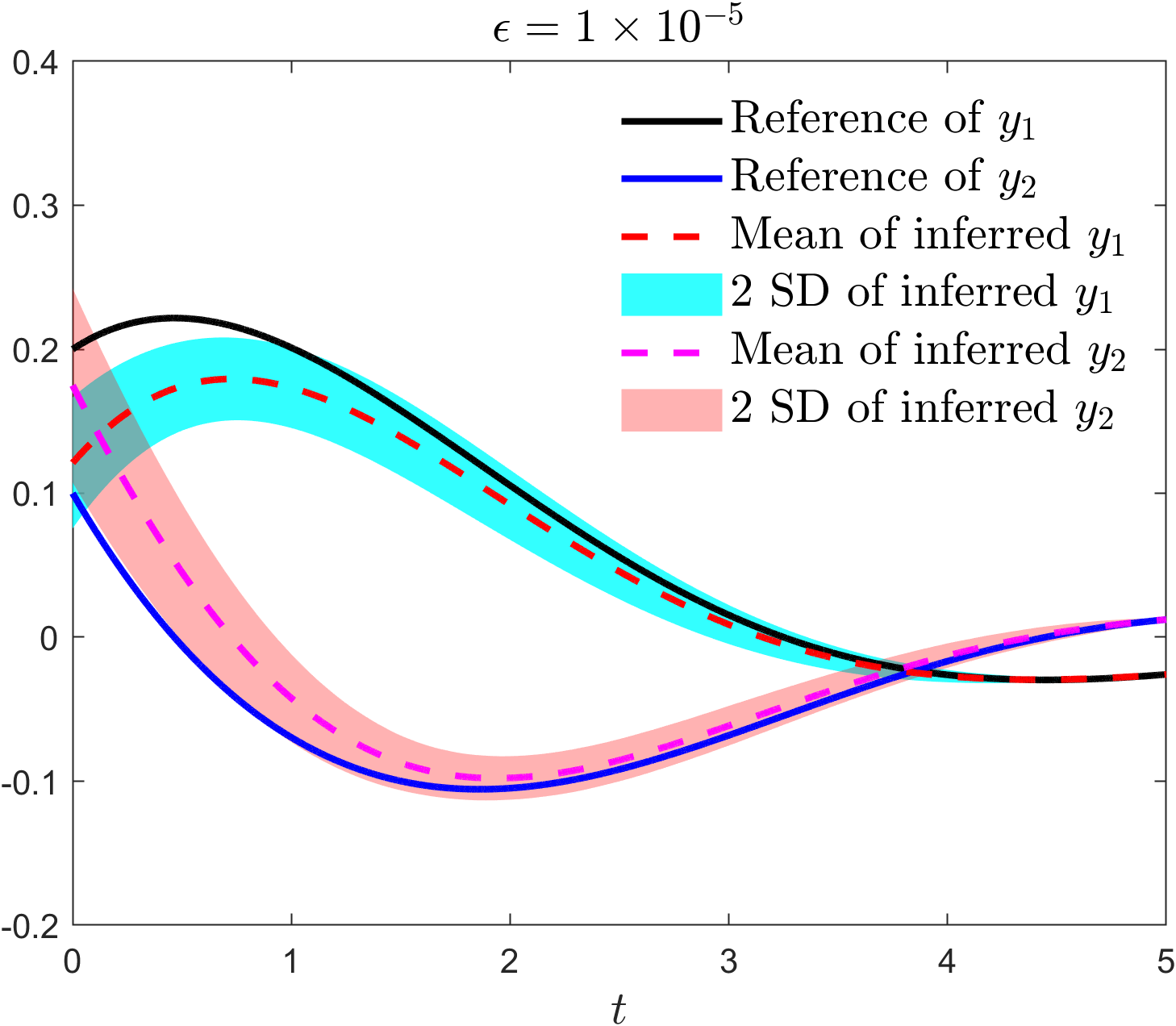}
    \includegraphics[width=0.23\textwidth]{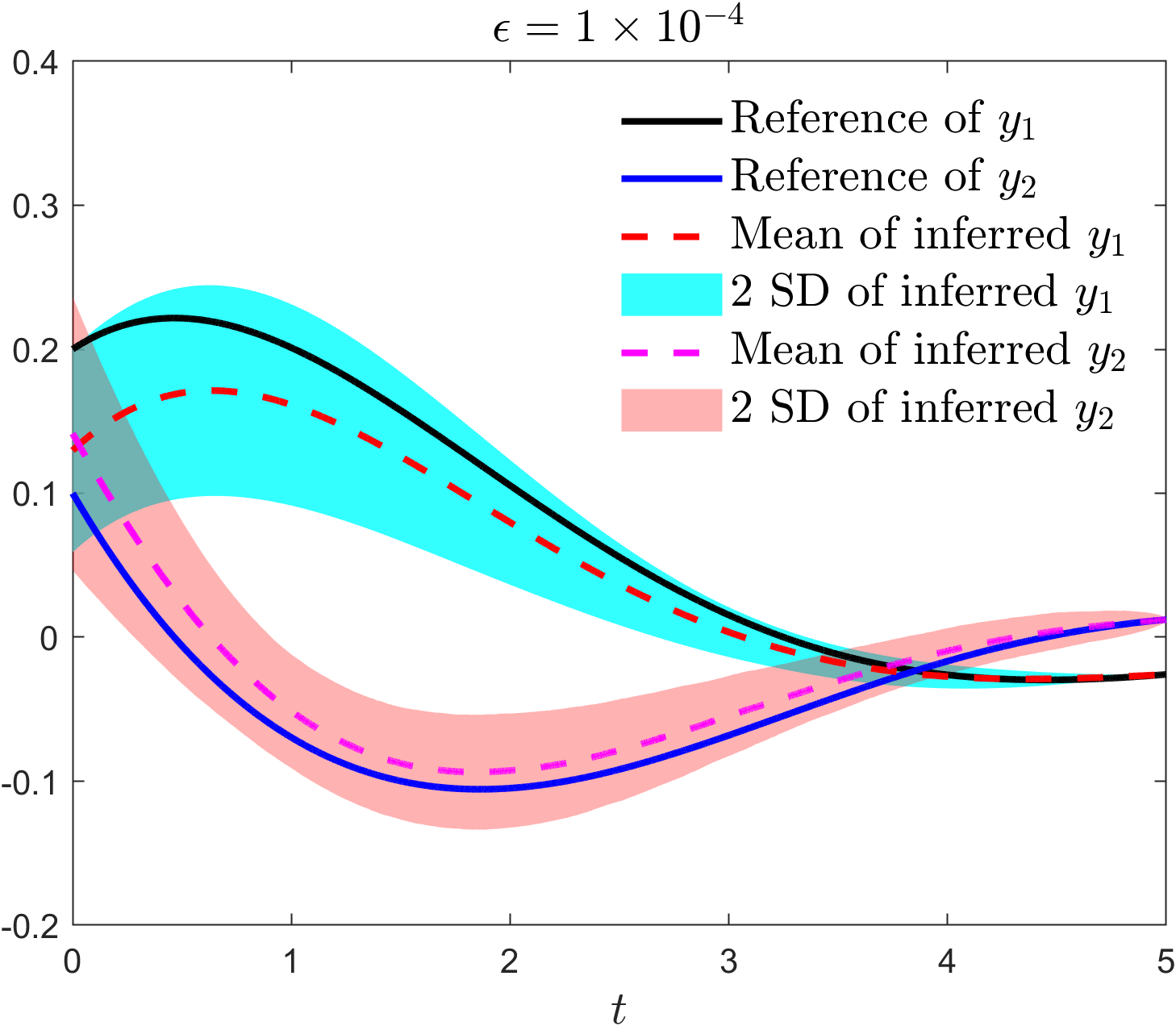}
    \includegraphics[width=0.23\textwidth]{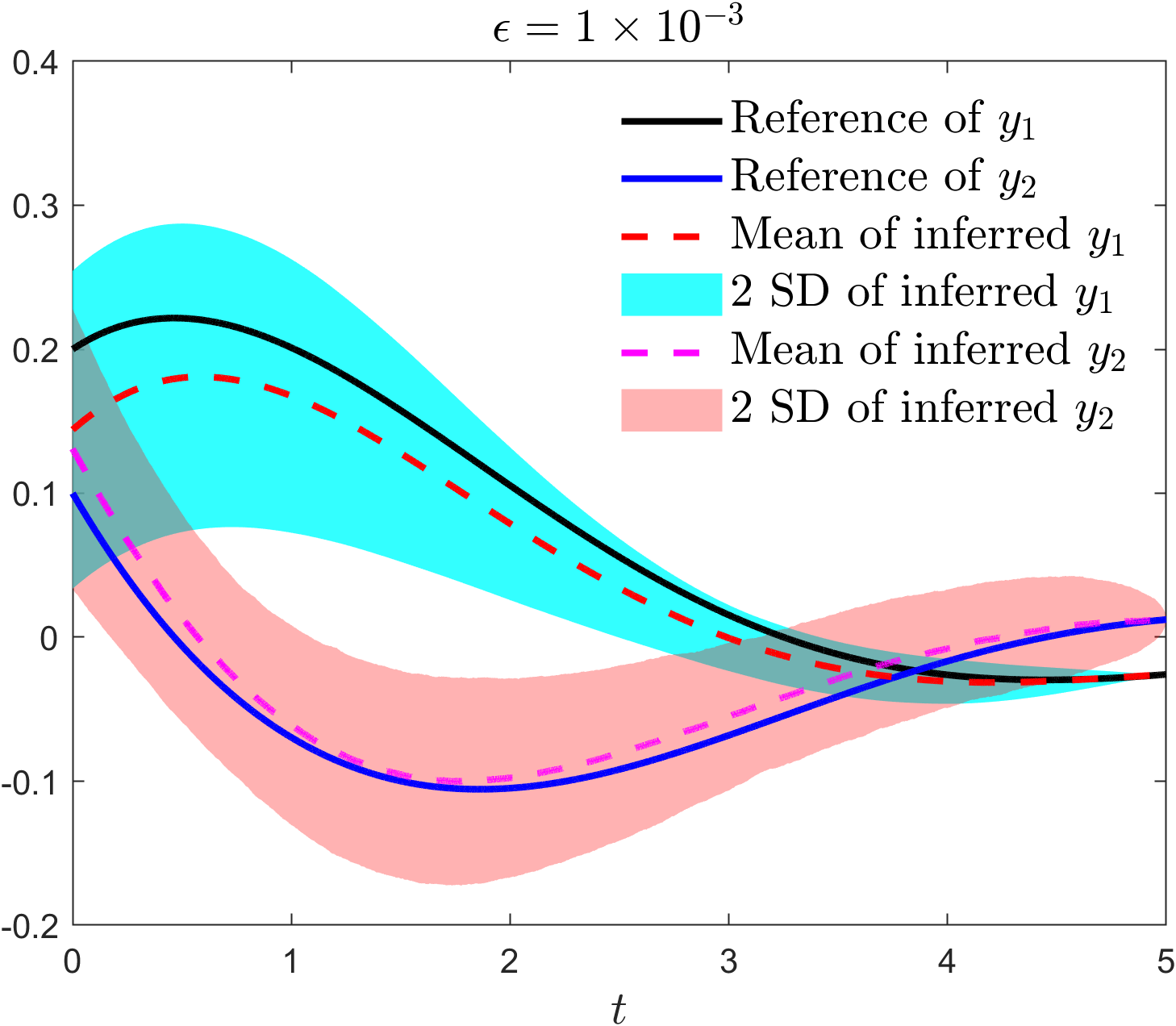}
\caption{
Quantifying model uncertainty in inferring the values of \(Y_1(t)\) and \(Y_2(t)\) for \(t \in [0, T)\), based on \(Y_1(T)\) and \(Y_2(T)\) at \(T = 5\), under different \(\epsilon\) values in~\eqref{eq:example_2_4}. The observations \(Y_1(T)\) and \(Y_2(T)\), along with the reference solutions (solid lines in each figure), are generated by the exact ODE system~\eqref{eq:example_2_2}. The parameter \(\epsilon\) represents the uncertainty level in the misspecified model. The SGM-HJ-sampler is employed to generate posterior samples, which are then used to compute the posterior means (dashed lines) and 2-standard-deviation intervals (color-shaded regions) in the right three figures. Model uncertainty refers to the uncertainty in the differential equation due to misspecification of a term on the right-hand side of~\eqref{eq:example_2_2b}. The leftmost figure illustrates the reference trajectories of \(y_1\) and \(y_2\) (solid lines, from the exact ODE system~\eqref{eq:example_2_2}) and the inferred \(y_1\) and \(y_2\) (dashed lines, from the misspecified ODE system~\eqref{eq:example_2_3}). The discrepancy between the solid and dashed lines highlights the impact of model misspecification, emphasizing the need to account for the associated uncertainty. 
} 
\label{fig:example_2_3}
\end{figure}

In this section, we illustrate how the proposed method addresses model misspecification. Consider an exact but unknown ODE given by \( u''(t) = -u(t) + u(t)^2 - u'(t) \), which is misspecified as \( \tilde u''(t) = -\tilde u(t) - \tilde u'(t) \). This discrepancy necessitates modeling the error and quantifying the uncertainty induced by the incorrect model, as discussed in \cite{zou2024correcting}. By defining \( y_1(t) = u(t) \) and introducing the auxiliary variable \( y_2(t) = u'(t) \), the exact ODE transforms into:
\begin{subequations}\label{eq:example_2_2}
    \begin{align}
        \frac{dy_1}{dt} &= y_2, \label{eq:example_2_2a}\\
        \frac{dy_2}{dt} &= -y_1 + y_1^2 - y_2. \label{eq:example_2_2b}
    \end{align}
\end{subequations}
Similarly, letting \( \tilde y_1(t) = \tilde u(t) \) and \( \tilde y_2(t) = \tilde u'(t) \), the misspecified model becomes:
\begin{subequations}\label{eq:example_2_3}
    \begin{align}
        \frac{d\tilde y_1}{dt} &= \tilde y_2, \label{eq:example_2_3a}\\
        \frac{d\tilde y_2}{dt} &= -\tilde y_1 - \tilde y_2.\label{eq:example_2_3b}
    \end{align}
\end{subequations}

To account for the model uncertainty, we reformulate the misspecified ODE system \eqref{eq:example_2_3} into the following SDE:
\begin{equation}\label{eq:example_2_4}
    \begin{split}
        dY_1(t) &= Y_2(t) \, dt,\\
        dY_2(t) &= (-Y_1(t) - Y_2(t)) \, dt + \sqrt{\epsilon} \, dW_t,
    \end{split}
\end{equation}
where we use capital letters \(Y_1(t)\) and \(Y_2(t)\) to denote random variables, which approximate the corresponding deterministic values denoted by lowercase letters \(y_1(t)\) and \(y_2(t)\) from the exact ODE model~\eqref{eq:example_2_2}. The capital letters represent the stochastic versions of the corresponding lowercase variables, reflecting the uncertainty in the misspecified model. Here, \(\epsilon\) is a hyperparameter controlling the level of uncertainty in the system; a larger \(\epsilon\) corresponds to greater uncertainty in the model equations, while a smaller \(\epsilon\) indicates higher confidence in the misspecified ODE.

Our goal is to infer the values of \(Y_1(t)\) and \(Y_2(t)\) for \(t \in [0, T)\) based on the values of \(Y_1(T)\) and \(Y_2(T)\), with \(T=5\), while accounting for the model uncertainty in \eqref{eq:example_2_4} using the SGM-HJ-sampler. Specifically, the data for \(Y_1(T)\) and \(Y_2(T)\), which are based on the exact ODE solutions \(y_1(T)\) and \(y_2(T)\), as well as the reference trajectories \(y_1(t)\) and \(y_2(t)\) for \(t \in [0, T)\), are generated by numerically solving the exact nonlinear ODE \eqref{eq:example_2_2} with the initial conditions \(y_1(0) = 0.2\) and \(y_2(0) = 0.1\) using the SciPy \textit{odeint} function \cite{2020SciPy-NMeth}. We consider three levels of uncertainty, \(\epsilon = 1 \times 10^{-3}, 1 \times 10^{-4}, 1 \times 10^{-5}\), and assume the prior distributions of \(Y_1(0)\) and \(Y_2(0)\) are independent log-normal distributions, specifically \( \text{LogNormal}(-2, 0.5^2) \). Since the prior distribution is neither Gaussian nor a Gaussian mixture, the Riccati-HJ-sampler is not applicable. Therefore, we use the SGM-HJ-sampler with $\Delta\tau=0.001$ to solve this problem.

We present the results in Figure~\ref{fig:example_2_3}, where each panel displays the reference trajectories \(y_1(t)\) and \(y_2(t)\) as solid lines. In the leftmost panel, we show the inference of \(y_1(t)\) and \(y_2(t)\) for \(t \in [0, T)\), based solely on \(y_1(T)\) and \(y_2(T)\), by solving the misspecified linear ODE \eqref{eq:example_2_3} backward in time. The gap between the reference trajectories \((y_1(t), y_2(t))\) and the inferred trajectories \((\tilde y_1(t), \tilde y_2(t))\) highlights the necessity of incorporating model uncertainty when addressing model misspecification.

The remaining panels in Figure~\ref{fig:example_2_3} show the results from the SGM-HJ-sampler, with each panel corresponding to a different uncertainty level, \(\epsilon\). Specifically, after obtaining the posterior samples ($1\times10^3$) using the SGM-HJ-sampler, we compute the posterior means and standard deviations. The posterior means are represented by dashed lines, while the 2-standard-deviation regions are shown in color, visualizing the uncertainty in the inferred solutions. From these panels, we observe that the uncertainty in the inferred values of \(Y_1(t)\) and \(Y_2(t)\) increases as \(\epsilon\) increases. In practice, \(\epsilon\) can be interpreted as a confidence hyperparameter for the model \eqref{eq:example_2_3b}, where higher confidence corresponds to a smaller \(\epsilon\).

In this example, three distinct neural networks (\(s_\NNparam\)) were trained, one for each value of \(\epsilon\), leading to three separate SGM-HJ-samplers. To enhance flexibility with respect to the confidence hyperparameter \(\epsilon\), we could include \(\epsilon\) as an input to the neural network, allowing it to adapt to different confidence levels without requiring retraining. This example also highlights the capability of our method to handle partial uncertainty, where only certain equations within the system are subject to uncertainty.

\subsection{Model misspecification for a nonlinear ODE}\label{sec:numerics_eg3}

\begin{figure}[ht!]
    \begin{subfigure}[b]{1\textwidth}
         \centering
         \includegraphics[width=0.23\textwidth]{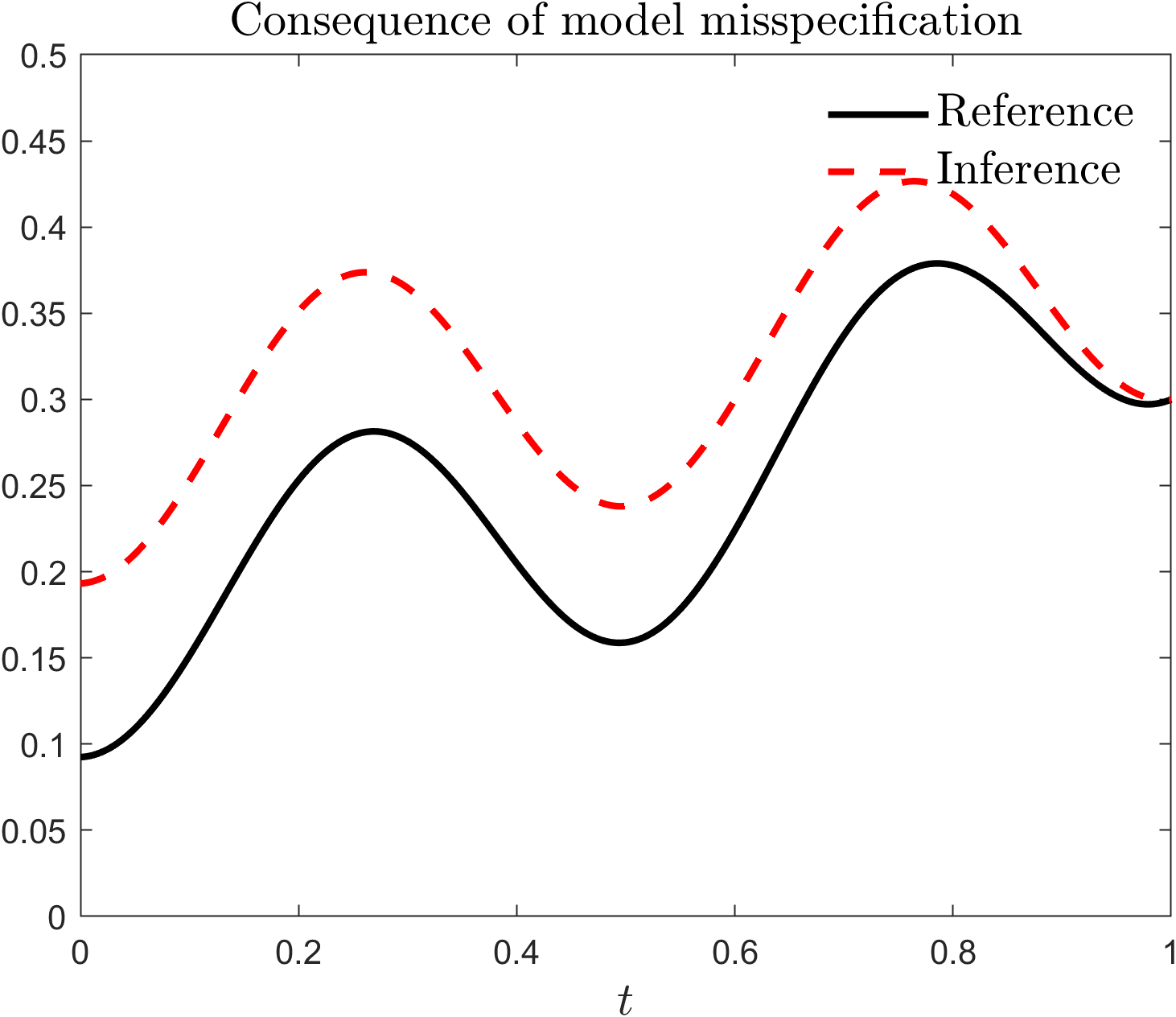}
         \includegraphics[width=0.23\textwidth]{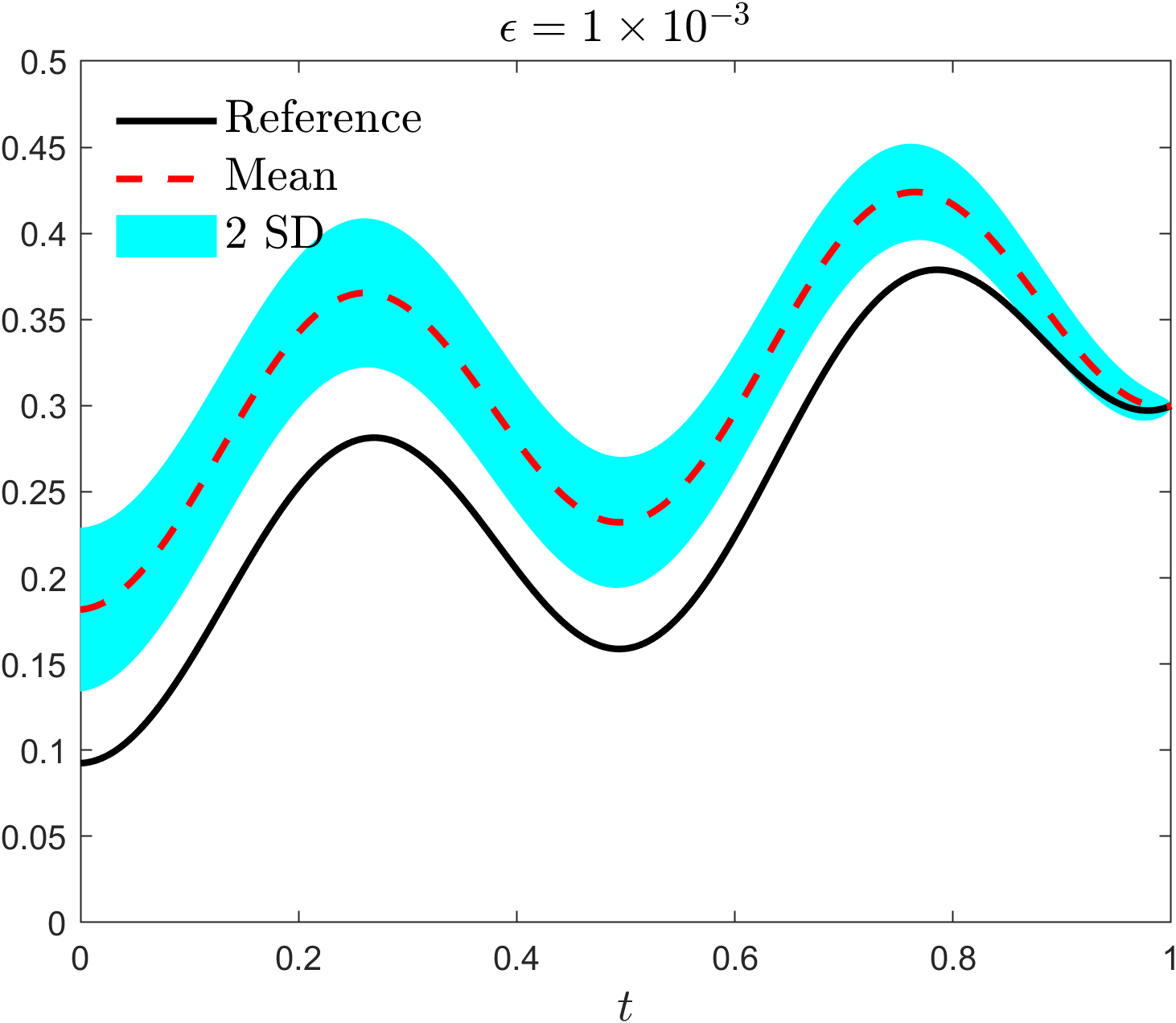}
         \includegraphics[width=0.23\textwidth]{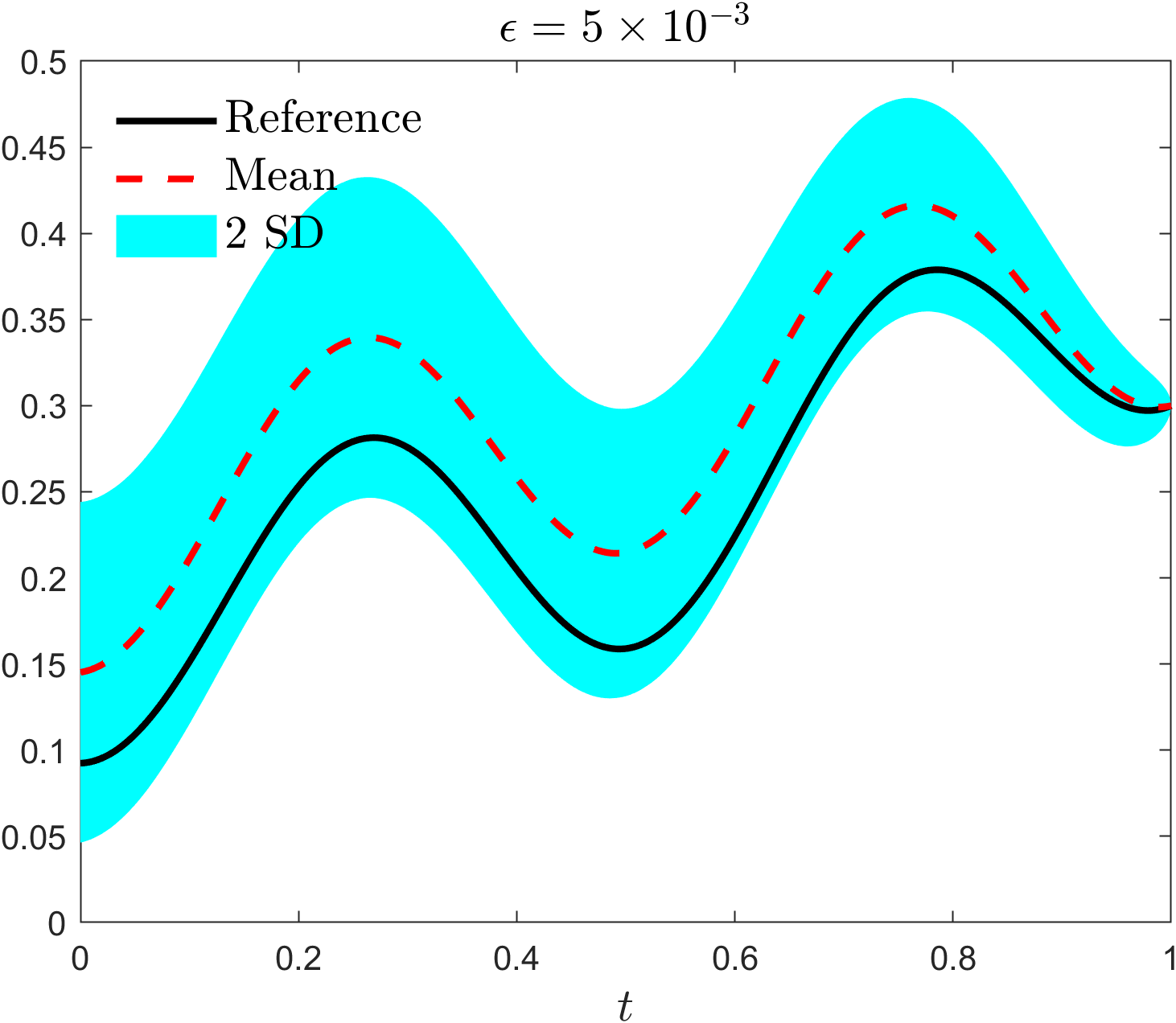}
         \includegraphics[width=0.23\textwidth]{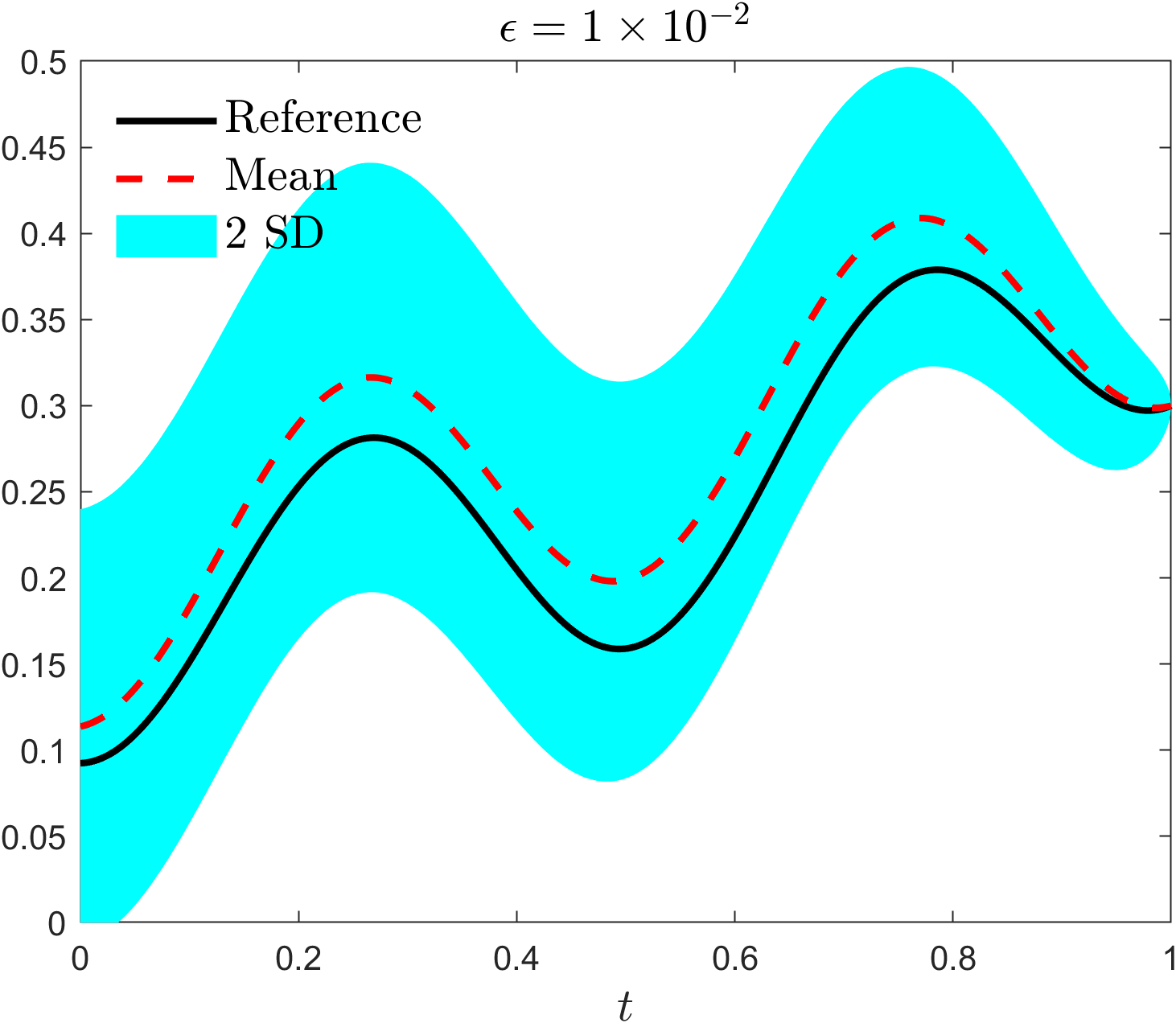}
         \caption{The exact model is $g(y) = 3 y^2$, misspecified as $\tilde{g}(y) = y^2$.}
    \end{subfigure}
    \begin{subfigure}[b]{1\textwidth}
         \centering
         \includegraphics[width=0.23\textwidth]{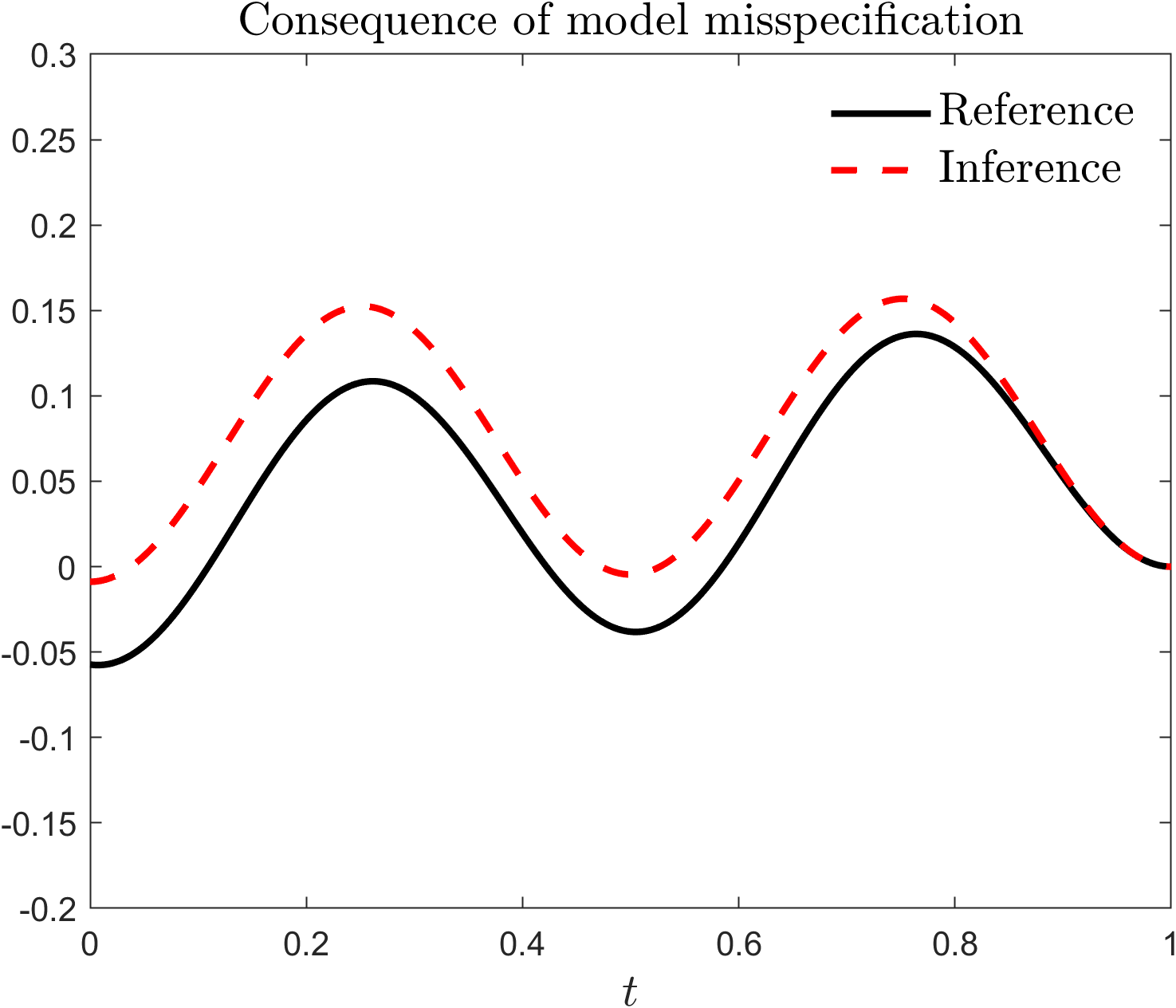}
         \includegraphics[width=0.23\textwidth]{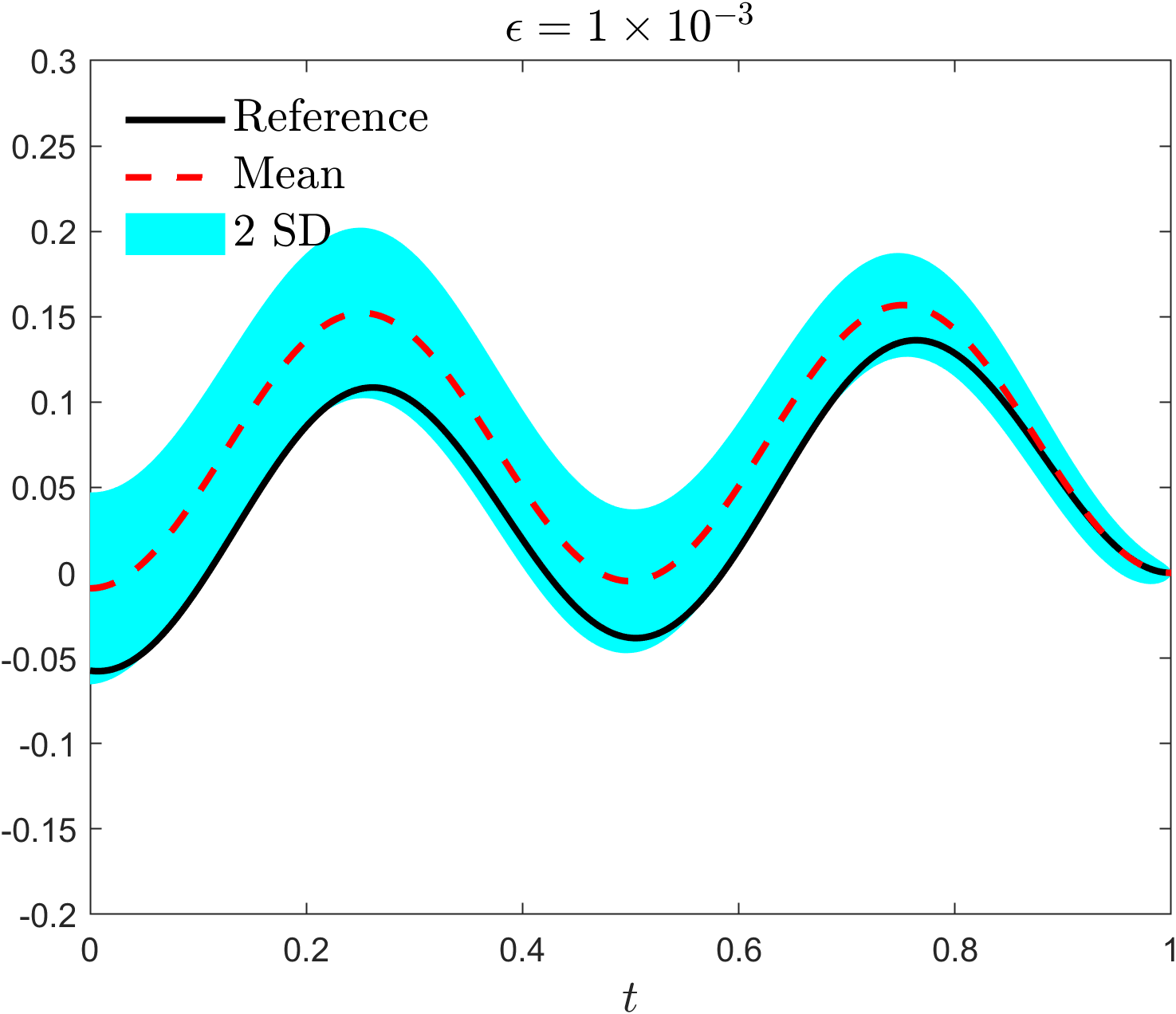}
         \includegraphics[width=0.23\textwidth]{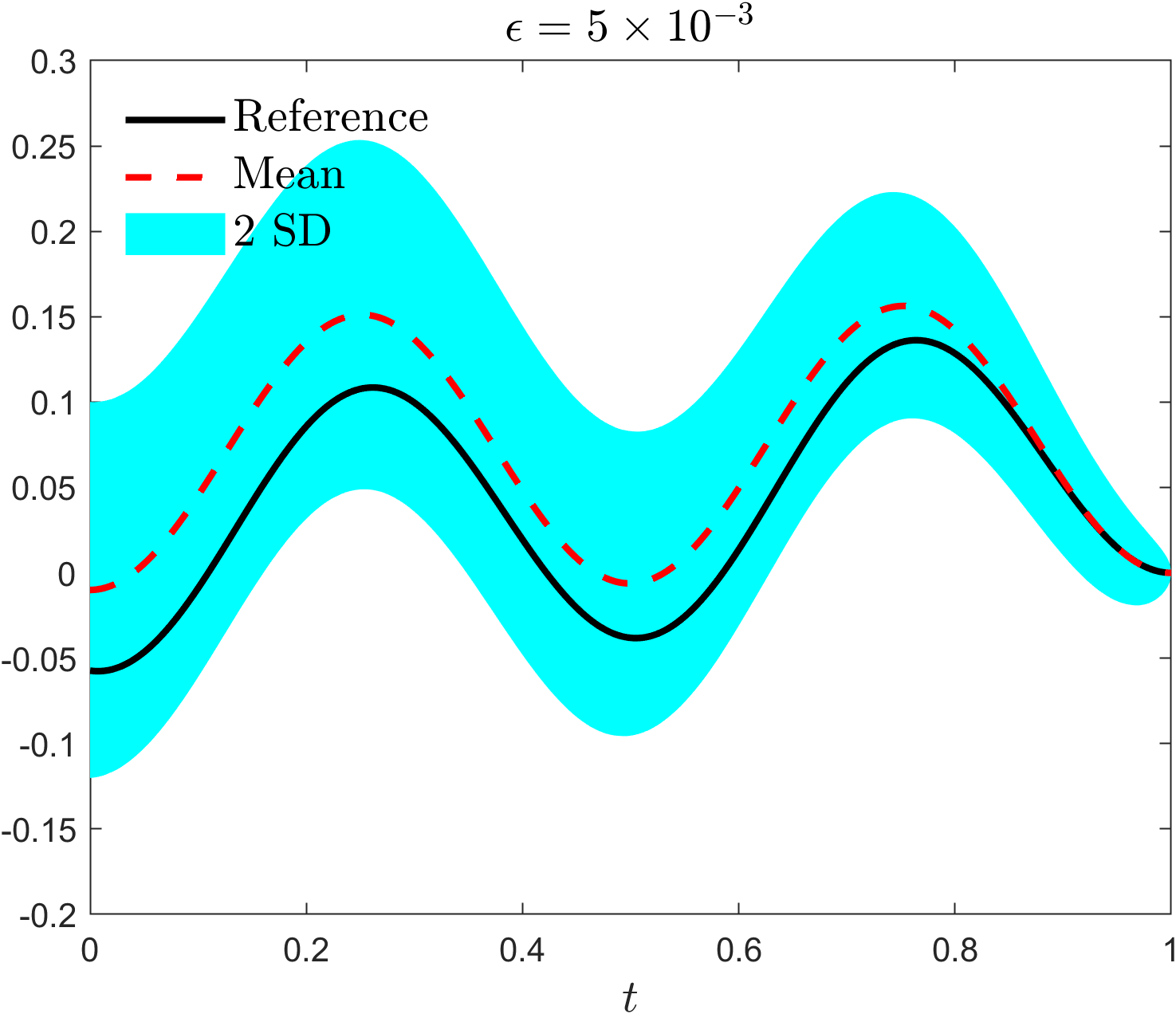}
         \includegraphics[width=0.23\textwidth]{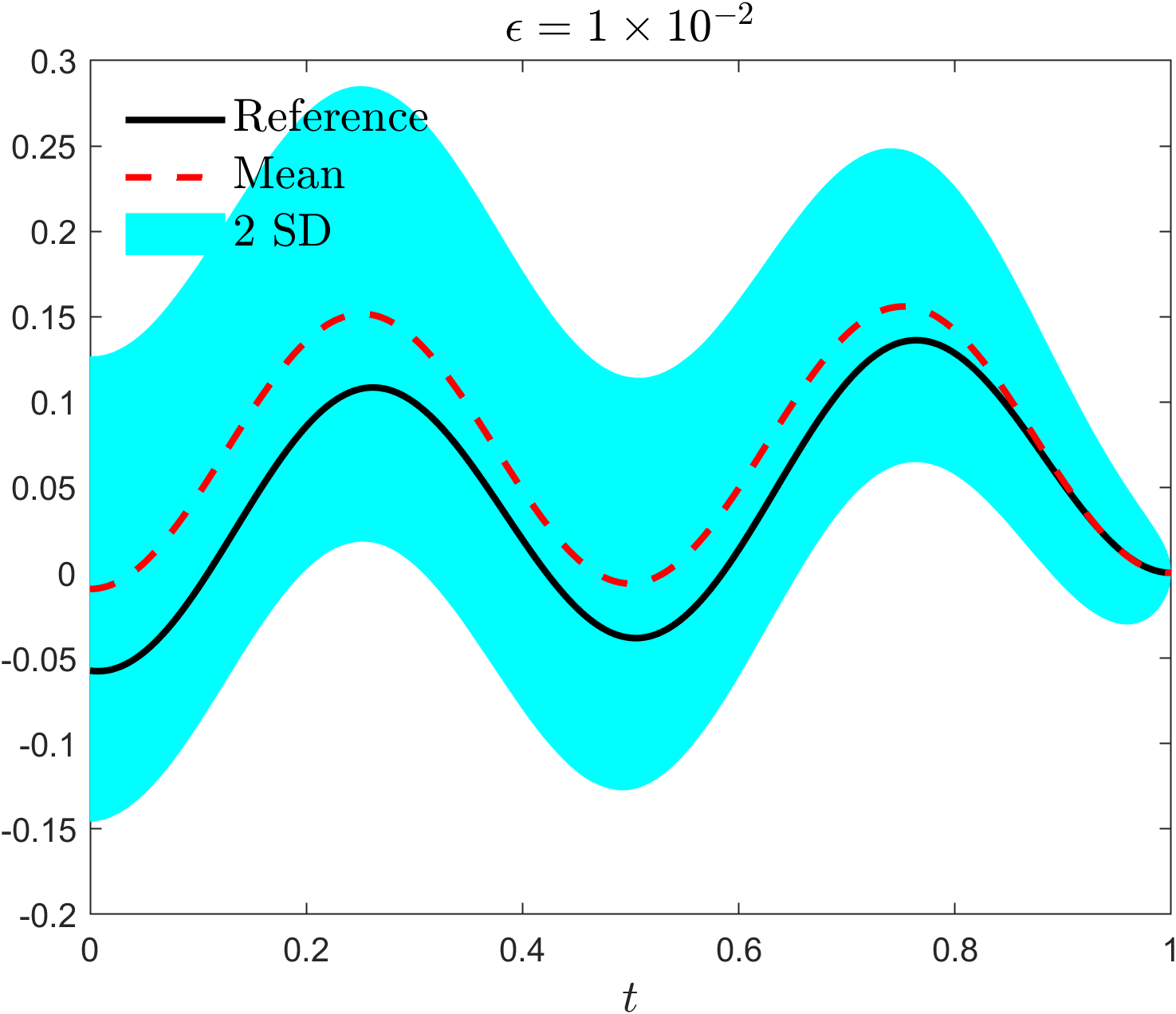}
         \caption{The exact model is $g(y) = 1.5y(1-y)$, misspecified as $\tilde{g}(y) = y^2$.}
    \end{subfigure}
\caption{Quantifying model uncertainty in inferring \(y(t)\), \(t \in [0, T)\), from \(y(T)\) (\(T=1\)) under different cases of model misspecification. The leftmost panel shows the reference solution (exact ODE) and the inference obtained by solving the misspecified ODE backward in time. The difference highlights the consequence of model misspecification, emphasizing the need to quantify uncertainty. The right three panels display the references (solid), posterior means (dashed), and 2-standard-deviation regions (colored) computed using the SGM-HJ-sampler with varying uncertainty levels (\(\epsilon\)).} 
\label{fig:example_3}
\end{figure}

In this section, we demonstrate the effectiveness of the proposed approach in addressing model misspecification in a nonlinear 1D ODE through a Bayesian inverse problem. Specifically, we consider the following ODE:
\begin{equation}\label{eq:example_3}
    \begin{aligned}
        &\frac{dy(t)}{dt} = f(t) + g(y(t)), \quad t\in[0, T],\\
        & y(0) = y_0,
    \end{aligned}
\end{equation}
where \( y_0 \) is the initial condition, \( f(t) = \sin(4\pi t) \) is the known source term, and \( g(y(t)) \) is the term that is misspecified as \( \tilde{g}(y(t)) \).
Our objective is to infer the values of \( y(t) \) for \( t \in [0, T) \), given the observed value \(\ydata= y(T) \), while accounting for uncertainty due to the misspecification of \( g(y) \). To model this uncertainty, we reformulate the ODE as the following SDE:
\begin{equation}\label{eq:example_3_1}
    dY_t = (f(t) + \tilde{g}(Y_t)) dt + \sqrt{\epsilon} dW_t,
\end{equation}
where 
\( \epsilon \) is a positive constant that controls the level of uncertainty. The parameter \( \epsilon \) can be interpreted as a confidence level: smaller values of \( \epsilon \) indicate higher confidence in the model.
The prior distribution of \(Y_0\) is assumed to be \(\mathcal{N}(0, 0.1^2)\). We discretize the time domain \([0, T]\) uniformly with \(\Delta t = 0.01\) to generate the training data.

We assume \( T = 1 \) and consider two distinct cases of model misspecification:
\begin{enumerate}
    \item[(a)] The exact model is \( g(y) = 3 y^2 \), but it is misspecified as \( \tilde{g}(y) = y^2 \).
    \item[(b)] The exact model is \( g(y) = 1.5y(1-y) \), but it is misspecified as \( \tilde{g}(y) = y^2 \).
\end{enumerate}
In Case (a), the model form is correct but the coefficient is wrong, while in Case (b), the model \( g \) itself is misspecified. The values of \( y(T) \) and the reference solutions for \( y(t) \), for \( t \in [0, T) \), are obtained by numerically solving the correctly specified ODE with initial conditions \( y_0 = 0.05 \) in (a) and \( y_0 = -0.1 \) in~(b). 

The results are presented in Figure~\ref{fig:example_3}, following a similar presentation style as Figure~\ref{fig:example_2_3}. The leftmost panel shows the consequences of model misspecification, where the inferred solution is computed by solving the misspecified ODE backward in time. The remaining three panels show the posterior mean (dashed lines) and 2-standard-deviation regions (colored areas) computed from the posterior samples ($1\times 10^5$) generated by the SGM-HJ-sampler, with different levels of uncertainty corresponding to \(\epsilon = 1 \times 10^{-3}\), \(5 \times 10^{-3}\), and \(1 \times 10^{-2}\). We set $\Delta\tau=0.001$ in solving the controlled SDE. The solid lines represent the reference solutions for \( y(t) \) from the correctly specified ODE. These results show that as \( \epsilon \) decreases, the uncertainty in the inferred values of \( y(t) \) decreases accordingly. Overall, this example demonstrates the SGM-HJ-sampler's capability to effectively account for model misspecification in nonlinear systems.

\subsection{A high-dimensional example}\label{sec:numerics_eg4}

\begin{figure}[ht!]
    \begin{subfigure}[b]{1\textwidth}
         \centering
         \includegraphics[width=0.18\textwidth]{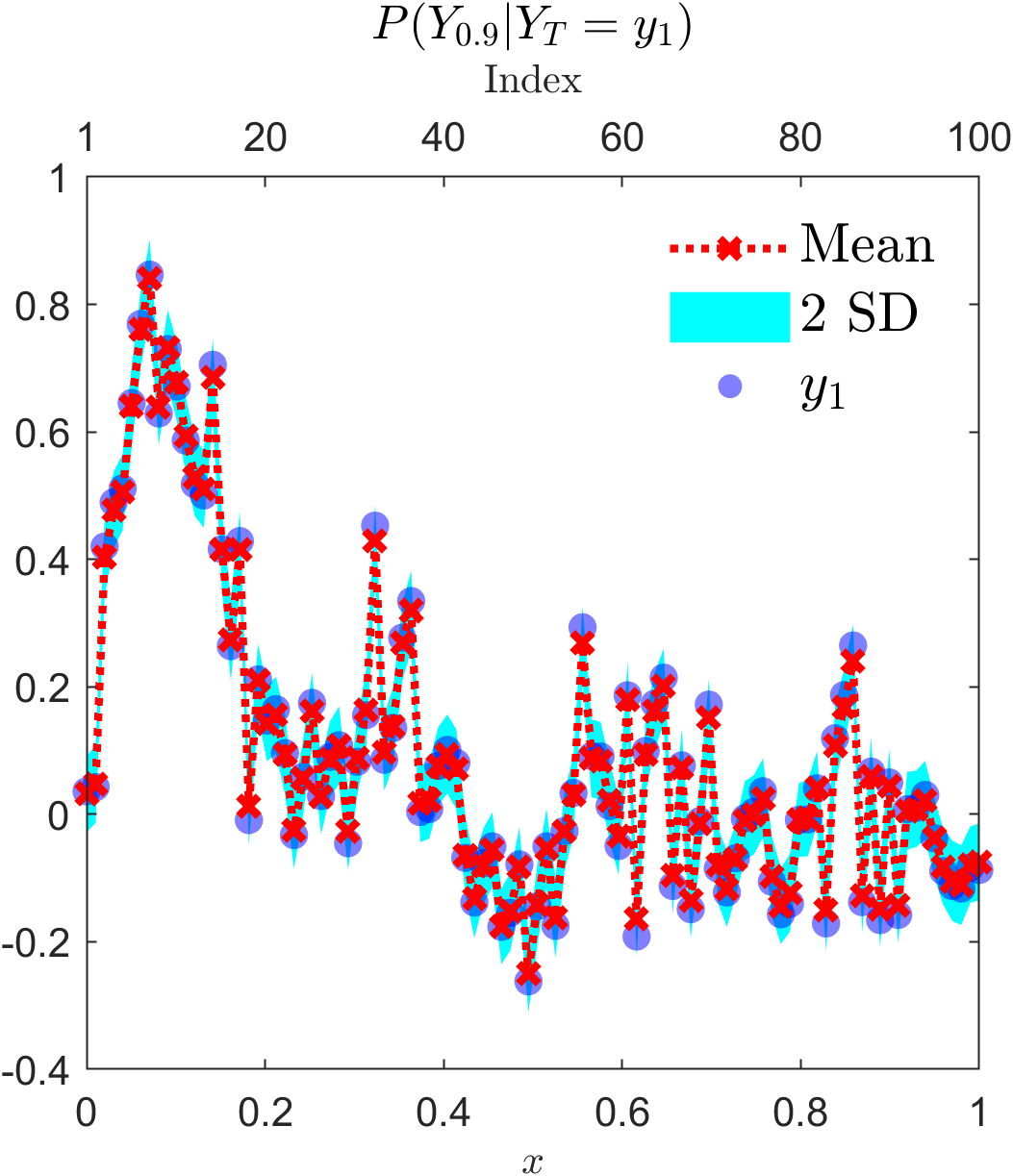}
         \includegraphics[width=0.18\textwidth]{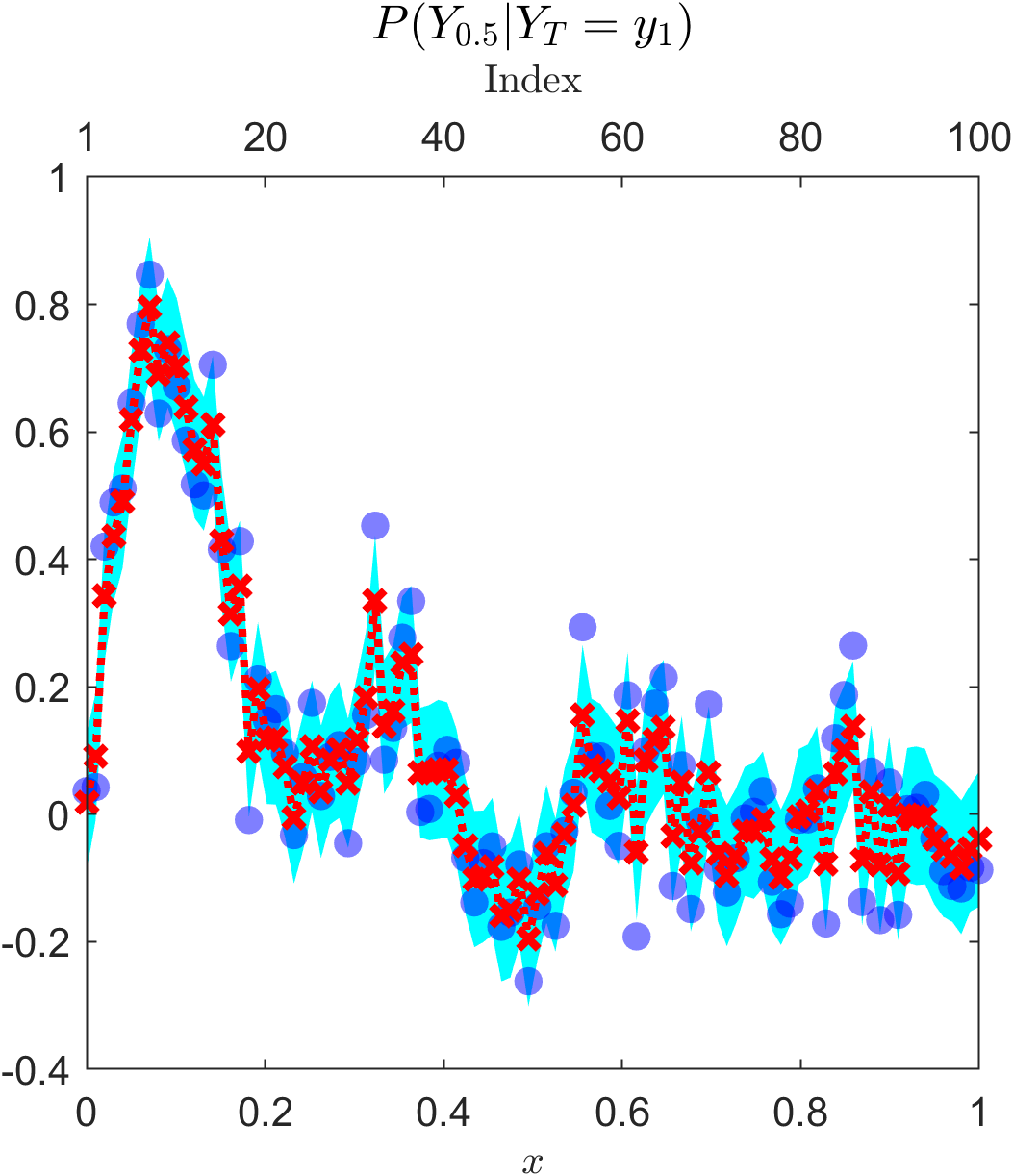}
         \includegraphics[width=0.18\textwidth]{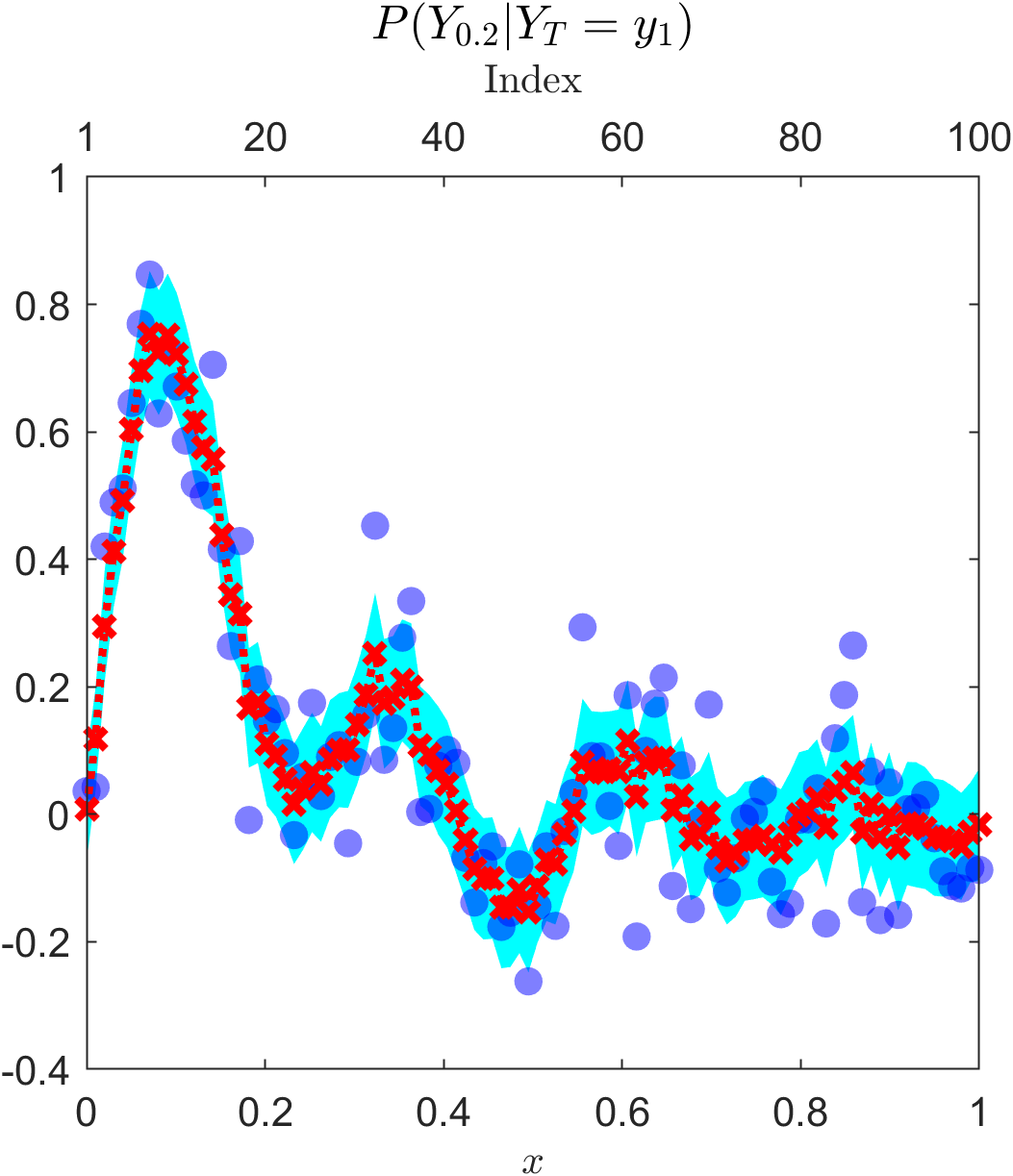}
         \includegraphics[width=0.18\textwidth]{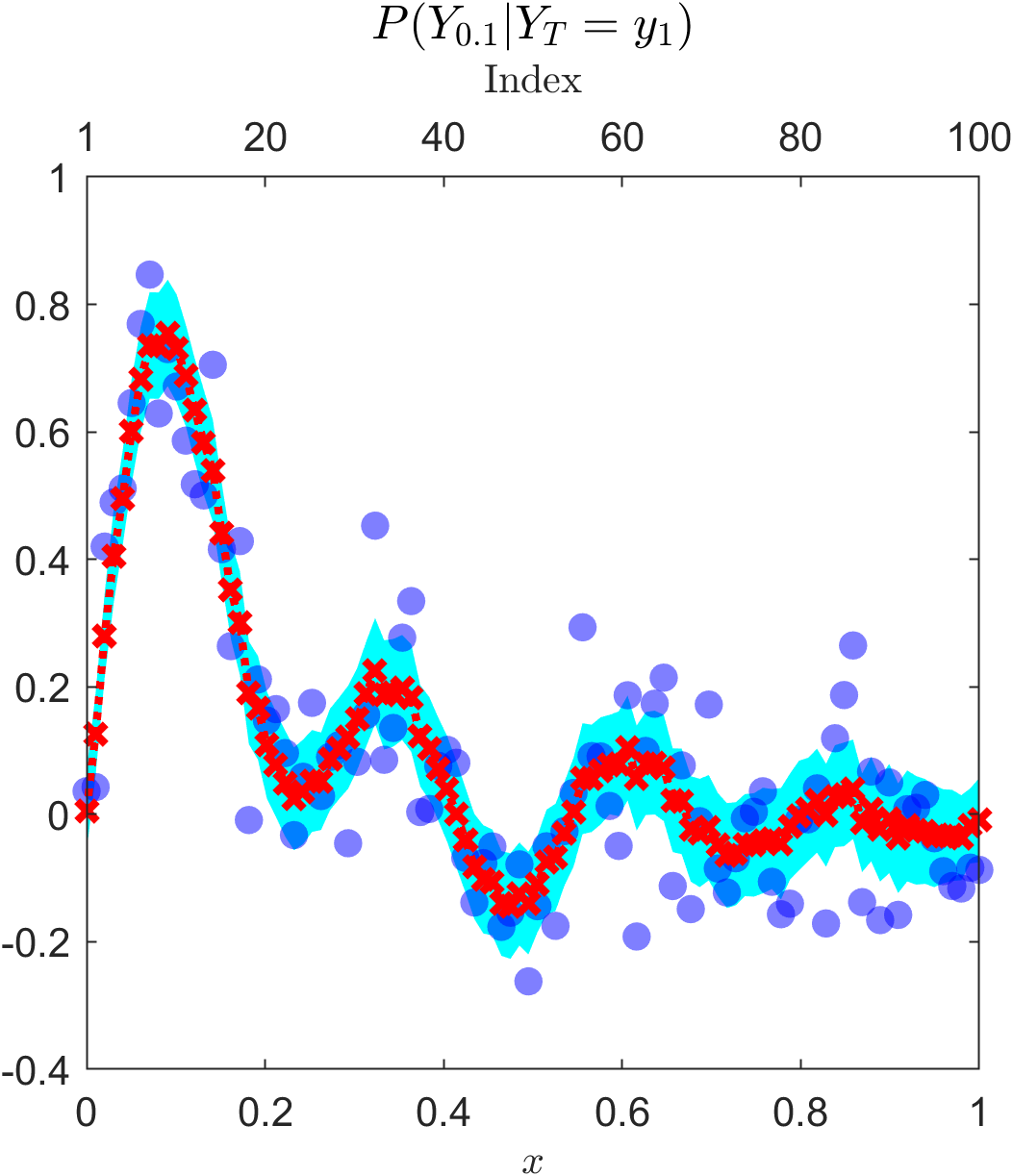}
         \includegraphics[width=0.18\textwidth]{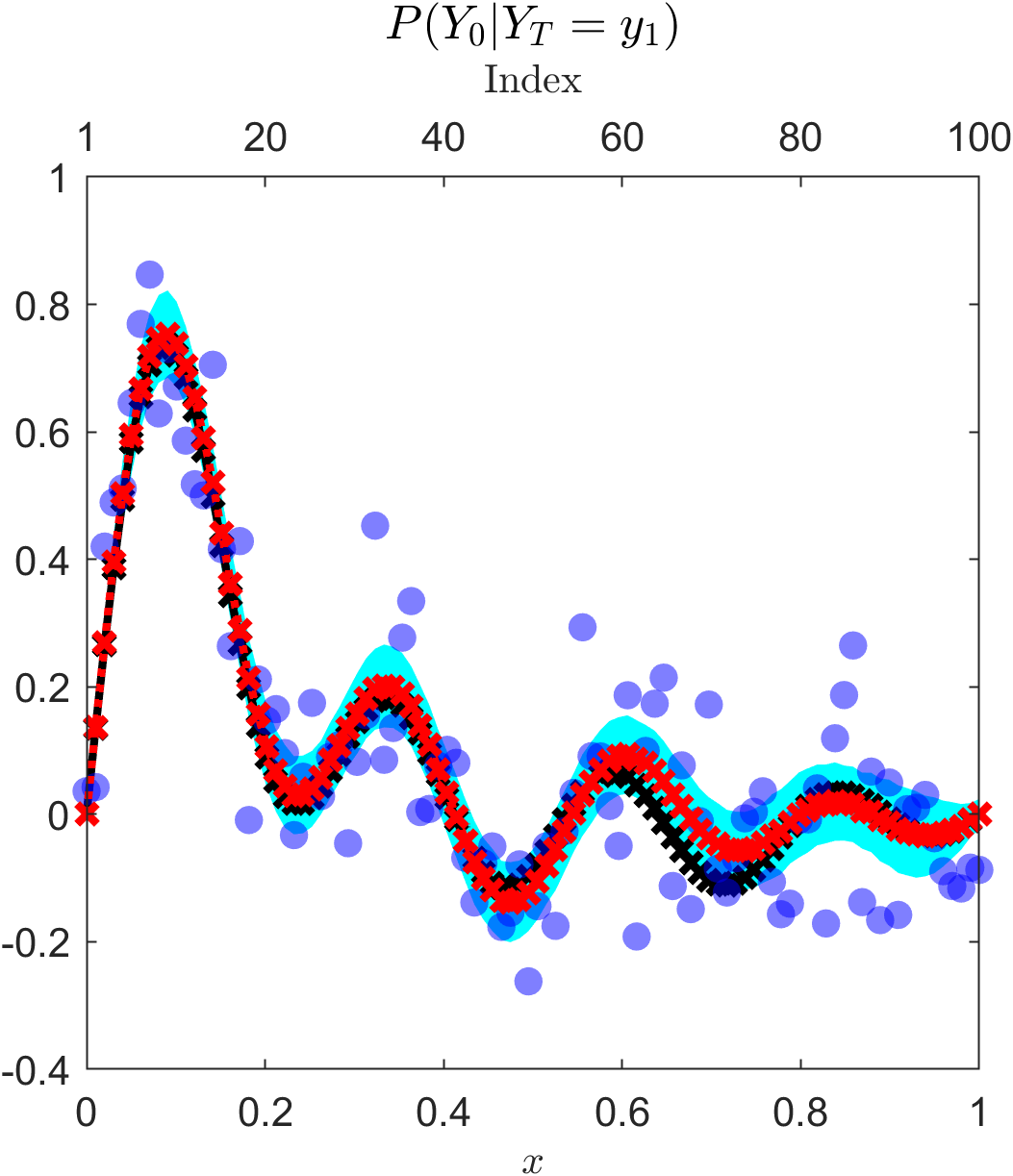}
         \caption{Posterior samples of $Y_t, t\in[0, T)$ given $Y_T=y_1$. }
    \end{subfigure}
    \begin{subfigure}[b]{1\textwidth}
         \centering
         \includegraphics[width=0.18\textwidth]{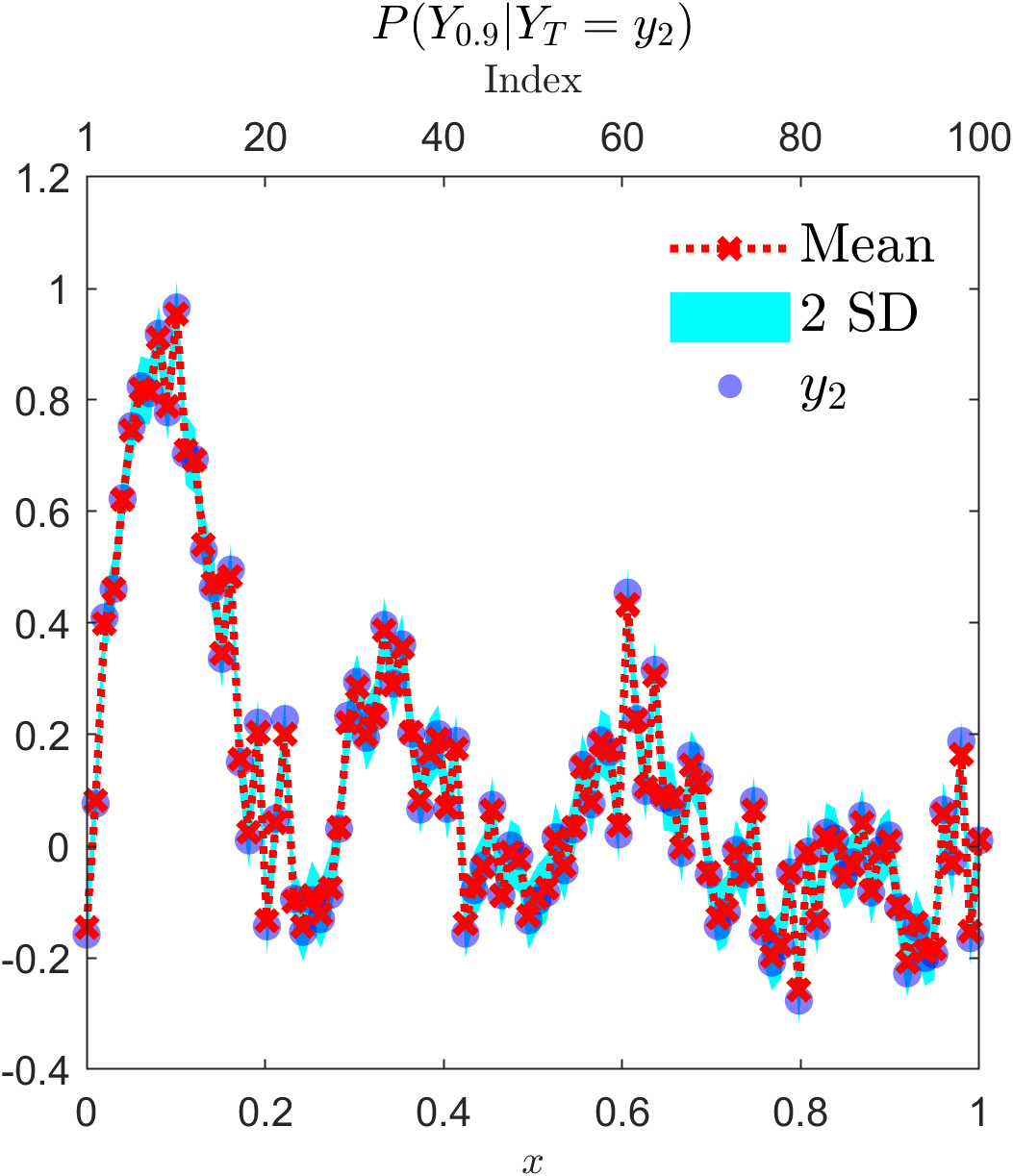}
         \includegraphics[width=0.18\textwidth]{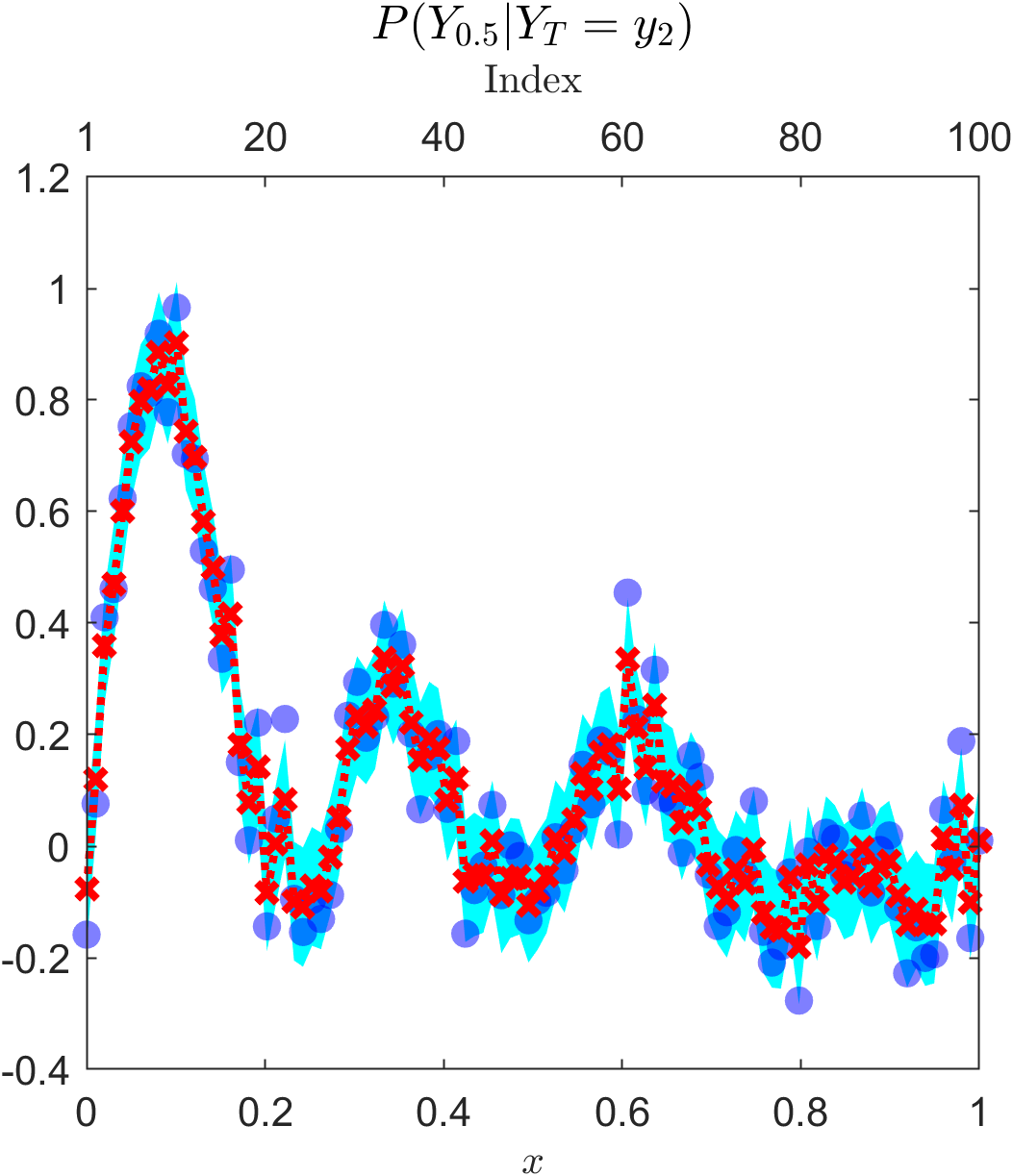}
         \includegraphics[width=0.18\textwidth]{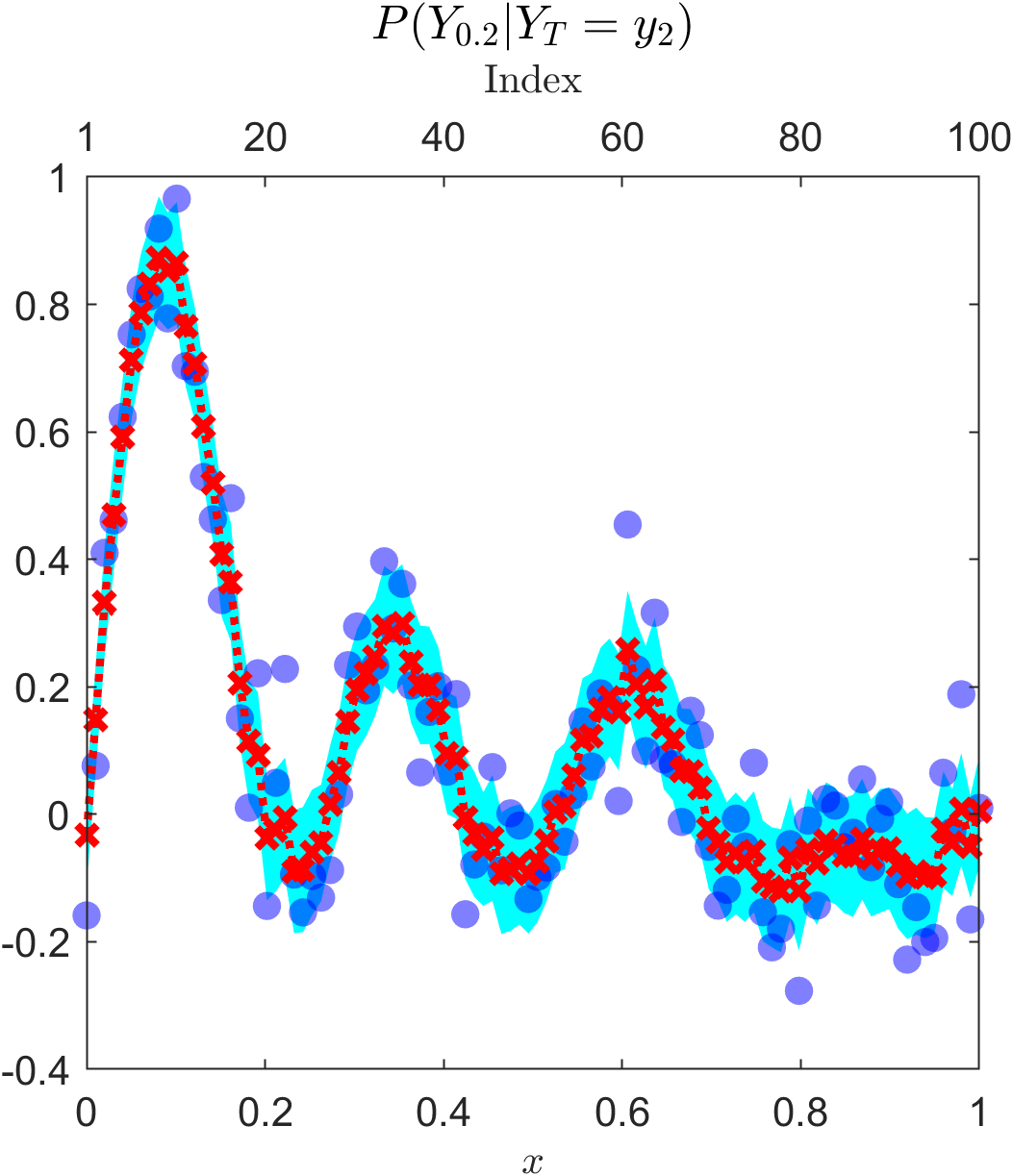}
         \includegraphics[width=0.18\textwidth]{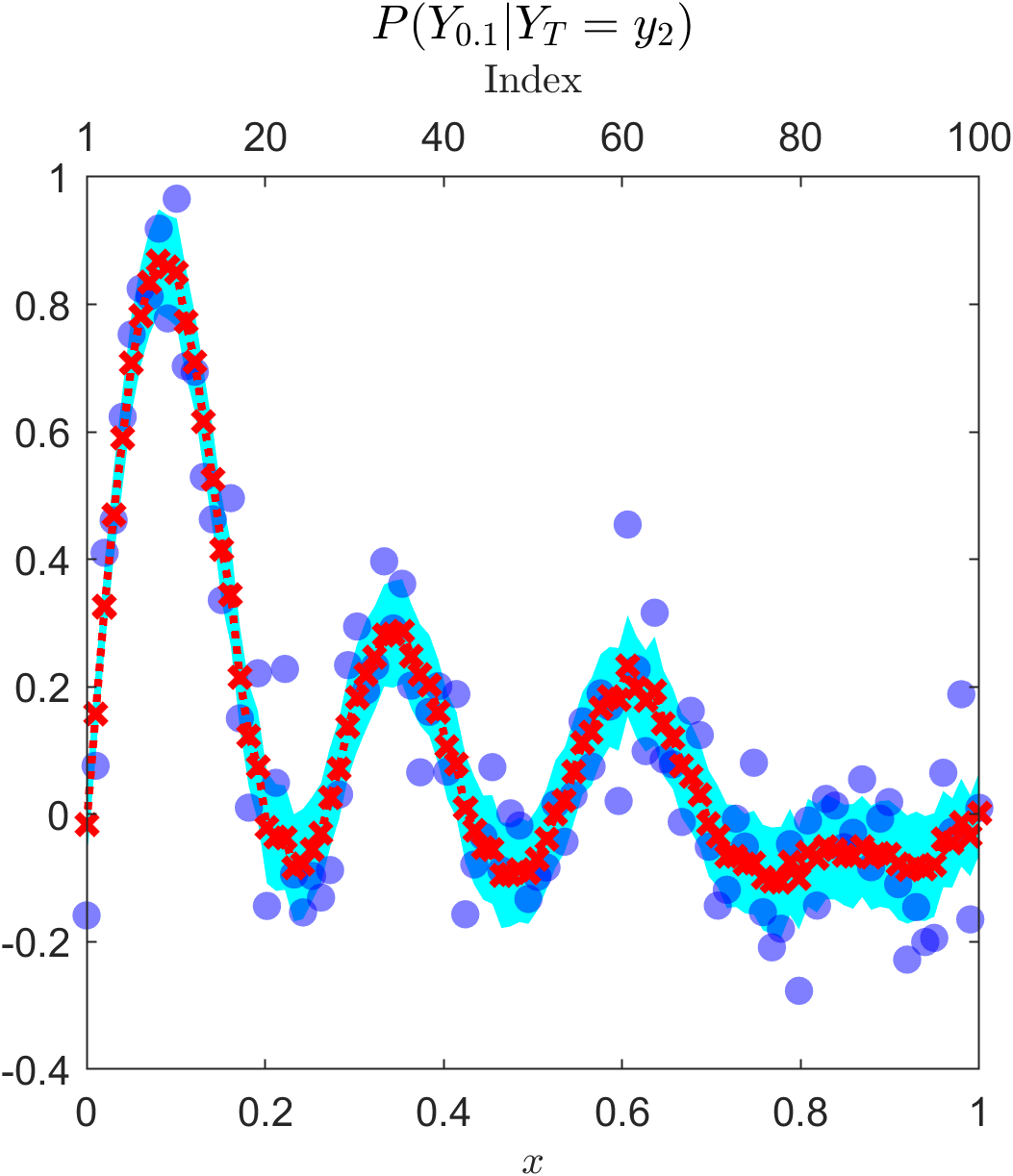}
         \includegraphics[width=0.18\textwidth]{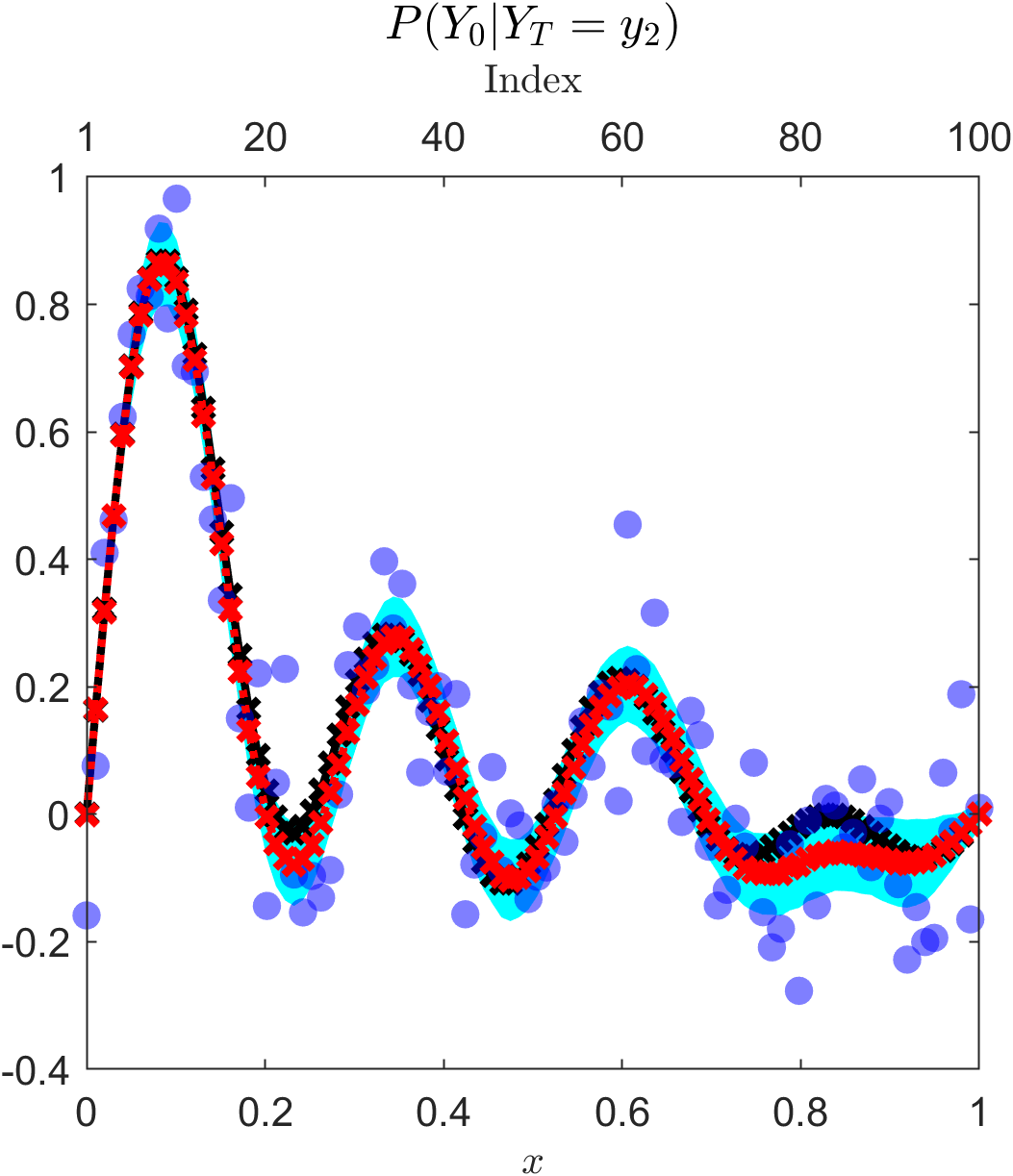}
         \caption{Posterior samples of $Y_t, t\in[0, T)$ given $Y_T=y_2$. }
    \end{subfigure}
\caption{
Results from the SGM-HJ-sampler used to infer \(Y_t\) for \(t \in [0, T)\) given \(Y_T = y_1\) or \(y_2\). Each figure shows a snapshot at a different time \(t\). The horizontal axis represents the spatial domain of the underlying function \(f_t\), while the vertical axis represents the function's value. Since \(Y_t \in \mathbb{R}^n\) with \(n = 100\) denotes the discretization of \(f_t\), we plot these values at the grid points in each figure. Figures (a) and (b) correspond to the cases with observations \(Y_T = y_1\) and \(Y_T = y_2\), respectively, where \(y_1\) and \(y_2\) are vectors randomly selected from the testing data of \(Y_T\). The component values of \(y_1\) and \(y_2\) are marked as {\color{blue}\textbf{blue}} circles. The {\color{red}\textbf{red}} crosses represent the posterior means of the inferred \(Y_t\) components, connected by red dots to illustrate the inferred underlying function \(f_t\). The colored region (\textcolor{myCyan}{$\blacksquare$}) indicates the uncertainty, computed as the mean \(\pm\) two standard deviations. In the rightmost figures, the \textbf{black} lines and crosses show the reference values of \(f_0\), the underlying unknown function used to generate the observation data \(Y_T\).
}
\label{fig:example_4}
\end{figure}

This section presents an example where the proposed algorithm is applied to a Bayesian inference problem for a function \( f_t \) defined on \([0, 1]\). The domain is discretized using a uniform grid, and we focus on the grid points. The values of \( f_t(x_i) \) at each grid point \( x_i \) can be represented as an \(n\)-dimensional vector, where \(n\) is the number of grid points. 
We denote this vector, with components \( f_t(x_1), \dots, f_t(x_n) \), as \( Y_t \). 
The underlying stochastic process for \(Y_t\) is modeled as a scaled Brownian motion, \(dY_t = \sqrt{\epsilon}dW_t, t \in [0, T]\). The objective is to solve the inverse problem of inferring the value of \(Y_t, t \in [0, T)\), given an observation of \(Y_T\) and prior knowledge of \(Y_0\). This example also evaluates the algorithm’s capability to handle high-dimensional problems.
The prior information assumes that \(Y_0\) corresponds to the grid values of the function \( f_0 \)~\cite{meng2022learning, zou2023hydra}:
\begin{equation}\label{eq:high_d}
    f_0(x\mid \xi_1, \dots, \xi_8) = \frac{1}{16} \sum_{j=1}^8 \xi_j\sin(j\pi x), \quad x\in[0, 1],
\end{equation}
where \(\xi_j, j=1,...,8\) are i.i.d. random variables uniformly distributed on \([1, 3)\).

We set \(n = 100\), \(\epsilon = 0.01\), and \(T = 1\). 
The SGM-HJ-sampler is used to solve the problem, utilizing the sliced score matching loss function~\eqref{eqt:slice_sgm_loss}, which enhances scalability when direct computation of the divergence in the loss~\eqref{eqt:sgm_loss} becomes inefficient.
The training data for \(s_\NNparam\) are generated by sampling \(Y_0\) from~\eqref{eq:high_d} and numerically discretizing the Brownian motion with \(\Delta t = 0.02\). The inference step is carried out with \(\Delta \tau = 0.01\). We generate posterior samples of \(Y_t, t \in [0, T)\), given two specific values of \(Y_T\), randomly selected from the test set. The results are shown in Figure~\ref{fig:example_4}, where we depict the posterior mean (red cross) and uncertainty (two standard deviation intervals, colored areas) of \(Y_t\) for certain \(t \in [0, T)\), based on 1,000 posterior samples from the SGM-HJ-sampler. Unlike Figures~\ref{fig:example_2_3} and~\ref{fig:example_3}, the x-axis in these figures represents the spatial domain \([0, 1]\) for the underlying function \(f_t\), rather than the temporal domain \(t \in [0, T]\). Each figure is a snapshot for a specific \(t\), and since \(Y_t\) contains the grid values of an underlying 1D function, we visualize the results by connecting the component values with lines to represent the reference, posterior mean, and uncertainties.

The results indicate that when \(t\) is close to \(T\), such as \(t = 0.9\), the uncertainty is small, and the posterior mean of \(Y_t\) is close to the observed value \(Y_T\). As \(t\) decreases, the posterior mean becomes smoother. In both cases, when \(t = 0\), the reference values (black lines and crosses) lie within the two-standard-deviation regions around the posterior mean, demonstrating the effectiveness of the SGM-HJ-sampler in quantifying uncertainty for this high-dimensional problem.

\section{Summary}\label{sec:summary}

In this paper, we leverage the log transform to establish connections between certain Bayesian inference problems, stochastic optimal control, and diffusion models. Building on this connection, we propose the HJ-sampler, an algorithm designed to sample from the posterior distribution of inverse problems involving SDE processes. 
We have developed three specific HJ-samplers: the analytic-HJ-sampler, Riccati-HJ-sampler, and SGM-HJ-sampler, applying them to various SDEs and different prior distributions of the initial condition. Notably, we have demonstrated the potential of these algorithms in addressing (1) uncertainty induced by model misspecification \cite{zou2024correcting} in nonlinear ODEs, (2) mixtures of certainty and uncertainty in ODE systems, and (3) high dimensionality. The results showcase the accuracy, flexibility, and generalizability of our approach, highlighting new avenues for solving such inverse problems by utilizing techniques originally developed for control problems and diffusion models. Despite these advancements, several open problems and extensions remain to be explored.

There are multiple ways to generalize the current method. First, although the method is initially applied to a single observational data point, it can be extended sequentially to handle multiple observations. After obtaining posterior samples from several observation points, the prior distribution can be updated, and the HJ-sampler can be reapplied to the updated prior as new observations become available. When employing the machine learning-based version of the HJ-sampler, this process can be integrated with operator learning to potentially eliminate the need for retraining neural networks, enabling more efficient updates. Second, while the HJ-sampler is tailored for cases where \(Y_t\) is governed by an SDE, the underlying log transform framework is applicable to any process with a well-defined infinitesimal generator. The SDE case represents a specific instance, and future research will focus on extending the numerical implementation of the HJ-sampler to more general processes. This broader perspective also hints at potential extensions of diffusion models to handle non-continuous distributions or processes driven by more complex noise structures.

Beyond generalization, several improvements to the current HJ-sampler method are possible. For instance, alternative numerical methods for solving viscous HJ PDEs or stochastic optimal control problems could be incorporated, leading to new variants of the HJ-sampler. Additionally, various machine learning techniques could be integrated with the HJ-sampler to enhance efficiency and accuracy. This could involve the development of novel loss functions or improved strategies for generating training data. Moreover, in cases where certain parts of the model are poorly understood or lack sufficient information, neural network surrogate models could be used to shift from a model-driven to a data-driven approach. For example, if the underlying process is not well-characterized, NeuralODE \cite{chen2018neural}, NeuralSDE \cite{kidger2021neural, liu2019neural}, or reinforcement learning methods could be employed to approximate the dynamics.

\section*{Acknowledgement}

We would like to express our gratitude to Dr. Molei Tao for providing valuable references on Sequential Monte Carlo (SMC) methods, and to Dr. Arnaud Doucet for contributing the idea presented in the second paragraph of Remark~\ref{rem:similarity_SGM}. T.M. is supported by ONR MURI N00014-20-1-2787. Z.Z., J.D., and G.E.K. are supported by the MURI/AFOSR FA9550-20-1-0358 project and the DOE-MMICS SEA-CROGS DE-SC0023191 award.

\bibliographystyle{siam}
\bibliography{references}

\appendix
\def\appendixcontrol{\alpha}
\def\LD{L_{\partial \Dinsde}}

\section{Log transform applied to a general SDE}\label{appendix:log_sde}
We consider the stochastic process $X_t$, represented as an SDE in $\Rn$, defined by $dX_t = b(X_t,t)dt + \sqrt{\epsilon} \sigma(X_t,t)dW_t$. Here, $W_t$ denotes a Brownian motion in $\R^m$. The functions $b: \Rn \times [0,T] \to \Rn$ and $\sigma: \Rn \times [0,T] \to \R^{n \times m}$ ensure the SDE's proper formulation. 
We introduce $\Dinsde: \Rn \times [0,T] \to \R^{n \times n}$ defined by $\Dinsde(x,t) = \sigma(x,t)\sigma(x,t)^T$ for any $x \in \Rn$ and $t \in [0,T]$. 
The constant \(\epsilon\) in the SDE is a positive value that indicates the level of stochasticity and corresponds to the hyperparameter \(\epsilon\) in the general definition of the operator $\epsgeneratorfwd$ in Section~\ref{sec:log_transform}. The specific formulation of $\epsgeneratorfwd$ in this SDE context and its adjoint $\epsgeneratorfwd^*$ are described by the equations:
\begin{equation}
\epsgeneratorfwd f = b(x,t) \cdot \nabla_x f + \frac{\epsilon}{2} \Tr(\Dinsde(x,t) \nabla_x^2 f), \quad
\epsgeneratorfwd^* f = -\nabla_x \cdot (b(x,t) f) + \frac{\epsilon}{2} \sum_{i,j=1}^n \frac{\partial^2 (\Dinsde_{ij}(x,t) f)}{\partial x_i \partial x_j},
\end{equation}
with $\Dinsde_{ij}(x,t)$ representing the $(i,j)$-th component of the matrix $\Dinsde(x,t)$.
In this appendix, if the variables in a function or a formula are not explicitly specified, they default to $(x,t)$.

The KBE and KFE in~\eqref{eqt:forward_backward_Kolmogorov} are given by:
\begin{equation}
\partial_t \mubwd = b(x,T-t) \cdot \nabla_x \mubwd + \frac{\epsilon}{2} \Tr(\Dinsde(x,T-t) \nabla_x^2 \mubwd), \quad
\partial_t \mufwd = -\nabla_x \cdot (b(x,t) \mufwd) + \frac{\epsilon}{2} \sum_{i,j=1}^n \frac{\partial^2 (\Dinsde_{ij}(x,t) \mufwd)}{\partial x_i \partial x_j},
\end{equation}
with all calculations based on the variables $(x,t)$.
The coupled nonlinear differential equations in~\eqref{eqt:coupled_PDE_fwd} are:
\begin{equation}
\begin{dcases}
\partial_t \rho +\nabla_x \cdot((b(x,t) + \Dinsde(x,t)\nabla_x \Sepsbwd) \rho) = \frac{\epsilon}{2} \sum_{i,j=1}^n \partial_{x_i}\partial_{x_j}(\rho \Dinsde_{ij}(x,t)), \\
\partial_t \Sepsbwd + b(x,t)\cdot \nabla_x \Sepsbwd + \frac{1}{2} (\nabla_x \Sepsbwd)^T \Dinsde(x,t) \nabla_x \Sepsbwd + \frac{\epsilon}{2}\Tr(\Dinsde(x,t) \nabla_x^2 \Sepsbwd) = 0,
\end{dcases}
\end{equation}
where calculations and functions are evaluated at $(x,t)$.

With terminal condition $-J$ on $\Sepsbwd$ and initial condition $\delta_{z_0}$ on $\rho$, these equations form the first order conditions for optimality in the following stochastic optimal control problem:
\begin{equation}
\min\left\{ \E\left[\int_0^T \frac{1}{2}v_s^T D(Z_s,s)^{-1}v_s ds + J(Z_T)\right]\colon dZ_s = (b(Z_s, s) + v_s)ds + \sqrt{\epsilon}\sigma(Z_s,s) dW_s, Z_0 = z_0\right\},
\end{equation}
whose value equals $-\Sepsbwd(z_0, 0)$.
Alterations in initial or terminal conditions for $\rho$ and $\Sepsbwd$ relate these differential equations to specific MFGs or SOT problems, as discussed previously in Example~\ref{eg:log_transform_BM}. 
Details of the computations are provided in the following section.

\subsection{Computational details}
By the definition of $\epsgeneratorfwd$, we have
\begin{equation}
\begin{split}
e^{-\frac{\Sepsbwd}{\epsilon}}\epsgeneratorfwd e^{\frac{\Sepsbwd}{\epsilon}}
&= e^{-\frac{\Sepsbwd}{\epsilon}} \left(b(x,t)\cdot \nabla_x e^{\frac{\Sepsbwd}{\epsilon}} + \frac{\epsilon}{2}\sum_{i,j=1}^n \Dinsde_{ij}(x,t)\partial_{x_i}\partial_{x_j} e^{\frac{\Sepsbwd}{\epsilon}}\right)\\
&= e^{-\frac{\Sepsbwd}{\epsilon}} \left(\frac{e^{\frac{\Sepsbwd}{\epsilon}}}{\epsilon} b(x,t)\cdot \nabla_x \Sepsbwd + \frac{\epsilon}{2}\sum_{i,j=1}^n \Dinsde_{ij}(x,t) e^{\frac{\Sepsbwd}{\epsilon}} \left(\frac{(\partial_{x_i}\Sepsbwd)(\partial_{x_j}\Sepsbwd)}{\epsilon^2} + \frac{\partial_{x_i}\partial_{x_j} \Sepsbwd}{\epsilon} \right)\right)\\
&= \frac{b(x,t)\cdot \nabla_x \Sepsbwd}{\epsilon} + \frac{1}{2\epsilon}\sum_{i,j=1}^n \Dinsde_{ij}(x,t) (\partial_{x_i}\Sepsbwd)(\partial_{x_j}\Sepsbwd) + \frac{1}{2}\sum_{i,j=1}^n \Dinsde_{ij}(x,t) (\partial_{x_i}\partial_{x_j} \Sepsbwd)\\
&= \frac{b(x,t)\cdot \nabla_x \Sepsbwd}{\epsilon} + \frac{1}{2\epsilon} (\nabla_x \Sepsbwd)^T \Dinsde(x,t) \nabla_x \Sepsbwd + \frac{1}{2}\Tr(\Dinsde(x,t) \nabla_x^2 \Sepsbwd).
\end{split}
\end{equation}
Therefore, the HJ equation in~\eqref{eqt:coupled_PDE_fwd} becomes
\begin{equation}
0 = \partial_t \Sepsbwd + \epsilon e^{-\frac{\Sepsbwd}{\epsilon}}\epsgeneratorfwd e^{\frac{\Sepsbwd}{\epsilon}}
= \partial_t \Sepsbwd + b(x,t)\cdot \nabla_x \Sepsbwd + \frac{1}{2} (\nabla_x \Sepsbwd)^T \Dinsde(x,t) \nabla_x \Sepsbwd + \frac{\epsilon}{2}\Tr(\Dinsde(x,t) \nabla_x^2 \Sepsbwd).
\end{equation}
Similarly, according to the formula for $\epsgeneratorfwd^*$, we get
\begin{equation}
\begin{split}
e^{\frac{\Sepsbwd}{\epsilon}}\epsgeneratorfwd^* (\rho e^{-\frac{\Sepsbwd}{\epsilon}})
&= e^{\frac{\Sepsbwd}{\epsilon}} \left( -\nabla_x \cdot (b(x,t) \rho e^{-\frac{\Sepsbwd}{\epsilon}}) + \frac{\epsilon}{2} \sum_{i,j=1}^n \frac{\partial^2 (\Dinsde_{ij}(x,t) \rho e^{-\frac{\Sepsbwd}{\epsilon}})}{\partial x_i \partial x_j}\right)\\
&= e^{\frac{\Sepsbwd}{\epsilon}} \Bigg( -e^{-\frac{\Sepsbwd}{\epsilon}} \nabla_x \cdot (b(x,t)\rho) + \rho e^{-\frac{\Sepsbwd}{\epsilon}} b(x,t)\cdot \frac{\nabla_x \Sepsbwd}{\epsilon} + \frac{\epsilon}{2} \sum_{i,j=1}^n \Big(e^{-\frac{\Sepsbwd}{\epsilon}} \partial_{x_i}\partial_{x_j}(\rho \Dinsde_{ij}(x,t)) \\
&\quad\quad\quad\quad - \frac{2}{\epsilon}e^{-\frac{\Sepsbwd}{\epsilon}}(\partial_{x_i} \Sepsbwd)(\partial_{x_j} (\rho \Dinsde_{ij}(x,t))) + \rho \Dinsde_{ij}(x,t) e^{-\frac{\Sepsbwd}{\epsilon}} \Big(\frac{(\partial_{x_i}\Sepsbwd) (\partial_{x_j}\Sepsbwd)}{\epsilon^2} - \frac{\partial_{x_i}\partial_{x_j}\Sepsbwd}{\epsilon}\Big) \Big)\Bigg)\\
&= -\nabla_x \cdot (b(x,t)\rho) + \rho b(x,t)\cdot \frac{\nabla_x \Sepsbwd}{\epsilon} + \frac{\epsilon}{2} \sum_{i,j=1}^n \Big( \partial_{x_i}\partial_{x_j}(\rho \Dinsde_{ij}(x,t)) \\
&\quad\quad\quad\quad - \frac{2}{\epsilon}(\partial_{x_i} \Sepsbwd)(\partial_{x_j} (\rho \Dinsde_{ij}(x,t))) + \rho \Dinsde_{ij}(x,t) \Big(\frac{(\partial_{x_i}\Sepsbwd) (\partial_{x_j}\Sepsbwd)}{\epsilon^2} - \frac{\partial_{x_i}\partial_{x_j}\Sepsbwd}{\epsilon}\Big) \Big).
\end{split}
\end{equation}
Then, we have
\begin{equation}
\begin{adjustbox}{width=0.99\textwidth}$
\begin{split}
\rho e^{-\frac{\Sepsbwd}{\epsilon}}\epsgeneratorfwd e^{\frac{\Sepsbwd}{\epsilon}} - e^{\frac{\Sepsbwd}{\epsilon}}\epsgeneratorfwd^* (\rho e^{-\frac{\Sepsbwd}{\epsilon}})
&= \frac{\rho b(x,t)\cdot \nabla_x \Sepsbwd}{\epsilon} + \frac{\rho }{2\epsilon} (\nabla_x \Sepsbwd)^T \Dinsde(x,t) \nabla_x \Sepsbwd + \frac{\rho}{2}\Tr(\Dinsde(x,t) \nabla_x^2 \Sepsbwd) 
+\nabla_x \cdot (b(x,t)\rho) \\
&\quad\quad - \rho b(x,t)\cdot \frac{\nabla_x \Sepsbwd}{\epsilon} - \frac{\epsilon}{2} \sum_{i,j=1}^n \Big( \partial_{x_i}\partial_{x_j}(\rho \Dinsde_{ij}(x,t)) - \frac{2}{\epsilon}(\partial_{x_i} \Sepsbwd)(\partial_{x_j} (\rho \Dinsde_{ij}(x,t))) \\
&\quad\quad + \rho \Dinsde_{ij}(x,t) \Big(\frac{(\partial_{x_i}\Sepsbwd) (\partial_{x_j}\Sepsbwd)}{\epsilon^2} - \frac{\partial_{x_i}\partial_{x_j}\Sepsbwd}{\epsilon}\Big) \Big)\\
&= \rho \Tr(\Dinsde(x,t) \nabla_x^2 \Sepsbwd) 
+\nabla_x \cdot (b(x,t)\rho) - \frac{\epsilon}{2} \sum_{i,j=1}^n \partial_{x_i}\partial_{x_j}(\rho \Dinsde_{ij}(x,t)) + \sum_{i,j=1}^n (\partial_{x_i} \Sepsbwd)(\partial_{x_j} (\rho \Dinsde_{ij}(x,t)))\\
&= \nabla_x \cdot((b(x,t) + \Dinsde(x,t) \nabla_x \Sepsbwd) \rho) - \frac{\epsilon}{2} \sum_{i,j=1}^n \partial_{x_i}\partial_{x_j}(\rho \Dinsde_{ij}(x,t)).
\end{split}
$\end{adjustbox}
\end{equation}
Therefore, the Fokker-Planck equation in~\eqref{eqt:coupled_PDE_fwd} becomes
\begin{equation}
0 = \partial_t \rho + \rho e^{-\frac{\Sepsbwd}{\epsilon}}\epsgeneratorfwd e^{\frac{\Sepsbwd}{\epsilon}} - e^{\frac{\Sepsbwd}{\epsilon}}\epsgeneratorfwd^* (\rho e^{-\frac{\Sepsbwd}{\epsilon}}) = \partial_t \rho +\nabla_x \cdot((b(x,t) + \Dinsde(x,t)\nabla_x \Sepsbwd) \rho) - \frac{\epsilon}{2} \sum_{i,j=1}^n \partial_{x_i}\partial_{x_j}(\rho \Dinsde_{ij}(x,t)).
\end{equation}

\section{The log transform for a one-dimensional scaled Poisson process}\label{appendix:log_Poisson}
In this section, we consider a stochastic process defined as a one-dimensional scaled Poisson process, namely $X_t = \epsilon N(t)$, where $N(t)$ denotes a Poisson process with a rate of $\lambda = \frac{1}{\epsilon}$.
Similar to the approach in Section~\ref{sec:log_Xt}, we define the linear operator $\epsgeneratorfwd$ as the infinitesimal generator of $X_t$. Consequently, the operator $\epsgeneratorfwd$ and its adjoint, $\epsgeneratorfwd^*$, are represented by:
\begin{equation*}
\epsgeneratorfwd f = \frac{f(x + \epsilon, t) - f(x, t)}{\epsilon}, \quad\quad \epsgeneratorfwd^* f = \frac{f(x - \epsilon, t) - f(x,t)}{\epsilon}.
\end{equation*}
This results in specific adaptations of the general coupled linear system~\eqref{eqt:forward_backward_Kolmogorov} and the general coupled nonlinear system~\eqref{eqt:coupled_PDE_fwd} to the case of the scaled Poisson process, represented by the following equations:
    \begin{equation}
    \partial_t \mubwd(x,t) = \frac{\mubwd(x + \epsilon, t) - \mubwd(x, t)}{\epsilon},\quad\quad 
    \partial_t \mufwd(x,t) = \frac{\mufwd(x-\epsilon, t) - \mufwd(x, t)}{\epsilon},
    \end{equation}
for the linear system, and
    \begin{equation}
    \begin{dcases}
    0 = \partial_t \rho + \rho e^{-\frac{\Sepsbwd}{\epsilon}}\epsgeneratorfwd e^{\frac{\Sepsbwd}{\epsilon}} - e^{\frac{\Sepsbwd}{\epsilon}}\epsgeneratorfwd^* (\rho e^{-\frac{\Sepsbwd}{\epsilon}}) = \partial_t \rho(x,t) + \frac{1}{\epsilon} \left(e^{\frac{\Sepsbwd(x+\epsilon,t) - \Sepsbwd(x,t)}{\epsilon}}\rho(x,t) - \rho(x-\epsilon,t) e^{\frac{\Sepsbwd(x,t)- \Sepsbwd(x-\epsilon,t)}{\epsilon}}\right), \\
    0 = \partial_t \Sepsbwd + \epsilon e^{-\frac{\Sepsbwd}{\epsilon}}\epsgeneratorfwd e^{\frac{\Sepsbwd}{\epsilon}} = \partial_t \Sepsbwd(x,t) + \exp\left(\frac{\Sepsbwd(x + \epsilon, t) - \Sepsbwd(x,t)}{\epsilon}\right) -  1,
    \end{dcases}
    \end{equation}
for the nonlinear system.

With the terminal condition $-J$ on $\Sepsbwd$ and the initial condition $\delta_{z_0}$ on $\rho$, these PDEs serve as the first order optimality conditions for the subsequent stochastic optimal control problem:
\begin{equation}
\begin{split}
\min\Bigg\{ \int_0^T\int_{\R} \left(g(x,s) \log g(x,s) -g(x,s) + 1\right)\rho(x,s)dx ds + \int_\R J(x)\rho(x,T)dx \colon 
\\
\partial_t \rho(x,t) = \frac{g(x-\epsilon,t)\rho(x-\epsilon,t) - g(x,t)\rho(x,s)}{\epsilon}, \rho(x,0) = \delta_{z_0}(x)\Bigg\}.
\end{split}
\end{equation}
The optimal control function $g$ satisfies the relation $g(x,t) = \exp(\frac{\Sepsbwd(x+\epsilon,t) - \Sepsbwd(x,t)}{\epsilon})$. This represents a stochastic optimal control scenario where the control influences the jump rate of a process through the function $g$, and the running loss is derived from the entropy function of the rate $g$. As in the SDE context, varying the initial and terminal conditions on $\rho$ and $\Sepsbwd$ associates these two PDEs with specific MFGs and Schrödinger bridge problems, where the underlying stochastic process is a controlled jump process, given $X_t$ is a jump process. We defer exploration of more complicated jump processes to future research.



\section{HJ-sampler for SDE cases}\label{appendix:HJsampler_sde}

In this section, we focus on the scenario where the stochastic process $Y_t$ is described by the general SDE $dY_t = b(Y_t, t) \, dt + \sqrt{\epsilon} \sigma(Y_t, t) \, dW_t$, where $W_t$ is an $m$-dimensional Brownian motion. Here, $b \colon \mathbb{R}^n \times [0, T] \to \mathbb{R}^n$ and $\sigma \colon \mathbb{R}^n \times [0, T] \to \mathbb{R}^{n \times m}$ are functions ensuring that the SDE is well-defined. We define $\Dinsde(x,t) = \sigma(x,t) \sigma(x,t)^T$. 
Unless specified otherwise, the input variables in this appendix are assumed to be \((x, t)\).
The constant \(\epsilon\) in the SDE corresponds to the hyperparameter \(\epsilon\) in the general definition of the operator $\epsgeneratorinv$ in Section~\ref{sec:log_transform}.
In this case, the infinitesimal generator of $Y_t$ is defined as $f \mapsto b(x,t) \cdot \nabla_x f + \frac{\epsilon}{2} \operatorname{Tr}(\Dinsde(x,t) \nabla^2 f)$. 
Consequently, the linear operators $\epsgeneratorinv$ and its adjoint $\epsgeneratorinv^*$ are defined by
\begin{equation}
\epsgeneratorinv f = -\nabla_x \cdot (b(x,t) f) +  \frac{\epsilon}{2} \sum_{i,j=1}^n \frac{\partial^2 (\Dinsde_{ij}(x,t) f)}{\partial x_i \partial x_j}, \quad
\epsgeneratorinv^* f = b(x,t) \cdot \nabla_x f + \frac{\epsilon}{2} \operatorname{Tr}(\Dinsde(x,t) \nabla_x^2 f),
\end{equation}
where $\Dinsde_{ij}(x,t)$ denotes the $(i,j)$-th element of the matrix $\Dinsde(x,t)$. The HJ equation~\eqref{eqt:HJsampler_general_HJ} becomes
\begin{equation}\label{eqt:HJsampler_SDE_HJ_general}
\begin{dcases}
\partial_t \Sepsbwd  - b(x,T-t)\cdot \nabla_x \Sepsbwd  + \epsilon \sum_{i,j=1}^n (\partial_{x_i} \Sepsbwd) (\partial_{x_j} \Dinsde_{ij}(x,T-t)) + \frac{1}{2} (\nabla_x \Sepsbwd)^T \Dinsde(x,T-t) \nabla_x \Sepsbwd  \\
\quad\quad\quad\quad + \frac{\epsilon}{2}\Tr( \Dinsde(x,T-t) \nabla_x^2 \Sepsbwd) + \frac{\epsilon^2}{2} \sum_{i,j=1}^n \partial_{x_i}\partial_{x_j}\Dinsde_{ij}(x,T-t) -\epsilon \nabla_x \cdot b(x,T-t) = 0,\\
\Sepsbwd(x,T) = \epsilon\log P_{prior}(x),
\end{dcases}
\end{equation}
which is a traditional viscous HJ PDE.
The Fokker-Planck equation in~\eqref{eqt:HJsampler_general_rho} becomes
\begin{equation}\label{eqt:HJsampler_SDE_rho_general}
\begin{dcases}
\partial_t \rho -\nabla_x \cdot (b(x,T-t)\rho) +\sum_{i,j=1}^n \partial_{x_i} \left(\rho (\Dinsde_{ij}(x,T-t) \partial_{x_j}\Sepsbwd + \epsilon \partial_{x_j}\Dinsde_{ij}(x,T-t))\right) 
\\
\quad\quad\quad\quad - \frac{\epsilon}{2}\sum_{i,j=1}^n \partial_{x_i}\partial_{x_j}(\Dinsde_{ij}(x,T-t)\rho) =0, \\
\rho(x,0) = \delta_{\ydata}(x).
\end{dcases}
\end{equation}

Similar to the case in Appendix~\ref{appendix:log_sde}, these two PDEs are related to the following stochastic optimal control problem:
\begin{equation}
\begin{split}
\min\Bigg\{ \E\Bigg[\int_0^T \Bigg(\frac{1}{2}v_s^T D_{T-s}(Z_s)^{-1}v_s + \epsilon\nabla_x \cdot b_{T-s}(Z_s) - \frac{\epsilon^2}{2} \sum_{i,j=1}^n \partial_{x_i}\partial_{x_j}\Dinsde_{ij,T-s}(Z_s) \Bigg)ds -\epsilon\log \Pprior(Z_T)\Bigg]\colon \\
dZ_s = (-b(Z_s,T-s)+\epsilon \nabla_x \cdot \Dinsde(Z_s, T-s) + v_s)ds + \sqrt{\epsilon}\sigma(Z_s,T-s) dW_s, Z_0 = \ydata\Bigg\},
\end{split}
\end{equation}
where $\nabla_x \cdot \Dinsde(x,t)$ represents a vector-valued function whose $i$-th component is $\nabla_x \cdot \Dinsde_i(x,t)$, with $\Dinsde_i(x,t)$ being the $i$-th column of $\Dinsde(x,t)$.
To simplify notation, we denote $b(x,t)$ and $\Dinsde_{ij}(x,t)$ by $b_t(x)$ and $\Dinsde_{ij,t}(x)$, respectively.
In the scenario where $\Dinsde$ is non-invertible, the following optimal control problem arises:
\begin{equation}
\begin{split}
\min\Bigg\{ \E\Bigg[\int_0^T \Bigg(\frac{1}{2}u_s^T D_{T-s}(Z_s)u_s + \epsilon\nabla_x \cdot b_{T-s}(Z_s) - \frac{\epsilon^2}{2} \sum_{i,j=1}^n \partial_{x_i}\partial_{x_j}\Dinsde_{ij,T-s}(Z_s) \Bigg)ds -\epsilon\log \Pprior(Z_T)\Bigg]\colon \\
dZ_s = (-b_{T-s}(Z_s)+\epsilon \nabla_x \cdot \Dinsde(Z_s, T-s) + \Dinsde_{T-s}(Z_s) u_s)ds + \sqrt{\epsilon}\sigma_{T-s}(Z_s) dW_s, Z_0 = \ydata\Bigg\}.
\end{split}
\end{equation}

The proposed HJ-sampler algorithm consists of two main steps:
\begin{enumerate}
    \item Numerically solve the viscous HJ PDE~\eqref{eqt:HJsampler_SDE_HJ_general} to obtain $\Sepsbwd$;
    \item Generate samples from $\rho$, which satisfies~\eqref{eqt:HJsampler_SDE_rho_general}, by sampling from the controlled SDE given below:
\begin{equation}\label{eqt:appendixC_controlledsde_cont}
\begin{split}
dZ_t &= \left(\Dinsde(Z_t,T-t)\nabla_x \Sepsbwd(Z_t, t) -b(Z_t,T-t) + \epsilon \sum_{j=1}^n \partial_{x_j} \Dinsde_j(Z_t,T-t)\right) dt + \sqrt{\epsilon}\sigma(Z_t,T-t)dW_t, \\
Z_0 &= \ydata,
\end{split}
\end{equation}
where $\Dinsde_j(x,t)$ denotes the $j$-th column of the matrix $\Dinsde(x,t)$. In practice, the Euler–Maruyama method is employed to discretize this SDE as follows:
\begin{equation}
\begin{adjustbox}{width=0.9\textwidth}$
\begin{split}
Z_{k+1} &= Z_k + \left(\Dinsde(Z_k,T-t_k)\nabla_x \Sepsbwd(Z_k, t_k) -b(Z_k,T-t_k) + \epsilon \sum_{j=1}^n \partial_{x_j} \Dinsde_j(Z_k,T-t_k)\right) \Delta t + \sqrt{\epsilon \Delta t} \sigma(Z_k,T-t_k) \xi_k, \\
Z_0 &= \ydata,
\end{split}
$\end{adjustbox}
\end{equation}
where $\xi_0, \dots, \xi_{n_t-1}$ are i.i.d. samples drawn from the $m$-dimensional standard normal distribution.
\end{enumerate}

The obtained samples for $Z_k$ provide an approximation to the posterior samples of $Y_{T - k\Delta t}$ given $Y_T = \ydata$.
Computational details are presented in the subsequent section.

\subsection{Computational details}
In this section, we provide the computational details for the results provided in Section~\ref{sec:HJsampler_sde}. 
By the definition of $\epsgeneratorinv$, we have
\begin{equation}
\begin{split}
e^{-\frac{\Sepsbwd}{\epsilon}}\epsgeneratorfwd e^{\frac{\Sepsbwd}{\epsilon}}
&= e^{-\frac{\Sepsbwd}{\epsilon}} \left(-\nabla_x \cdot(b(x,T-t)e^{\frac{\Sepsbwd}{\epsilon}}) + \frac{\epsilon}{2}\sum_{i,j=1}^n \partial_{x_i}\partial_{x_j}( \Dinsde_{ij}(x,T-t) e^{\frac{\Sepsbwd}{\epsilon}})\right)\\
&= e^{-\frac{\Sepsbwd}{\epsilon}} \Bigg( -e^{\frac{\Sepsbwd}{\epsilon}} \nabla_x \cdot b(x,T-t) - e^{\frac{\Sepsbwd}{\epsilon}} b(x,T-t)\cdot \frac{\nabla_x \Sepsbwd}{\epsilon} + \frac{\epsilon}{2} \sum_{i,j=1}^n \Big(e^{\frac{\Sepsbwd}{\epsilon}} (\partial_{x_i}\partial_{x_j}\Dinsde_{ij}(x,T-t)) \\
&\quad\quad\quad\quad + \frac{2}{\epsilon}e^{\frac{\Sepsbwd}{\epsilon}}(\partial_{x_i} \Sepsbwd)(\partial_{x_j} \Dinsde_{ij}(x,T-t)) + \Dinsde_{ij}(x,T-t) e^{\frac{\Sepsbwd}{\epsilon}} \Big(\frac{(\partial_{x_i}\Sepsbwd) (\partial_{x_j}\Sepsbwd)}{\epsilon^2} +\frac{\partial_{x_i}\partial_{x_j}\Sepsbwd}{\epsilon}\Big) \Big)\Bigg)\\
&= -\nabla_x \cdot b(x,T-t) - b(x,T-t)\cdot \frac{\nabla_x \Sepsbwd}{\epsilon} + \frac{\epsilon}{2} \sum_{i,j=1}^n \Big( \partial_{x_i}\partial_{x_j}\Dinsde_{ij}(x,T-t) \\
&\quad\quad\quad\quad + \frac{2}{\epsilon}(\partial_{x_i} \Sepsbwd)(\partial_{x_j} \Dinsde_{ij}(x,T-t)) + \Dinsde_{ij}(x,T-t) \Big(\frac{(\partial_{x_i}\Sepsbwd) (\partial_{x_j}\Sepsbwd)}{\epsilon^2} + \frac{\partial_{x_i}\partial_{x_j}\Sepsbwd}{\epsilon}\Big) \Big)\\
&= -\nabla_x \cdot b(x,T-t) - b(x,T-t)\cdot \frac{\nabla_x \Sepsbwd}{\epsilon} + \frac{\epsilon}{2} \sum_{i,j=1}^n \partial_{x_i}\partial_{x_j}\Dinsde_{ij}(x,T-t) \\
&\quad\quad\quad\quad + \sum_{i,j=1}^n (\partial_{x_i} \Sepsbwd)(\partial_{x_j} \Dinsde_{ij}(x,T-t)) + \frac{1}{2\epsilon} (\nabla_x \Sepsbwd)^T \Dinsde(x,T-t) \nabla_x \Sepsbwd + \frac{1}{2}\Tr( \Dinsde(x,T-t) \nabla_x^2 \Sepsbwd).
\end{split}
\end{equation}
Therefore, the HJ equation in~\eqref{eqt:coupled_PDE_fwd} becomes
\begin{equation}
\begin{split}
0 &= \partial_t \Sepsbwd + \epsilon e^{-\frac{\Sepsbwd}{\epsilon}}\epsgeneratorfwd e^{\frac{\Sepsbwd}{\epsilon}}\\
&= \partial_t \Sepsbwd  - b(x,T-t)\cdot \nabla_x \Sepsbwd  + \epsilon \sum_{i,j=1}^n (\partial_{x_i} \Sepsbwd) (\partial_{x_j} \Dinsde_{ij}(x,T-t)) + \frac{1}{2} (\nabla_x \Sepsbwd)^T \Dinsde(x,T-t) \nabla_x \Sepsbwd  \\
&\quad\quad\quad\quad  + \frac{\epsilon}{2}\Tr( \Dinsde(x,T-t) \nabla_x^2 \Sepsbwd) + \frac{\epsilon^2}{2} \sum_{i,j=1}^n \partial_{x_i}\partial_{x_j}\Dinsde_{ij}(x,T-t) -\epsilon \nabla_x \cdot b(x,T-t).
\end{split}
\end{equation}
Similarly, according to the formula of $\epsgeneratorfwd^*$, we obtain
\begin{equation}
\begin{split}
e^{\frac{\Sepsbwd}{\epsilon}}\epsgeneratorfwd^* (\rho e^{-\frac{\Sepsbwd}{\epsilon}})
&= e^{\frac{\Sepsbwd}{\epsilon}} \left(  b(x,T-t)\cdot \nabla_x ( \rho e^{-\frac{\Sepsbwd}{\epsilon}}) + \frac{\epsilon}{2}\sum_{i,j=1}^n \Dinsde_{ij}(x,T-t)\partial_{x_i}\partial_{x_j} (\rho e^{-\frac{\Sepsbwd}{\epsilon}})\right)\\
&= e^{\frac{\Sepsbwd}{\epsilon}} \Bigg( -e^{-\frac{\Sepsbwd}{\epsilon}} \rho b(x,T-t) \cdot \frac{\nabla_x \Sepsbwd}{\epsilon} + e^{-\frac{\Sepsbwd}{\epsilon}} b(x,T-t)\cdot \nabla_x \rho + \frac{\epsilon}{2} \sum_{i,j=1}^n \Big(e^{-\frac{\Sepsbwd}{\epsilon}} \Dinsde_{ij}(x,T-t) \partial_{x_i}\partial_{x_j}\rho \\
&\quad\quad\quad\quad - \frac{2}{\epsilon}e^{-\frac{\Sepsbwd}{\epsilon}} \Dinsde_{ij}(x,T-t)(\partial_{x_i} \Sepsbwd)(\partial_{x_j} \rho) + \rho \Dinsde_{ij}(x,T-t) e^{-\frac{\Sepsbwd}{\epsilon}} \Big(\frac{(\partial_{x_i}\Sepsbwd) (\partial_{x_j}\Sepsbwd)}{\epsilon^2} - \frac{\partial_{x_i}\partial_{x_j}\Sepsbwd}{\epsilon}\Big) \Big)\Bigg)\\
&= - \rho b(x,T-t)\cdot \frac{\nabla_x \Sepsbwd}{\epsilon} + b(x,T-t) \cdot \nabla_x \rho + \frac{\epsilon}{2} \sum_{i,j=1}^n \Big( \Dinsde_{ij}(x,T-t) \partial_{x_i}\partial_{x_j}\rho \\
&\quad\quad\quad\quad - \frac{2}{\epsilon}\Dinsde_{ij}(x,T-t)(\partial_{x_i} \Sepsbwd)(\partial_{x_j} \rho) + \rho \Dinsde_{ij}(x,T-t) \Big(\frac{(\partial_{x_i}\Sepsbwd) (\partial_{x_j}\Sepsbwd)}{\epsilon^2} - \frac{\partial_{x_i}\partial_{x_j}\Sepsbwd}{\epsilon}\Big) \Big).
\end{split}
\end{equation}
Then, we have
\begin{equation}
\begin{adjustbox}{width=0.99\textwidth}$
\begin{split}
\rho e^{-\frac{\Sepsbwd}{\epsilon}}\epsgeneratorfwd e^{\frac{\Sepsbwd}{\epsilon}} - e^{\frac{\Sepsbwd}{\epsilon}}\epsgeneratorfwd^* (\rho e^{-\frac{\Sepsbwd}{\epsilon}})
&= -\rho \nabla_x \cdot b(x,T-t) - \rho b(x,T-t)\cdot \frac{\nabla_x \Sepsbwd}{\epsilon} + \frac{\epsilon \rho}{2} \sum_{i,j=1}^n \partial_{x_i}\partial_{x_j}\Dinsde_{ij}(x,T-t) \\
&\quad\quad\quad\quad + \rho \sum_{i,j=1}^n (\partial_{x_i} \Sepsbwd)(\partial_{x_j} \Dinsde_{ij}(x,T-t)) + \frac{\rho}{2\epsilon} (\nabla_x \Sepsbwd)^T \Dinsde(x,T-t) \nabla_x \Sepsbwd + \frac{\rho}{2}\Tr( \Dinsde(x,T-t) \nabla_x^2 \Sepsbwd) \\
&\quad\quad\quad\quad + \rho b(x,T-t)\cdot \frac{\nabla_x \Sepsbwd}{\epsilon} - b(x,T-t) \cdot \nabla_x \rho - \frac{\epsilon}{2} \sum_{i,j=1}^n \Big( \Dinsde_{ij}(x,T-t) \partial_{x_i}\partial_{x_j}\rho \\
&\quad\quad\quad\quad - \frac{2}{\epsilon}\Dinsde_{ij}(x,T-t)(\partial_{x_i} \Sepsbwd)(\partial_{x_j} \rho) + \rho \Dinsde_{ij}(x,T-t) \Big(\frac{(\partial_{x_i}\Sepsbwd) (\partial_{x_j}\Sepsbwd)}{\epsilon^2} - \frac{\partial_{x_i}\partial_{x_j}\Sepsbwd}{\epsilon}\Big) \Big)\\
&= -\rho \nabla_x \cdot b(x,T-t) - b(x,T-t) \cdot \nabla_x \rho + \frac{\epsilon \rho}{2} \sum_{i,j=1}^n \partial_{x_i}\partial_{x_j}\Dinsde_{ij}(x,T-t) - \frac{\epsilon}{2} \sum_{i,j=1}^n \Dinsde_{ij}(x,T-t) \partial_{x_i}\partial_{x_j}\rho \\
&\quad\quad\quad\quad + \rho \sum_{i,j=1}^n (\partial_{x_i} \Sepsbwd)(\partial_{x_j} \Dinsde_{ij}(x,T-t)) + \rho \Tr( \Dinsde(x,T-t) \nabla_x^2 \Sepsbwd) + \sum_{i,j=1}^n\Dinsde_{ij}(x,T-t)(\partial_{x_i} \Sepsbwd)(\partial_{x_j} \rho)\\
&= -\nabla_x \cdot (b(x,T-t) \rho) 
+ \frac{\epsilon \rho}{2} \sum_{i,j=1}^n \partial_{x_i}\partial_{x_j}\Dinsde_{ij}(x,T-t) - \frac{\epsilon}{2} \sum_{i,j=1}^n \Dinsde_{ij}(x,T-t) \partial_{x_i}\partial_{x_j}\rho \\
&\quad\quad\quad\quad + \nabla_x\cdot (\rho \Dinsde(x,T-t) \nabla_x \Sepsbwd)\\
&= -\nabla_x \cdot ((b(x,T-t) -  \Dinsde(x,T-t) \nabla_x \Sepsbwd)\rho) 
- \frac{\epsilon}{2}\sum_{i,j=1}^n \partial_{x_i}\partial_{x_j}(\Dinsde_{ij}(x,T-t)\rho) + \epsilon \rho \sum_{i,j=1}^n \partial_{x_i}\partial_{x_j}\Dinsde_{ij}(x,T-t) \\
&\quad\quad\quad\quad + \epsilon \sum_{i,j=1}^n (\partial_{x_i}\Dinsde_{ij}(x,T-t)) (\partial_{x_j}\rho) \\
&= -\nabla_x \cdot (b(x,T-t)\rho) + \sum_{i,j=1}^n \partial_{x_i} \left(\rho ( \Dinsde_{ij}(x,T-t) \partial_{x_j}\Sepsbwd + \epsilon \partial_{x_j}\Dinsde_{ij}(x,T-t))\right) 
- \frac{\epsilon}{2}\sum_{i,j=1}^n \partial_{x_i}\partial_{x_j}(\Dinsde_{ij}(x,T-t)\rho).
\end{split}
$\end{adjustbox}
\end{equation}
Therefore, the Fokker-Planck equation in~\eqref{eqt:coupled_PDE_fwd} becomes
\begin{equation}
\begin{adjustbox}{width=0.99\textwidth}$
\begin{split}
0 &= \partial_t \rho + \rho e^{-\frac{\Sepsbwd}{\epsilon}}\epsgeneratorfwd e^{\frac{\Sepsbwd}{\epsilon}} - e^{\frac{\Sepsbwd}{\epsilon}}\epsgeneratorfwd^* (\rho e^{-\frac{\Sepsbwd}{\epsilon}}) \\
&= \partial_t \rho -\nabla_x \cdot (b(x,T-t)\rho) +\sum_{i,j=1}^n \partial_{x_i} \left(\rho (\Dinsde_{ij}(x,T-t) \partial_{x_j}\Sepsbwd + \epsilon \partial_{x_j}\Dinsde_{ij}(x,T-t))\right) 
- \frac{\epsilon}{2}\sum_{i,j=1}^n \partial_{x_i}\partial_{x_j}(\Dinsde_{ij}(x,T-t)\rho).
\end{split}
$\end{adjustbox}
\end{equation}

After specifying the terminal condition $-J$ for $\Sepsbwd$ and the initial condition $\delta_{z_0}$ for $\rho$, the two PDEs relate to the following stochastic optimal control problem:
\begin{equation}
\begin{split}
&\min\Bigg\{ \E\Bigg[\int_0^T \frac{1}{2}v_s^T D(Z_s,T-s)^{-1}v_s - \frac{\epsilon^2}{2} \sum_{i,j=1}^n \partial_{x_i}\partial_{x_j}\Dinsde_{ij}(Z_s,T-s) + \epsilon \nabla_x \cdot b(Z_s,T-s)  ds + J(Z_T)\Bigg]\colon \\
&\quad\quad \quad\quad dZ_s = \Big(v_s -b(Z_s,T-s) + \epsilon \sum_{j=1}^n \partial_{x_j} D_j(Z_s,T-s)\Big)ds + \sqrt{\epsilon}\sigma(Z_s,T-s) dW_s, Z_0 = z_0\Bigg\},
\end{split}
\end{equation}
whose value is equal to $-\Sepsbwd(z_0, 0)$.
With different initial and terminal conditions on $\Sepsbwd$ and $\rho$, these two PDEs relate to certain MFGs or SOT problems, similar to what was discussed in Example~\ref{eg:log_transform_BM} and Section~\ref{appendix:log_sde}.

\subsection{Numerical solvers for the viscous HJ PDE~\eqref{eqt:HJsampler_SDE_HJ_general}}\label{appendx:numerical_HJsampler}

In the proposed HJ-sampler algorithm, a numerical solver is required to address the viscous HJ PDE~\eqref{eqt:HJsampler_SDE_HJ_general} in this general case. Similar to the approaches discussed in Sections~\ref{sec:HJsampler_riccati} and~\ref{sec:HJsampler_diffusion}, we will describe the Riccati method and the SGM method in the following sections.

\subsubsection{Riccati method}
The Riccati method can be applied in cases where the function \(b(x,t)\) depends linearly on \(x\), \(\sigma(x,t)\) depends only on \(t\), and the prior distribution is Gaussian. This version of the HJ-sampler is called the Riccati-HJ-sampler. Specifically, we assume that \(b\) takes the form \(b(x,t) = A_t x + \beta_t\), where \(A_t \in \mathbb{R}^{n \times n}\) and \(\beta_t \in \mathbb{R}^n\) for any \(t \in [0,T]\). We also assume that the matrix \(\Dinsde(x,t) = \sigma(x,t)\sigma(x,t)^T\) depends only on \(t\). Additionally, the prior distribution is assumed to be \(\Pprior(\Bayesparam) \propto \exp\left(-\frac{1}{2}(\Bayesparam - \Bayesparam^0)^T\Sigma^{-1} (\Bayesparam - \Bayesparam^0)\right)\). Under these assumptions, the viscous HJ PDE~\eqref{eqt:HJsampler_SDE_HJ_general} in the first step of the Riccati-HJ-sampler can be solved using Riccati ODEs, following a similar approach to that in Section~\ref{sec:HJsampler_riccati}. 
For simplicity, in this section, we will interchangeably use both \(\Dinsde_t\) and \(\Dinsde(x,t)\), as well as \(\Dinsde_{t,ij}\) and \(\Dinsde_{ij}(x,t)\), as needed.

Let $\tilde S(x,t) = -\Sepsbwd(x,T-t)$, where $\Sepsbwd$ is the solution to~\eqref{eqt:HJsampler_SDE_HJ_general}. 
Then, $\tilde S$ satisfies
\begin{equation}
\begin{dcases}
\partial_t \tilde S + b(x,t)\cdot \nabla_x \tilde S  - \epsilon \sum_{i,j=1}^n (\partial_{x_i} \tilde S) (\partial_{x_j} \Dinsde_{ij}(x,t)) + \frac{1}{2} (\nabla_x \tilde S)^T \Dinsde(x,t) \nabla_x \tilde S  \\
\quad\quad\quad\quad - \frac{\epsilon}{2}\Tr( \Dinsde(x,t) \nabla_x^2 \tilde S) + \frac{\epsilon^2}{2} \sum_{i,j=1}^n \partial_{x_i}\partial_{x_j}\Dinsde_{ij}(x,t) -\epsilon \nabla_x \cdot b(x,t) = 0,\\
\tilde S(x,0) = -\epsilon\log P_{prior}(x).
\end{dcases}
\end{equation}
Under the assumptions on $b$, $\Dinsde$, and $\Pprior$, this PDE simplifies to
\begin{equation}
\begin{dcases}
\partial_t \tilde S + (A_{t} x + \beta_{t})^T \nabla_x \tilde S  + \frac{1}{2} (\nabla_x \tilde S)^T \Dinsde_{t} \nabla_x \tilde S  - \frac{\epsilon}{2}\Tr( \Dinsde_{t} \nabla_x^2 \tilde S) -\epsilon \Tr(A_{t}) = 0,\\
\tilde S(x,0) = \frac{\epsilon}{2} (x - \Bayesparam^0)^T\Sigma^{-1} (x - \Bayesparam^0) + \frac{n\epsilon}{2}\log(2\pi) + \frac{\epsilon}{2}\log\det(\Sigma).
\end{dcases}
\end{equation}
In this viscous HJ PDE, both the Hamiltonian and the initial condition are quadratic, allowing the equation to be solved using Riccati ODEs. Specifically, the solution is given by \(\tilde S(x,t) = \frac{1}{2} (x-\Sx(t))^T\Sxx(t)^{-1}(x-\Sx(t)) + \Sc(t)\), where $\Sxx\colon [0,T]\to\sympos$, $\Sx\colon [0,T]\to\Rn$, and $\Sc\colon [0,T]\to\R$ satisfy the following Riccati ODE system:
\begin{equation}
\begin{dcases}
\dot \Sxx(t)= \Dinsde_{t} + \Sxx(t)A_{t}^T + A_{t} \Sxx(t), \\
\dot \Sx(t) = A_{t} \Sx(t) + \beta_{t},\\
\dot \Sc(t) = \frac{\epsilon}{2} \Tr(2A_{t} + \Dinsde_{t}\Sxx(t)^{-1}),
\end{dcases}
\quad\quad \quad\quad 
\begin{dcases}
\Sxx(0) = \frac{1}{\epsilon} \Sigma, \\
\Sx(0) = \Bayesparam^0,\\
\Sc(0) = \frac{n\epsilon}{2}\log(2\pi) + \frac{\epsilon}{2}\log\det(\Sigma).
\end{dcases}
\end{equation}
Since sampling only requires $\nabla_x \Sepsbwd(x,t)$, it suffices to solve for $\Sxx$ and $\Sx$, and then obtain
\begin{equation}
\nabla_x \Sepsbwd(x,t) = - \nabla_x \tilde S(x,T-t)= -\Sxx(T-t)^{-1}( x - \Sx(T-t)).
\end{equation}

Beyond the Gaussian prior, we can also handle Gaussian mixture prior distributions. Similar to the approach in Section~\ref{sec:HJsampler_riccati}, we first solve the Riccati ODE for each Gaussian component and then compute \(\nabla_x \Sepsbwd\) using~\eqref{eqt:control_mixedGaussian}.

\subsubsection{SGM method}\label{appendix:C22_SGM}
For more general cases, we can approximate the score function using a neural network trained on samples of \(Y_t\) via the SGM method, similar to the approach in Section~\ref{sec:HJsampler_diffusion}. This approach is called the SGM-HJ-sampler. In this case, the training data are sampled from the discretized SDE as follows:
\begin{equation}\label{eqt:appendix_C2_discretizedode}
     Y_{k+1, j} = Y_{k, j} + b(Y_{k, j}, t_k)\Delta t + \sqrt{\epsilon \Delta t}\sigma(Y_{k,j},t_k) \xi_{k,j}.
    \end{equation}
The neural network can also be trained using the loss functions~\eqref{eqt:sgm_loss} and~\eqref{eqt:slice_sgm_loss}. After training, the term \(\nabla_x \Sepsbwd(Z_k, t_k)\) in the sampling step can be approximated by the pretrained model \(\epsilon s_\NNparam(Z_k, T-t_k)\).

There are alternative choices for the loss function in the training of diffusion models. A popular choice is:
\begin{equation}
\sum_{k=1}^{n_t} \sum_{j=1}^N \lambda_k \|s_\NNparam(Y_{k,j}, t_k) - \nabla_{y_k} \log P(Y_{t_k} = Y_{k, j} | Y_0 = Y_{0, j})\|^2,
\end{equation}
where \(\lambda_k\) are positive weighting terms, and various methods for selecting them in diffusion models are discussed in~\cite{song2020score,karras2022elucidating}.

In our case, when the model for $Y_t$ is complicated, we do not have an analytical formula for $\log P(Y_{t_k} | Y_0)$. 
However, using a similar derivation, we can employ the following loss instead:
\begin{equation}
\begin{split}
&\sum_{k=1}^{n_t} \sum_{j=1}^N \lambda_k\|s_\NNparam(Y_{k,j}, t_k) - \nabla_{y_k} \log P(Y_{t_k} = Y_{k, j} | Y_{t_{k-1}} = Y_{k-1, j})\|^2\\
\approx &\sum_{k=1}^{n_t}\sum_{j=1}^N \lambda_k\left\|s_\NNparam(Y_{k, j}, t_k) 
+ \frac{1}{\epsilon\Delta t}\Dinsde(Y_{k-1,j},t_{k-1})^{-1}(Y_{k,j} -  Y_{k-1,j} - b(Y_{k-1,j}, t_{k-1})\Delta t)\right\|^2.
\end{split}
\end{equation}
In this scenario, the conditional distribution \(P(Y_{t_k} | Y_{t_{k-1}})\) does not have an analytical formula, so we approximate it using the conditional distribution from the discretized process in~\eqref{eqt:appendix_C2_discretizedode}.
Note that this loss function requires the matrix \(\Dinsde(y,t)\) to be invertible for all \(y=Y_{k,j}\) and \(t=t_k\). The equivalence of these loss functions, including the original score matching loss \(\sum_{k=1}^{n_t} \sum_{j=1}^N \lambda_k \|s_\NNparam(Y_{k,j}, t_k) - \nabla_{y_k} \log P(Y_{t_k} = Y_{k,j})\|^2\), is discussed in~\cite{vincent2011connection}.


\subsection{Error estimation for HJ-sampler}\label{appendix:C3_err_est}

In this section, we present a simple error estimation for the HJ-sampler. The total error in the HJ-sampler arises from two main sources: the numerical error in solving the viscous HJ PDE during the first step, and the sampling error in the second step. Here, we focus on analyzing the impact of the first error.

Let \(\rho(x, t|\ydata)\) or \(\rho_{\ydata}(x, t)\) denote the density of the stochastic process described in~\eqref{eqt:appendixC_controlledsde_cont}, and let \(\tilde\rho(x, t|\ydata)\) or \(\tilde\rho_{\ydata}(x, t)\) represent the density of the following stochastic process:
\begin{equation}
\begin{split}
dZ_\tau &= \left(\Dinsde(Z_\tau,T-\tau)\appendixcontrol(Z_{\tau}, \tau) - b(Z_\tau, T-\tau) + \epsilon \sum_{j=1}^n \partial_{x_j} \Dinsde_j(Z_\tau,T-\tau)\right) d\tau + \sqrt{\epsilon} \sigma(Z_\tau,T-\tau) dW_\tau, \\
Z_0 &= \ydata,
\end{split}
\end{equation}
where \(\appendixcontrol\) is an approximation of the control \(\nabla_x \Sepsbwd\). We compare the error between the true distribution \(\rho_{\ydata}(\cdot, t)\) and its numerical approximation \(\tilde\rho_{\ydata}(\cdot, t)\), measured by the Wasserstein-2 distance, denoted as \(W_2(\rho_{\ydata}(\cdot, t), \tilde \rho_{\ydata}(\cdot, t))\).

In the case of the Riccati-HJ-sampler, \(\appendixcontrol\) can be considered as an interpolation of the spatial gradient of the Riccati solution at temporal grid points \(t_k\). For the SGM-HJ-sampler, the function \(\appendixcontrol(x,t)\) is given by the pretrained neural network \(\epsilon s_\NNparam(x,T-t)\).

Following the proof in~\cite{kwon2022score}, we adapt the methodology to our setup. We provide an informal error estimation, focusing on the essential ideas while simplifying the analysis by omitting some technical details. We assume sufficient regularity of the functions so that the Fokker-Planck equations have smooth solutions \(\rho_{\ydata}(x,t)\) and \(\tilde\rho_{\ydata}(x,t)\) for any \(\ydata\), \(x \in \mathbb{R}^n\), and \(t \in (0,T]\). Additionally, we assume that the functions \(b\)  and \(\appendixcontrol\) are Lipschitz continuous with Lipschitz constants \(L_b\) and \(L_{\appendixcontrol}\), respectively.
We further assume that \(\Dinsde(x,t)\) depends only on \(t\) and is bounded by \(C_{\Dinsde}(t) I\), meaning that \(C_{\Dinsde}(t) I - \Dinsde(x,t)\) is positive semi-definite.
We assume that \(C_{\Dinsde}(t)\) and \(\exp\left(-2L_b t -2 L_{\appendixcontrol}\int_0^t C_{\Dinsde}(T-\tau)  d\tau\right) C_{\Dinsde}(T-t)^2\) are both integrable over \(t\in[0,T]\).
Moreover, we assume that the optimal transport map from \(\rho_{\ydata}(\cdot, t)\) to \(\tilde\rho_{\ydata}(\cdot, t)\) is given by \(\nabla \phi\), the gradient of a convex, second-order differentiable function \(\phi\) with invertible Hessian matrices. Note that the function $\phi$ may vary with $t$, but we omit this time dependence for simplicity of notation.

Under these assumptions, the function \(\rho_{\ydata}\) satisfies the following equation:
\begin{equation}
\begin{split}
0 &= \partial_t \rho_{\ydata} -\nabla_x \cdot (b(x,T-t)\rho_{\ydata}) +\sum_{i,j=1}^n \partial_{x_i} \left(\rho_{\ydata} \Dinsde_{ij}(T-t) \partial_{x_j}\Sepsbwd\right) 
- \frac{\epsilon}{2}\sum_{i,j=1}^n \partial_{x_i}\partial_{x_j}(\Dinsde_{ij}(T-t)\rho_{\ydata}) \\
&= \partial_t \rho_{\ydata} +\sum_{i=1}^n\partial_{x_i}\left(\rho_{\ydata}\left(-b_i(x,T-t) +\sum_{j=1}^n \left(\Dinsde_{ij}(T-t) \partial_{x_j}\Sepsbwd
- \frac{\epsilon \Dinsde_{ij}(T-t)\partial_{x_j}\rho_{\ydata}}{2\rho_{\ydata}}\right)\right)\right).
\end{split}
\end{equation}
Thus, \(\rho_{\ydata}(\cdot, t)\) is also the marginal density of the ODE \(\dot{x}_t = v(x_t, t | \rho_{\ydata})\), where \(v\) is given by:
\begin{equation}
\begin{split}
v(x, t | \rho_{\ydata}) &= -b(x,T-t) +\Dinsde(T-t) \nabla_x\Sepsbwd(x,t) 
- \frac{\epsilon \Dinsde(T-t)\nabla_x\rho_{\ydata}}{2\rho_{\ydata}}.
\end{split}
\end{equation}
Similarly, the function \(\tilde\rho_{\ydata}(\cdot, t)\) is the marginal density of the ODE \(\dot{x}_t = \tilde{v}(x_t, t | \tilde{\rho}_{\ydata})\), where \(\tilde{v}\) is given by:
\begin{equation}
\tilde v(x, t | \tilde \rho_{\ydata}) = -b(x,T-t) +\Dinsde(T-t) \appendixcontrol(x, t) 
- \frac{\epsilon \Dinsde(T-t)\nabla_x\tilde\rho_{\ydata}}{2\tilde\rho_{\ydata}}.
\end{equation}

According to~\cite[Cor.5.25]{santambrogio2015optimal}, the Wasserstein distance between \(\rho_{\ydata}(\cdot, t)\) and \(\tilde\rho_{\ydata}(\cdot, t)\), denoted by \(t \mapsto W_2(\rho_{\ydata}, \tilde \rho_{\ydata})(t) = W_2(\rho_{\ydata}(\cdot, t), \tilde \rho_{\ydata}(\cdot, t))\), satisfies 
\begin{equation}\label{eqt:appendixC3_flow_W2}
\frac{1}{2} \frac{d}{dt} W_2(\rho_{\ydata}, \tilde \rho_{\ydata})^2 = \E_{\pi(x,y)}[(x-y)^T (v(x, t | \rho_{\ydata}) - \tilde v(y, t | \tilde \rho_{\ydata}))],
\end{equation}
where \(\pi\) is the optimal transport plan between \(\rho_{\ydata}\) and \(\tilde \rho_{\ydata}\). For simplicity, we omit the variable \(t\) in the density functions.

Next, we will bound the right-hand side of~\eqref{eqt:appendixC3_flow_W2}. Through straightforward computation, we obtain
\begin{equation}\label{eqt:appendixC3_est_term}
\begin{split}
&\E_{\pi(x,y)}[(x-y)^T (v(x, t | \rho_{\ydata}) - \tilde v(y, t | \tilde \rho_{\ydata}))] = \E_{\pi(x,y)}[(x-y)^T (-b(x, T-t) + b(y, T-t))] \\
&\quad\quad\quad\quad\quad\quad\quad\quad + \E_{\pi(x,y)}\left[(x - y)^T\Dinsde(T-t)( \nabla_x\Sepsbwd(x,t) - \appendixcontrol(y, t))\right]\\
&\quad\quad\quad\quad\quad\quad\quad\quad -\frac{\epsilon}{2} \E_{\pi(x,y)}\left[(x - y)^T \Dinsde(T-t)\left(\frac{\nabla_x \rho_{\ydata}(x,t)}{\rho_{\ydata}(x,t)} - \frac{\nabla_y\tilde\rho_{\ydata}(y,t)}{\tilde\rho_{\ydata}(y,t)}\right)\right].
\end{split}
\end{equation}
Since the function \(b\) is Lipschitz continuous with respect to \(x\), with uniform Lipschitz constant \(L_b\), the first term on the right-hand side of~\eqref{eqt:appendixC3_est_term} is bounded above by \(L_b W_2(\rho_{\ydata}, \tilde \rho_{\ydata})^2\). Furthermore, since the matrix \(\Dinsde(t)\) is bounded above by \(C_D(t) I\) and the function \(\appendixcontrol\) is Lipschitz with respect to \(x\) with constant \(L_{\appendixcontrol}\), the second term on the right-hand side of~\eqref{eqt:appendixC3_est_term} is estimated by 
\begin{equation}
\begin{split}
&\E_{\pi(x,y)}\left[(x - y)^T \Dinsde(T-t) (\nabla_x\Sepsbwd(x,t) - \appendixcontrol(y, t))\right]\\
\leq \,& C_D(T-t) \E_{\pi(x,y)}\left[(x - y)^T (\nabla_x\Sepsbwd(x,t) - \appendixcontrol(y, t))\right]\\
=\,&  C_D(T-t) \E_{\pi(x,y)}\left[(x - y)^T (\nabla_x\Sepsbwd(x,t) - \appendixcontrol(x, t))\right] + C_D(T-t) \E_{\pi(x,y)}\left[(x - y)^T ( \appendixcontrol(x, t) -  \appendixcontrol(y, t))\right] \\
\leq \, & C_D(T-t)W_2(\rho_{\ydata}, \tilde \rho_{\ydata})\left(\E_{\pi(x,y)}\left[\|\nabla_x\Sepsbwd(x,t) - \appendixcontrol(x, t)\|^2\right]\right)^{1/2} + C_D(T-t)  L_{\appendixcontrol} W_2(\rho_{\ydata}, \tilde \rho_{\ydata})^2\\
= \, & C_D(T-t)W_2(\rho_{\ydata}, \tilde \rho_{\ydata})\left(\E_{\rho_{\ydata}}\left[\|\nabla_x\Sepsbwd(\cdot,t) - \appendixcontrol(\cdot, t)\|^2\right]\right)^{1/2} + C_D(T-t)  L_{\appendixcontrol} W_2(\rho_{\ydata}, \tilde \rho_{\ydata})^2.
\end{split}
\end{equation}

According to optimal transport theory, there exists a convex function \(\phi\) such that the optimal transport plan \(\pi\) is given by \((\text{Id}, \nabla \phi)_\# \rho_{\ydata}\), which denotes the push-forward of \(\rho_{\ydata}\) by \((\text{Id}, \nabla \phi)\) (where \(\text{Id}\) denotes the identity map \(x \mapsto x\)). Denote by \(\phi^*\) the Fenchel-Legendre transform of \(\phi\). Then, \(\pi\) can also be represented by \((\nabla \phi^*, \text{Id})_\# \tilde \rho_{\ydata}\).
The expectation in the last term on the right-hand side of~\eqref{eqt:appendixC3_est_term} simplifies as follows:
\begin{equation}\label{eqt:appendixC3_est_term3}
\begin{split}
&\E_{\pi(x,y)}\left[(x - y)^T \Dinsde(T-t)\left(\frac{\nabla_x\rho_{\ydata}(x,t))}{\rho_{\ydata}(x,t)} - \frac{\nabla_y\tilde\rho_{\ydata}(y,t))}{\tilde\rho_{\ydata}(y,t)}\right)\right]\\
= \, & \E_{\rho_{\ydata}}\left[ \frac{(x - \nabla\phi(x))^T\Dinsde(T-t)\nabla_x\rho_{\ydata}(x,t)}{\rho_{\ydata}(x,t)}\right] - \E_{\tilde\rho_{\ydata}}\left[\frac{(\nabla \phi^*(y) - y)^T\Dinsde(T-t)\nabla_y\tilde\rho_{\ydata}(y,t)}{\tilde\rho_{\ydata}(y,t)}\right]\\
= \, & \int (x - \nabla\phi(x))^T\Dinsde(T-t)\nabla_x\rho_{\ydata}(x,t) dx - \int (\nabla\phi^*(y) - y)^T\Dinsde(T-t)\nabla_{y}\tilde\rho_{\ydata}(y,t)dy\\
= \, & -\int \Tr\left((I - \nabla^2\phi(x))\Dinsde(T-t)\right)\rho_{\ydata}(x,t) dx + \int \Tr\left((\nabla^2\phi^*(y) - I)\Dinsde(T-t)\right)\tilde\rho_{\ydata}(y,t)dy\\
= \, &  - \E_{\rho_{\ydata}}\left[\Tr(\Dinsde(T-t)(I - \nabla^2 \phi))\right] + \E_{\tilde\rho_{\ydata}}\left[\Tr(\Dinsde(T-t)(\nabla^2 \phi^* - I))\right]\\
= \, &  - \E_{\rho_{\ydata}}\left[\Tr(\Dinsde(T-t)(I - \nabla^2 \phi))\right] + \E_{\rho_{\ydata}}\left[\Tr(\Dinsde(T-t)((\nabla^2 \phi)^{-1} - I))\right]\\
= \, & \E_{\rho_{\ydata}}\left[\Tr\left(\Dinsde(T-t)(\nabla^2 \phi(x) + (\nabla^2 \phi(x))^{-1} - 2I)\right)\right]\\
= \, & \E_{\rho_{\ydata}}\left[\Tr\left(\Dinsde(T-t)(\nabla^2 \phi(x))^{-1}(\nabla^2 \phi(x) - I)^2\right)\right]\geq 0,
\end{split}
\end{equation}
where the last term is non-negative because the matrix \(\Dinsde(T-t)(\nabla^2 \phi(x))^{-1}(\nabla^2 \phi(x) - I)^2\) is positive semi-definite for any \(t\in [0,T]\).

Combining all the estimates, we have
\begin{equation}
\begin{adjustbox}{width=0.99\textwidth}$
\begin{split}
\frac{1}{2} \frac{d}{dt} W_2(\rho_{\ydata}, \tilde \rho_{\ydata})^2\leq 
C_1(t) W_2(\rho_{\ydata}, \tilde \rho_{\ydata})^2 + C_D(T-t)W_2(\rho_{\ydata}, \tilde \rho_{\ydata})\left(\E_{\rho_{\ydata}}\left[\|\nabla_x\Sepsbwd(\cdot,t) - \appendixcontrol(\cdot, t)\|^2\right]\right)^{1/2},
\end{split}
$\end{adjustbox}
\end{equation}
where \(C_1(t) = L_b + C_D(T-t) L_{\appendixcontrol}\) is a function of \(t\). 
Next, we take the expectation with respect to \(\ydata \sim P(\ydata) = \mubwd(\ydata, T)\) and obtain
\begin{equation}
\begin{adjustbox}{width=0.99\textwidth}$
\begin{split}
&\frac{1}{2} \frac{d}{dt} \E_{P(\ydata)}\left[W_2(\rho_{\ydata}, \tilde \rho_{\ydata})^2\right]\\
\leq &
C_1(t) \E_{P(\ydata)}\left[W_2(\rho_{\ydata}, \tilde \rho_{\ydata})^2\right] 
+C_D(T-t)\E_{P(\ydata)}\left[W_2(\rho_{\ydata}, \tilde \rho_{\ydata})\left(\E_{\rho_{\ydata}}\left[\|\nabla_x\Sepsbwd(\cdot,t) - \appendixcontrol(\cdot, t)\|^2\right]\right)^{1/2}\right]\\
\leq & C_1(t) \E_{P(\ydata)}\left[W_2(\rho_{\ydata}, \tilde \rho_{\ydata})^2\right] 
+C_D(T-t)\left(\E_{P(\ydata)}\left[W_2(\rho_{\ydata}, \tilde \rho_{\ydata})^2\right]\right)^{1/2}\left(\E_{P(\ydata)}\E_{\rho_{\ydata}}\left[\|\nabla_x\Sepsbwd(\cdot,t) - \appendixcontrol(\cdot, t)\|^2\right]\right)^{1/2}\\
\leq & C_1(t) \E_{P(\ydata)}\left[W_2(\rho_{\ydata}, \tilde \rho_{\ydata})^2\right] 
+C_D(T-t)\left(\E_{P(\ydata)}\left[W_2(\rho_{\ydata}, \tilde \rho_{\ydata})^2\right]\right)^{1/2}\left(\E\left[\|\nabla_x\Sepsbwd(Y_{T-t},t) - \appendixcontrol(Y_{T-t}, t)\|^2\right]\right)^{1/2},
\end{split}
$\end{adjustbox}
\end{equation}
where the second inequality follows from the Cauchy-Schwarz inequality, and the last one holds because \(P(\ydata)\rho_{\ydata}(x,t) = P(Y_{T,1} = \ydata)P(Y_{T-t} = x | Y_{T,1} = \ydata) = P(Y_{T-t} = x, Y_{T,1} = \ydata)\) following~\eqref{eqt:rho_Bayesian_formula}.
Let \(\mathcal{L}(t) = \left(\E_{P(\ydata)}\left[W_2(\rho_{\ydata}, \tilde \rho_{\ydata})^2\right]\right)^{1/2}\). We obtain
\begin{equation}
\frac{d\mathcal{L}(t)}{dt} = \frac{\frac{1}{2} \frac{d}{dt} \E_{P(\ydata)}\left[W_2(\rho_{\ydata}, \tilde \rho_{\ydata})^2\right]}{\mathcal{L}(t)} \leq C_1(t) \mathcal{L}(t) + C_D(T-t) \left(\E\left[\|\nabla_x\Sepsbwd(Y_{T-t},t) - \appendixcontrol(Y_{T-t}, t)\|^2\right]\right)^{1/2}.
\end{equation}
Moreover, we have \(\mathcal{L}(0) = \left(\E_{P(\ydata)}\left[W_2(\rho_{\ydata}(\cdot,0), \tilde \rho_{\ydata}(\cdot,0))^2\right]\right)^{1/2} = \left(\E_{P(\ydata)}\left[W_2(\delta_{\ydata}, \delta_{\ydata})^2\right]\right)^{1/2} = 0\).
Then, by a similar argument as in Appendix A1 of~\cite{kwon2022score}, we conclude that
\begin{equation}
\mathcal{L}(t)\leq \frac{1}{I(t)}\int_0^t I(\tau)C_D(T-\tau) \left(\E\left[\|\nabla_x\Sepsbwd(Y_{T-\tau},\tau) - \appendixcontrol(Y_{T-\tau},\tau)\|^2\right]\right)^{1/2}d\tau,
\end{equation}
where the function \(I\) is defined by \(I(t) = \exp(-\int_0^t C_1(\tau)d\tau)\).

Note that \(\mathcal{L}(t)^2\) is the expected squared Wasserstein-2 error of the posterior distribution with respect to the observation \(\ydata\). According to Markov's inequality, the probability of the error exceeding \(e\) can be bounded as follows:
\begin{equation}
\begin{split}
&P(W_2(\rho_{\ydata}(\cdot,t), \tilde \rho_{\ydata}(\cdot,t)) \geq e) \leq \frac{\E_{P(\ydata)}\left[W_2(\rho_{\ydata}, \tilde \rho_{\ydata})^2\right]}{e^2} = \frac{\mathcal{L}(t)^2}{e^2}\\
\leq \, & \frac{1}{e^2}\left(\frac{1}{I(t)}\int_0^t I(\tau)C_D(T-\tau) \left(\E\left[\|\nabla_x\Sepsbwd(Y_{T-\tau},\tau) - \appendixcontrol(Y_{T-\tau}, \tau)\|^2\right]\right)^{1/2}d\tau\right)^2.
\end{split}
\end{equation}

\begin{rem}
In the error estimation above, beyond the regularity assumptions, one restrictive assumption is that the function \(\Dinsde\) depends only on \(t\). This excludes SDEs where the diffusion coefficient has \(x\) dependence. This assumption is primarily utilized in the estimation in~\eqref{eqt:appendixC3_est_term3}. Other parts of the proof can be straightforwardly generalized to cases where \(\Dinsde\) depends on \(x\). However, if \(\Dinsde\) does depend on \(x\), the term estimated in~\eqref{eqt:appendixC3_est_term3} may become negative. If this term can be appropriately bounded, the proof could be extended to more general cases.
\end{rem}


\begin{rem}
In the error analysis above, we examined how the numerical error from the first step of the HJ-sampler affects the continuous sampling process. The key term in the upper bound is the expectation \(\E\left[\|\nabla_x \Sepsbwd(Y_{T-\tau},\tau) - \appendixcontrol(Y_{T-\tau},\tau)\|^2\right]\). For the Riccati-HJ-sampler, this term can be calculated based on the errors in the solutions \(\Sxx\) and \(\Sx\) to the Riccati ODE system. For the SGM-HJ-sampler, this term is related to the loss function and can be bounded by quantities derived from the loss value.

The overall error of the sampling algorithm consists of two parts: the error from the first step and the discretization error from the second step. 
The error in the first step, as estimated in our analysis, can guide the choice of the numerical solver for the viscous HJ PDE, including decisions on the temporal discretization size for traditional methods or the data size and neural network size for AI-based methods.
The second step's error, resulting from the SDE discretization scheme, can guide the selection of the sampling method and its temporal discretization size. Balancing these two sources of error, along with considering the computational efficiency of different methods, is crucial for achieving optimal performance.

The error analysis for the second step follows directly from the established analysis of SDE discretization schemes, so we omit the details here. In this paper, we use Euler–Maruyama discretization in the second step, yielding an order of \(0.5\) in the strong sense and order \(1\) in the weak sense. If higher accuracy is needed, higher-order schemes such as the Runge-Kutta method could be considered.
\end{rem}

    


\section{Neural network training details for numerical examples}\label{sec:details}

This section provides the details of the neural network \( s_\NNparam \) used in the SGM-HJ-sampler for obtaining posterior samples, as well as the specifics of the training procedure. Across all numerical examples, \( s_\NNparam \) is implemented with the \(\tanh\) activation function for the nonlinear hidden layers, and the Adam optimizer \cite{kingma2014adam} is employed with a learning rate of \( 1 \times 10^{-4} \), unless stated otherwise.


In Section~\ref{sec:numerics_eg1}, we explore three 1D cases with varying prior distributions and one 2D case. For the 1D Gaussian prior case, a fully-connected neural network (FNN) with two hidden layers of 50 neurons each is utilized. The training of \( s_\NNparam \) involves \( 1,000,000 \) sample paths of \( Y_t \) over \( t \in [0, T] \), using mini-batch training with a batch size of \( 1,000 \) for \( 3,000 \) epochs. For the 1D Gaussian mixture case, the 1D mixture of uniform distributions case, and the 2D Gaussian mixture case, an FNN with three hidden layers and 50 neurons per layer is employed. The network is trained on \( 1,000,000 \) sample paths of \( Y_t \) over \( t \in [0, T] \) for \( 5,000 \) epochs with a batch size of \( 1,000 \).


In Section~\ref{sec:numerics_eg2}, the examples include one 1D case and two 2D cases. For the 1D case, an FNN with two hidden layers of 50 neurons each is trained on \( 1,000,000 \) sample paths of \( Y_t \) over \( t \in [0, T] \) for \( 3,000 \) epochs, using a batch size of \( 1,000 \) and a learning rate of \( 1 \times 10^{-4} \). For the 2D Gaussian mixture prior case, an FNN with three hidden layers of 50 neurons each is trained on \( 1,000,000 \) sample paths of \( Y_t \) over \( t \in [0, T] \) for \( 5,000 \) epochs with a batch size of \( 1,000 \). For the 2D case with a LogNormal prior, which considers model misspecification, an FNN with three hidden layers of 50 neurons each is trained separately for each value of \( \epsilon \). Each network is trained on \( 100,000 \) sample paths of \( Y_t \) over \( t \in [0, T] \) for \( 5,000 \) epochs, using a batch size of \( 1,000 \) and a learning rate of \( 1 \times 10^{-3} \).


In Section~\ref{sec:numerics_eg3}, for each value of \( \epsilon \), an FNN with three hidden layers and 50 neurons per layer is trained on \( 100,000 \) sample paths of \( Y_t \) over \( t \in [0, T] \) for \( 5,000 \) epochs with a batch size of \( 1,000 \). In Section~\ref{sec:numerics_eg4}, an FNN with three hidden layers and 200 neurons per layer is used, trained on \( 100,000 \) sample paths of \( Y_t \) over \( t \in [0, T] \) for \( 3,000 \) epochs with a batch size of \( 1,000 \) and a learning rate of \( 1 \times 10^{-3} \).


In the first three sections, the loss function~\eqref{eqt:sgm_loss} is used with a weight of \( \lambda_k = 1 \) for training. In the final section, the sliced version~\eqref{eqt:slice_sgm_loss} is used with a weight of \( \lambda_k = 1 \) and a sample size \( N_v = 1 \).



\section{Details of analytical formulas used in numerical results}\label{appendix:analytical_numeical_examples}

\subsection{Brownian motion}

In this section, we focus on cases involving Brownian motion. Specifically, we assume the process \(Y_t\) is governed by the SDE \(dY_t = \sqrt{\epsilon}dW_t\), where $\epsilon>0$ is a hyperparameter, and $W_t$ is standard Brownian motion, with different prior distributions for \(Y_0\).

\subsubsection{One-dimensional uniform prior}
Here, we provide the computational details for a one-dimensional uniform prior and a mixture of uniform priors. The generalization to higher-dimensional box-shaped domains is straightforward and therefore omitted.

First, assume the prior is a uniform distribution on \([a,b]\). The marginal distribution for \(Y_t\) is given by:
\begin{equation}
\begin{split}
P(Y_t = y) &= \int P(Y_t = y| Y_0 = x)P(Y_0 = x)dx
= \frac{1}{b-a}\int_a^b \frac{1}{\sqrt{2\pi \epsilon t}}\exp\left(-\frac{1}{2\epsilon t} |y-x|^2\right) dx\\
&= \frac{1}{b-a}\int_{\frac{y-b}{\sqrt{\epsilon t}}}^{\frac{y-a}{\sqrt{\epsilon t}}} \frac{1}{\sqrt{(2\pi)^n}}\exp\left(-\frac{1}{2} x^2\right) dx = \frac{1}{b-a}\left(\Phi\left(\frac{y - a}{\sqrt{\epsilon t}}\right) - \Phi\left(\frac{y - b}{\sqrt{\epsilon t}}\right)\right),
\end{split}
\end{equation}
where \(\Phi\) is the cumulative distribution function of the standard one-dimensional Gaussian distribution. The solution to the viscous HJ PDE~\eqref{eqt:HJsampler_SDE_HJ} is:
\begin{equation}
\Sepsbwd(x,T-t) = \epsilon\log P(Y_t = x) = \epsilon \log \left(\Phi\left(\frac{x - a}{\sqrt{\epsilon t}}\right) - \Phi\left(\frac{x - b}{\sqrt{\epsilon t}}\right)\right) - \epsilon \log (b-a).
\end{equation}
The inference process, as described in~\eqref{eq:Euler–Maruyama}, becomes:
\begin{equation}
\begin{split}
Z_{k+1} &= Z_{k} + \partial_{x} \Sepsbwd(Z_k, t_k) \Delta t + \sqrt{\epsilon \Delta t} \xi_k\\
&= Z_{k} + \epsilon \frac{\frac{1}{\sqrt{\epsilon (T-t_k)}} \left(\Phi'\left(\frac{Z_{k} - a}{\sqrt{\epsilon (T-t_k)}}\right) - \Phi'\left(\frac{Z_{k} - b}{\sqrt{\epsilon (T-t_k)}}\right)\right)}{\Phi\left(\frac{Z_{k} - a}{\sqrt{\epsilon (T-t_k)}}\right) - \Phi\left(\frac{Z_{k} - b}{\sqrt{\epsilon (T-t_k)}}\right)} \Delta t + \sqrt{\epsilon \Delta t} \xi_k\\
&= Z_{k} + \sqrt{\frac{\epsilon}{2\pi (T-t_k)}} \frac{\exp\left(-\frac{1}{2\epsilon (T-t_k)}|Z_{k} - a|^2\right) - \exp\left(-\frac{1}{2\epsilon (T-t_k)}|Z_{k} - b|^2\right)}{\Phi\left(\frac{Z_{k} - a}{\sqrt{\epsilon (T-t_k)}}\right) - \Phi\left(\frac{Z_{k} - b}{\sqrt{\epsilon (T-t_k)}}\right)} \Delta t + \sqrt{\epsilon \Delta t} \xi_k.
\end{split}
\end{equation}
The posterior distribution of \(Y_t\) given \(Y_T = \ydata\) for \(t \in (0,T)\) is:
\begin{equation}
\begin{adjustbox}{width=0.99\textwidth}$
\begin{split}
P(Y_t = \Bayesparam | Y_T =\ydata) &= \frac{P(Y_t = \Bayesparam) P(Y_T=\ydata | Y_t = \Bayesparam)}{P(Y_T = \ydata)}
= \frac{\left(\Phi\left(\frac{\Bayesparam - a}{\sqrt{\epsilon t}}\right) - \Phi\left(\frac{\Bayesparam- b}{\sqrt{\epsilon t}}\right)\right)\frac{1}{\sqrt{2\pi\epsilon (T-t)}} \exp\left(-\frac{1}{2\epsilon (T-t)} |\Bayesparam - \ydata|^2\right)}{\Phi\left(\frac{\ydata - a}{\sqrt{\epsilon T}}\right) - \Phi\left(\frac{\ydata - b}{\sqrt{\epsilon T}}\right)}.
\end{split}
$\end{adjustbox}
\end{equation}
When \(t=0\), the posterior distribution is:
\begin{equation}
\begin{split}
P(Y_0 = \Bayesparam | Y_T =\ydata) &= \frac{P(Y_0 = \Bayesparam) P(Y_T=\ydata | Y_0 = \Bayesparam)}{P(Y_T = \ydata)}
= \frac{\frac{1}{\sqrt{2\pi\epsilon T}} \exp\left(-\frac{1}{2\epsilon T} |\Bayesparam - \ydata|^2\right)}{\Phi\left(\frac{\ydata - a}{\sqrt{\epsilon T}}\right) - \Phi\left(\frac{\ydata - b}{\sqrt{\epsilon T}}\right)}\chi_{[a,b]} (\Bayesparam),
\end{split}
\end{equation}
where \(\chi_{[a,b]}\) is the indicator function that takes the value \(1\) on \([a,b]\) and \(0\) otherwise.

Next, consider the prior to be a mixture of uniform distributions: \(\Pprior(x) = \sum_{j=1}^M \frac{w_j}{b_j - a_j} \chi_{[a_j, b_j]}(x)\), where the weights \(w_j\) satisfy \(\sum_{j=1}^M w_j = 1\). Following the same process as above, the marginal distribution for \(Y_t\) is:
\begin{equation}
\begin{split}
P(Y_t = y) &= \sum_{j=1}^M \frac{w_j}{b_j - a_j}\left(\Phi\left(\frac{y - a_j}{\sqrt{\epsilon t}}\right) - \Phi\left(\frac{y - b_j}{\sqrt{\epsilon t}}\right)\right).
\end{split}
\end{equation}
The solution to the viscous HJ PDE~\eqref{eqt:HJsampler_SDE_HJ} is:
\begin{equation}
\Sepsbwd(x,T-t) = \epsilon\log P(Y_t = x) = \epsilon\log\left(\sum_{j=1}^M \frac{w_j}{b_j - a_j}\left(\Phi\left(\frac{x - a_{j}}{\sqrt{\epsilon t}}\right) - \Phi\left(\frac{x - b_{j}}{\sqrt{\epsilon t}}\right)\right)\right).
\end{equation}
The inference process, as described in~\eqref{eq:Euler–Maruyama}, becomes:
\begin{equation}
\begin{split}
&Z_{k+1} = Z_{k} + \partial_{x} \Sepsbwd(Z_k, t_k) \Delta t + \sqrt{\epsilon \Delta t} \xi_k\\
&= Z_{k} + \epsilon \frac{\frac{1}{\sqrt{\epsilon (T-t_k)}} \left(\sum_{j=1}^M \frac{w_j}{b_j-a_j}\left(\Phi'\left(\frac{Z_{k} - a_{j}}{\sqrt{\epsilon (T-t_k)}}\right) - \Phi'\left(\frac{Z_{k} - b_{j}}{\sqrt{\epsilon (T-t_k)}}\right)\right)\right)}{\sum_{j=1}^M \frac{w_j}{b_j-a_j}\left(\Phi\left(\frac{Z_{k} - a_{j}}{\sqrt{\epsilon (T-t_k)}}\right) - \Phi\left(\frac{Z_{k} - b_{j}}{\sqrt{\epsilon (T-t_k)}}\right)\right)} \Delta t + \sqrt{\epsilon \Delta t} \xi_k\\
&= Z_{k} + \sqrt{\frac{\epsilon}{2\pi (T-t_k)}}  \frac{\sum_{j=1}^M \frac{w_j}{b_j-a_j}\left(\exp\left(-\frac{1}{2\epsilon (T-t_k)}|Z_{k} - a_{j}|^2\right) - \exp\left(-\frac{1}{2\epsilon (T-t_k)}|Z_{k} - b_{j}|^2\right)\right)}{\sum_{j=1}^M \frac{w_j}{b_j-a_j}\left(\Phi\left(\frac{Z_{k} - a_{j}}{\sqrt{\epsilon (T-t_k)}}\right) - \Phi\left(\frac{Z_{k} - b_{j}}{\sqrt{\epsilon (T-t_k)}}\right)\right)} \Delta t + \sqrt{\epsilon \Delta t} \xi_k.
\end{split}
\end{equation}
The posterior distribution of \(Y_t\) given \(Y_T = \ydata\) for \(t \in (0,T)\) is:
\begin{equation}
\begin{adjustbox}{width=0.99\textwidth}$
\begin{split}
P(Y_t = \Bayesparam | Y_T =\ydata) &= \frac{P(Y_t = \Bayesparam) P(Y_T=\ydata | Y_t = \Bayesparam)}{P(Y_T = \ydata)}
= \frac{\left(\sum_{j=1}^M \frac{w_j}{b_j - a_j}\left(\Phi\left(\frac{\Bayesparam - a_j}{\sqrt{\epsilon t}}\right) - \Phi\left(\frac{\Bayesparam - b_j}{\sqrt{\epsilon t}}\right)\right)\right)\frac{1}{\sqrt{2\pi\epsilon (T-t)}} \exp\left(-\frac{1}{2\epsilon (T-t)} | \Bayesparam - \ydata|^2\right)}{\sum_{j=1}^M \frac{w_j}{b_j - a_j}\left(\Phi\left(\frac{\ydata - a_j}{\sqrt{\epsilon T}}\right) - \Phi\left(\frac{\ydata - b_j}{\sqrt{\epsilon T}}\right)\right)}.
\end{split}
$\end{adjustbox}
\end{equation}
The posterior distribution of \(Y_0\) given \(Y_T = \ydata\) is:
\begin{equation}
\begin{adjustbox}{width=0.99\textwidth}$
\begin{split}
P(Y_0 = \Bayesparam | Y_T =\ydata) &= \frac{P(Y_0 = \Bayesparam) P(Y_T=\ydata | Y_0 = \Bayesparam)}{P(Y_T = \ydata)}
= \frac{\left(\sum_{j=1}^M \frac{w_j}{b_j - a_j} \chi_{[a_j,b_j]} (\Bayesparam)\right)\frac{1}{\sqrt{2\pi\epsilon T}} \exp(-\frac{1}{2\epsilon T} | \Bayesparam-\ydata|^2)}{\sum_{j=1}^M \frac{w_j}{b_j - a_j}\left(\Phi\left(\frac{\ydata - a_j}{\sqrt{\epsilon T}}\right) - \Phi\left(\frac{\ydata - b_j}{\sqrt{\epsilon T}}\right)\right)}.
\end{split}
$\end{adjustbox}
\end{equation}

\subsubsection{$n$-dimensional Gaussian mixture prior}
We consider the \(n\)-dimensional Brownian motion problem with a Gaussian mixture prior. Let the prior be \(\Pprior\) as given in~\eqref{eqt:mixed_Gaussian}, and let the process be governed by \(dY_t = \sqrt{\epsilon}dW_t\). The marginal distribution is
\begin{equation}
P(Y_t = x) = \sum_{i=1}^\nprior \frac{w_i}{C_i(t)}\exp\left(-\frac{1}{2}(x-\Bayesparam_i)^T (\Sigma_i + \epsilon t I)^{-1} (x-\Bayesparam_i)\right),
\end{equation}
where $C_i(t) = \sqrt{(2\pi)^n\det(\Sigma_i + \epsilon t I)}$ is the normalization constant.

The solution to the viscous HJ PDE~\eqref{eqt:HJsampler_SDE_HJ} is
\begin{equation}
\Sepsbwd(x, T-t) = \epsilon\log \left(\sum_{i=1}^\nprior \frac{w_i}{C_i(t)}\exp\left(-\frac{1}{2}(x-\Bayesparam_i)^T (\Sigma_i + \epsilon t I)^{-1} (x-\Bayesparam_i)\right)\right).
\end{equation}
The inference process, as described in~\eqref{eq:Euler–Maruyama}, becomes  
\begin{equation}
\begin{adjustbox}{width=0.99\textwidth}$
\begin{split}
Z_{k+1} &= Z_k + (\nabla_x \Sepsbwd(Z_k, t_k)) \Delta t + \sqrt{\epsilon \Delta t} \xi_k\\
&= Z_k - \epsilon \frac{\sum_{i=1}^M \frac{w_i}{C_i(T-t_k)} (\Sigma_i + \epsilon (T-t_k) I)^{-1} (Z_k-\Bayesparam_i) \exp\left(-\frac{1}{2}(Z_k-\Bayesparam_i)^T (\Sigma_i + \epsilon (T-t_k) I)^{-1} (Z_k-\Bayesparam_i)\right)}{\sum_{i=1}^M \frac{w_i}{C_i(T-t_k)}\exp\left(-\frac{1}{2}(Z_k-\Bayesparam_i)^T (\Sigma_i + \epsilon (T-t_k) I)^{-1} (Z_k-\Bayesparam_i)\right)} \Delta t + \sqrt{\epsilon \Delta t} \xi_k.
\end{split}
$\end{adjustbox}
\end{equation}

The posterior distribution of \(Y_t\) given \(Y_T = \ydata\) for \(t \in [0,T)\) is 
\begin{equation}
\begin{adjustbox}{width=0.99\textwidth}$
\begin{split}
P(Y_t = \Bayesparam | Y_T =\ydata) &= \frac{P(Y_t = \Bayesparam) P(Y_T=\ydata | Y_t = \Bayesparam)}{P(Y_T = \ydata)}\\
&= \frac{\left(\sum_{i=1}^M \frac{w_i}{C_i(t)}\exp\left(-\frac{1}{2}(\Bayesparam-\Bayesparam_i)^T (\Sigma_i + \epsilon t I)^{-1}(\Bayesparam-\Bayesparam_i)\right)\right)\frac{1}{\sqrt{(2\pi\epsilon (T-t))^n}} \exp\left(-\frac{1}{2\epsilon (T-t)} \|\ydata - \Bayesparam\|^2\right)}{\sum_{i=1}^M \frac{w_i}{C_i(T)}\exp\left(-\frac{1}{2}(\ydata-\Bayesparam_i)^T(\Sigma_i + \epsilon T I)^{-1}(\ydata-\Bayesparam_i)\right)}\\
&= \sum_{i=1}^M \frac{\tilde w_i}{\sqrt{(2\pi)^n \det(M_i)}}\exp\left(-\frac{1}{2}(\theta - v_i)^T M_i^{-1} (\theta - v_i)\right),
\end{split}
$\end{adjustbox}
\end{equation}
which is a Gaussian mixture, where the $i$-th Gaussian has covariance matrix $M_i = \left((\Sigma_i + \epsilon t I)^{-1} + \frac{I}{\epsilon (T-t)}\right)^{-1}$ and mean $v_i = M_i\left((\Sigma_i + \epsilon t I)^{-1} \theta_i + \frac{\ydata}{\epsilon (T-t)}\right)$. The weight $\tilde w_i$ is 
\begin{equation}
\tilde w_i = \frac{\frac{w_i}{C_i(T)}\exp\left(-\frac{1}{2}(\ydata-\Bayesparam_i)^T(\Sigma_i + \epsilon T I)^{-1}(\ydata-\Bayesparam_i)\right)}{\sum_{j=1}^M \frac{w_j}{C_j(T)}\exp\left(-\frac{1}{2}(\ydata-\Bayesparam_j)^T(\Sigma_j + \epsilon T I)^{-1}(\ydata-\Bayesparam_j)\right)}.
\end{equation}

\subsection{OU process}

In general, if the matrix $B$ in the OU process is not diagonal, neither the posterior density function nor the sampling SDE~\eqref{eq:Euler–Maruyama} have analytical solutions. However, analytical formulas can be derived when $B$ is a diagonal matrix and the prior is a Gaussian mixture where each component has a diagonal covariance matrix.

Specifically, for one-dimensional cases, the analytical solutions can be obtained for the OU process with a Gaussian prior $\mathcal{N}(\Gpriorcenter, \sigma^2)$. Note that in this case, the matrix $B$ reduces to a scalar.
The solution to the Riccati ODE system is given by $\Sxx(t) = \frac{e^{-2B t}\sigma^2}{\epsilon} + \frac{1- e^{-2B t}}{2B}$, $\Sx(t) = e^{-B t}\Gpriorcenter$, and $\Sc(t) = \frac{\epsilon}{2}\log(2\pi e^{-2B t}\sigma^2 + \frac{\pi\epsilon(1- e^{-2B t})}{B})$.
The solution to the viscous HJ PDE~\eqref{eqt:HJsampler_SDE_HJ} is
\begin{equation}
\begin{adjustbox}{width=0.99\textwidth}$
\Sepsbwd(x, T-t) = -\frac{1}{2} (x-\Sx(t))^T\Sxx(t)^{-1}(x-\Sx(t)) - \Sc(t) = -\frac{\left|x-e^{-B t}\Gpriorcenter\right|^2}{\frac{2e^{-2B t}\sigma^2}{\epsilon} + \frac{1- e^{-2B t}}{B}} - \frac{\epsilon}{2}\log\left(2\pi e^{-2B t}\sigma^2 + \frac{\pi\epsilon(1- e^{-2B t})}{B}\right).
$\end{adjustbox}
\end{equation}
The inference process, as described in~\eqref{eq:Euler–Maruyama}, becomes  
\begin{equation}
\begin{split}
Z_{k+1} &= Z_k + (\partial_x \Sepsbwd(Z_k, t_k)) \Delta t + \sqrt{\epsilon \Delta t} \xi_k
= Z_k - \frac{Z_k-e^{-B (T-t_k)}\Gpriorcenter}{\frac{e^{-2B (T-t_k)}\sigma^2}{\epsilon} + \frac{1- e^{-2B (T-t_k)}}{2B}} \Delta t + \sqrt{\epsilon \Delta t} \xi_k.
\end{split}
\end{equation}
For any $0<t<s<T$, the process satisfies $Y_s = e^{-B (s-t)}Y_t + \sqrt{\frac{\epsilon(1- e^{-2B (s-t)})}{2B}}\xi$, where $\xi$ is a standard Gaussian random variable independent of $Y_t$.
The marginal distribution of $Y_t$ is
\begin{equation}
P(Y_t = x) = \frac{1}{\sqrt{2\pi e^{-2B t}\sigma^2 + \frac{\pi\epsilon(1- e^{-2B t})}{B}}}\exp\left(-\frac{1}{2 e^{-2B t}\sigma^2 + \frac{\epsilon(1- e^{-2B t})}{B}} |x-e^{-B t}\Gpriorcenter|^2 \right).
\end{equation}
The conditional distribution of $Y_T$ given $Y_t$ is
\begin{equation}
P(Y_T = y | Y_t = \Bayesparam) = \sqrt{\frac{B}{\epsilon \pi(1-e^{-2B (T-t)})}}\exp\left(-\frac{B}{\epsilon(1-e^{-2B (T-t)})} \left|y - e^{-B (T-t)} \Bayesparam\right|^2\right).
\end{equation}
The posterior distribution of $Y_t$ given $Y_T = \ydata$ for any $t\in [0,T)$ is 
\begin{equation}
\begin{adjustbox}{width=0.99\textwidth}$
\begin{split}
&P(Y_t = \Bayesparam | Y_T = \ydata) = \frac{P(Y_t = \Bayesparam)P(Y_T = \ydata | Y_t = \Bayesparam)}{P(Y_T = \ydata)} \\
&=\frac{\frac{1}{\sqrt{2\pi e^{-2B t}\sigma^2 + \frac{\pi\epsilon(1- e^{-2B t})}{B}}}\exp\left(-\frac{1}{2 e^{-2B t}\sigma^2 + \frac{\epsilon(1- e^{-2B t})}{B}} |\Bayesparam-e^{-B t}\Gpriorcenter|^2 \right) \sqrt{\frac{B}{\epsilon \pi(1-e^{-2B (T-t)})}}\exp\left(-\frac{B}{\epsilon(1-e^{-2B (T-t)})} |\ydata - e^{-B (T-t)} \Bayesparam|^2\right)}{\frac{1}{\sqrt{2\pi e^{-2B T}\sigma^2 + \frac{\pi\epsilon(1-e^{-2B T})}{B}}}\exp\left(-\frac{1}{2e^{-2B T}\sigma^2 + \frac{\epsilon(1-e^{-2B T})}{B}} |\ydata-e^{-B T}\Gpriorcenter|^2\right)}.
\end{split}
$\end{adjustbox}
\end{equation}
After simplification, this posterior distribution is Gaussian, with mean and variance given by:
\begin{equation}
\begin{split}
\E[Y_t | Y_T = \ydata] &= \frac{\epsilon(e^{-B t} - e^{B(t-2T)})\Gpriorcenter + 2B \sigma^2 e^{-B(T+t)}\ydata + \epsilon (e^{-B(T-t)} - e^{-B(T+t)})\ydata}{\epsilon(1-e^{-2B T}) + 2B \sigma^2e^{-2B T}},\\
\text{Var}[Y_t | Y_T = \ydata] &= \frac{\epsilon\left(\sigma^2e^{-2Bt} + \frac{\epsilon(1-e^{-2Bt})}{2B}\right)(1-e^{-2B(T-t)})}{\epsilon(1-e^{-2B T}) + 2B \sigma^2e^{-2B T}}.
\end{split}    
\end{equation}

\section{Supplementary results for the 1D Brownian motion example}\label{appendix:additional_numerics}

\begin{figure}[h]
    \centering
    \begin{subfigure}[b]{1\textwidth}
         \centering
         \includegraphics[width=\textwidth]{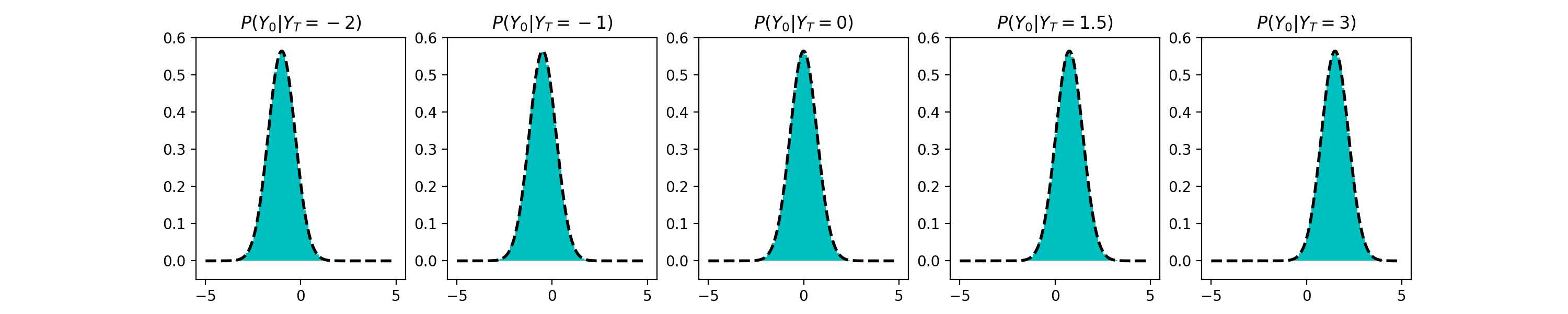}
         \caption{The analytic-HJ-sampler.}
     \end{subfigure}
     \begin{subfigure}[b]{1\textwidth}
         \centering
         \includegraphics[width=\textwidth]{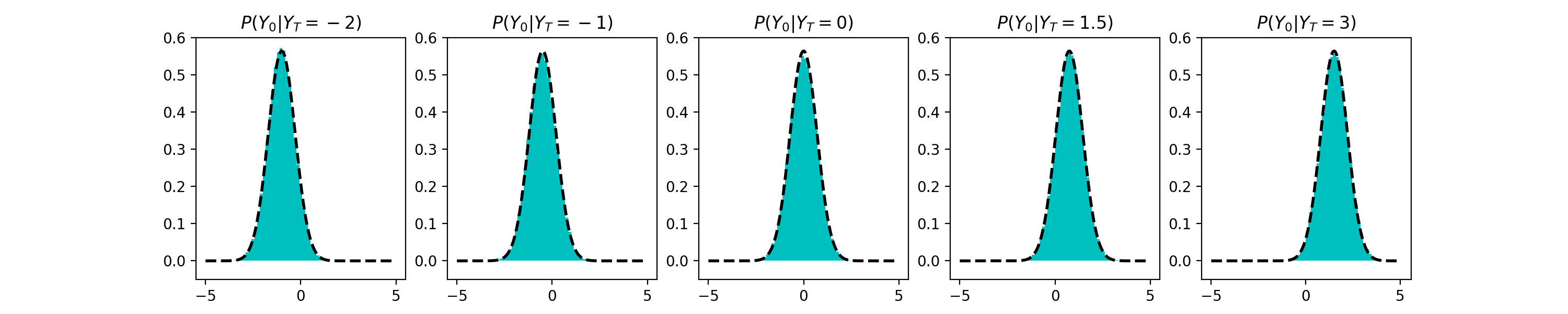}
         \caption{The SGM-HJ-sampler.}
     \end{subfigure}
    \caption{Histograms depicting the distribution of posterior samples for the scaled 1D Brownian motion case with a Gaussian prior, across different observation values $\ydata$. The SGM-HJ-sampler employs a neural network trained on \( t \in [0, T] \) with \( T = 1 \). The \textbf{black} dashed lines represent the exact posterior density functions (Gaussian). Each histogram is generated from $1\times10^6$ samples.}
    \label{fig:example_1_0}
\end{figure}

In this section, we provide supplementary results for the 1D Brownian motion example in Section~\ref{sec:example_1_1}, illustrating the inference of \( Y_0 \) given the observation of \( Y_T \) using both the analytic-HJ-sampler and the SGM-HJ-sampler. Figure \ref{fig:example_1_0} displays histograms of the posterior samples. For quantitative comparison, please refer to Table \ref{tab:example_1_1} in the main content.

\end{document}